\documentclass[twoside,11pt,dvipsnames]{article}

\usepackage{amsmath}
\usepackage{amssymb}
\usepackage{booktabs}
\usepackage{bbm}
\usepackage{mathtools}
\usepackage{xcolor}
\usepackage{subcaption}
\usepackage{multirow}
\usepackage{float}  %

\usepackage{jmlr2e}

\newcommand{\argmin}{\operatorname*{arg\,min}}
\newcommand{\fracpartial}[2]{\frac{\partial #1}{\partial  #2}}
\newcommand{\fracpartialsq}[2]{\frac{\partial^2 #1}{\partial  #2^2}}

\newtheorem{simp_assumption}{Simplifying Assumption}  %

\newif\ifcomments
\commentsfalse
\ifcomments
    \providecommand{\daniel}[2][]{{\protect\color{purple}{[Daniel:\textbf{#1} #2]}}}
    \providecommand{\jonathan}[2][]{{\protect\color{violet}{[Jonathan:\textbf{#1} #2]}}}
    \providecommand{\zayd}[2][]{{\protect\color{brown}{[Zayd:\textbf{#1} #2]}}}
\else
    \providecommand{\daniel}[2][]{}
    \providecommand{\jonathan}[2][]{}
    \providecommand{\zayd}[2][]{}
\fi

\usepackage{graphicx}
\newsavebox\CBox
\def\textBF#1{\sbox\CBox{#1}\resizebox{\wd\CBox}{\ht\CBox}{\textbf{#1}}}

\newcommand{\citepos}[1]{\citeauthor{#1}'s~(\citeyear{#1})}

\usepackage{comment}

\newcommand{\revised}[1]{#1}
\usepackage{lastpage}
\jmlrheading{24}{2023}{1-\pageref{LastPage}}{4/22}{5/23}{22-0449}{Jonathan Brophy, Zayd Hammoudeh, and Daniel Lowd}

\ShortHeadings{Influence Estimation for Gradient-Boosted Decision Trees}{Brophy, Hammoudeh, and Lowd}
\firstpageno{1}

\makeatletter%
\begin{document}

\title{Adapting and Evaluating Influence-Estimation Methods for Gradient-Boosted Decision Trees}

\author{\name Jonathan Brophy \email jbrophy@cs.uoregon.edu \\
        \name Zayd Hammoudeh \email zayd@cs.uoregon.edu \\
        \name Daniel Lowd \email lowd@cs.uoregon.edu \\
       \addr Department of Computer and Information Science \\
       University of Oregon \\
       Eugene, OR 97403, USA
}

\editor{Eric Laber}

\maketitle

\begin{abstract}%
Influence estimation analyzes how changes to the training data can lead to different model predictions; this analysis can help us better understand these predictions, the models making those predictions, and the data sets they're trained on. However, most influence-estimation techniques are designed for deep learning models with continuous parameters. Gradient-boosted decision trees (GBDTs) are a powerful and widely-used class of models; however, these models are black boxes with opaque decision-making processes. In the pursuit of better understanding GBDT predictions and generally improving these models, we adapt recent and popular influence-estimation methods designed for deep learning models to GBDTs. Specifically, we adapt representer-point methods and TracIn, denoting our new methods TREX and BoostIn, respectively; source code is available at \url{https://github.com/jjbrophy47/tree\_influence}. We compare these methods to LeafInfluence and other baselines using 5 different evaluation measures on 22 real-world data sets with 4 popular GBDT implementations. These experiments give us a comprehensive overview of how different approaches to influence estimation work in GBDT models. We find BoostIn is an efficient influence-estimation method for GBDTs that performs equally well or better than existing work while being four orders of magnitude faster. Our evaluation also suggests the gold-standard approach of leave-one-out~(LOO) retraining consistently identifies the single-most influential training example but performs poorly at finding the most influential set of training examples for a given target prediction.
\end{abstract}


\section{Introduction}
\label{sec:introduction}

\emph{Influence estimation} analyzes how changes to the training data can lead to different model predictions and helps investigate questions such as, ``how would this prediction change if I were to \emph{remove} these training examples?'' Analyzing the influence of training data on model predictions can provide a better understanding of model behavior~\citep{adadi2018peeking,ferreira2020people,tjoa2020survey}, improve model quality via data set or model debugging~\citep{yeh2018representer}, help quantify the value of different training examples~\citep{ghorbani2019data}, and much more~\citep{koh2017understanding}.

\revised{
For example, \citet{braun2022influence} leverage training data influence analysis to gain actionable insights into real-world black box models -- specifically, to mitigate the effect of labeling noise on automated medical image diagnosis. 
Mislabeled training instances are often memorized by expressive models, resulting in these instances displaying abnormal influence scores \citep{feldman2020neural}.
\citeauthor{braun2022influence} apply this insight by automatically downweighting training instances with outlier influence scores~\citep{pruthi2020estimating}, resulting in medical imaging models that are both more accurate and more robust. 
By mitigating the effect of mislabeled data, \citeauthor{braun2022influence}'s improved models also require less training data, reducing the need for expensive, domain-specific annotators. 
\citepos{hammoudeh2022identifying} influence estimation survey details additional applications of influence analysis, including algorithmic fairness~\citep{black2021leaveoneout}, defenses against poisoning attacks~\citep{hammoudeh2022identifying}, active learning~\citep{liu2021influenceactivelearning},
and data augmentation~\citep{oh2021influencetensorcompletion}.%
}

Existing work in this burgeoning subfield has focused mainly on deep learning systems~\citep{koh2017understanding,yeh2018representer,pruthi2020estimating}. However, gradient-boosted decision trees~(GBDTs) are a powerful class of models that generally outperform deep learning and other more traditional learning algorithms on \emph{tabular} data~\citep{prokhorenkova2018catboost,popov2019neural}, one of the most common types of data used in industry~\citep{sun2019supertml}; they are also regularly used to win challenges on data-competition websites such as Kaggle and DrivenData~\citep{fogg2016anthony,bojer2021kaggle}. Despite their predictive prowess, GBDTs are black-box models with opaque decision-making processes~\citep{lundberg2017unified,lundberg2018consistent}. Influence estimation may help us better understand how GBDTs make predictions, remove unwanted biases, and ultimately improve our models.

Influence functions were one of the first methods proposed for influence estimation, initially in differentiable models~\citep{koh2017understanding} and later in trees~\citep{sharchilev2018finding}. Representer point methods~\citep{yeh2018representer} offer greater efficiency and an interpretation of the model as a sum over contributions from all training points, based on representer theorems. More recently, dynamic influence estimation methods such as TracIn~\citep{pruthi2020estimating} and HyDRA~\citep{chen2021hydra} look at the influence of training examples throughout the training process, rather than only focusing on the properties of the final model. 

However, tree ensembles have not yet benefited from the advantages of new representer point and dynamic influence estimation methods.
Thus, we adapt these techniques designed for deep learning models
to GBDTs. Specifically, we propose \emph{TREX}~(\S\ref{sec:trex}), our adaptation of representer-point methods~\citep{yeh1998modeling}, and \emph{BoostIn}~(\S\ref{sec:boostin}), our adaptation of TracIn~\citep{pruthi2020estimating}.

Since influence estimation may behave differently in trees than in deep learning models, we evaluate a wide range of techniques with varying methodologies to better understand how best to do influence estimation in GBDTs.
However, empirically evaluating the merit of different influence methods tends to vary in the literature, in part because there is no single criterion for defining the optimal set of training examples that influence a prediction~\citep{koh2017understanding,yeh2018representer,hanawa2021evaluation}.
Thus, to get a broad overview of performance, we focus on quantitative evaluations rather than the qualitative analysis that is sometimes used to evaluate influence-estimation methods. Specifically, we approximate the optimal set by ranking training examples based on their individual influences and measure changes in model predictions after performing some operation on a subset of the most influential examples~(e.g., removal) and retraining the model; this evaluation methodology provides a quantitative measure of the fidelity of each method, that is, how well a method predicts actual model behavior. We use different evaluation measures since the effectiveness of an influence method may be operation dependent. For example, the training instances identified as most influential may have the biggest impact on a target prediction when \emph{removed}, but not when \emph{their labels are flipped} instead; hence, we evaluate each influence method in various contexts~(\S\ref{sec:methodology}).

We conduct extensive experiments on 22 real-world data sets to compare eight different influence-estimation methods using 5~different evaluation measures and 4 popular modern GBDT implementations~: LightGBM~\citep{ke2017lightgbm}, XGBoost~\citep{chen2016xgboost}, CatBoost~\citep{prokhorenkova2018catboost}, and gradient boosting from Scikit-Learn~\citep{scikit-learn}. Our results suggest the adaptation of TracIn~\citep{pruthi2020estimating}---which we denote~\emph{BoostIn}---is a solid default choice for influence estimation in GBDTs. BoostIn is over \emph{four orders of magnitude} faster than the current state-of-the-art:~LeafInfluence~\citep{sharchilev2018finding}---the adaptation of Influence Functions~\citep{koh2017understanding} to GBDTs---and provides better influence estimates than competing methods in the majority of tested contexts, on average~(\S\ref{sec:result_summary}).

Furthermore, leave-one-out~(LOO)---removing training examples one at a time, retraining the model after each removal, and measuring the prediction difference between the original and retrained models---consistently identifies the~\emph{single}-most influential training example for a given target prediction by definition; however, we find LOO does a poor job of selecting the most influential~\emph{set} of training examples that contribute to a given target prediction~(\S\ref{sec:loo_fragility}). We find LOO often induces small but significant changes to the structure of the trees as a result of removing one or very few training examples; we also observe LOO has a very low correlation with any of the other influence-estimation methods~(\S\ref{sec:correlation}).
Thus, when considering more training examples for an influence estimate, we find methods using a fixed tree-structure assumption are better able to find a \emph{set} of training examples that most influence a given target prediction than methods that do not.

In the following sections, we detail the problem of influence estimation both generally and in the context of GBDTs~(\S\ref{sec:problem_formulation}). We review previous work adapting influence functions to GBDTs, and then present our new methods adapting more recent approaches~(\S\ref{sec:adapting_influence_methods}). We evaluate all methods in multiple contexts, including removing and relabeling the most influential training examples and observing the effect on one and many target predictions after retraining the model~(\S\ref{sec:methodology},~\S\ref{sec:result_summary}). We then describe additional related work~(\S\ref{sec:related_work}), and finally conclude and discuss avenues for future work~(\S\ref{sec:conclusion}).

\section{Preliminaries and Problem Formulation}
\label{sec:problem_formulation}

\revised{For \emph{feature space} ${\mathcal{X} \subseteq \mathbb{R}^p}$ and \emph{target space} ${\mathcal{Y} \in \{\{-1, +1\}, \mathbb{Z}, \mathbb{R}\}}$, let ${\mathcal{Z} \coloneqq \mathcal{X} \times \mathcal{Y}}$.}
Let \revised{${D \coloneqq \{(x_i, y_i)\}_{i=1}^{n} \subseteq \mathcal{Z}}$} be the full \emph{training set}, where\ ${x_i \in \mathcal{X}}$ is a $p$-dimensional vector ${(x_{i,j})_{j=1}^{p}}$ and ${y_i \in \mathcal{Y}}$.
\revised{%
We use ${\textbf{x}\coloneqq\{x_i\}_{i=1}^{n}}$ and ${\textbf{y}\coloneqq\{y_i\}_{i=1}^{n}}$ to denote the full training \emph{feature matrix} and \emph{target vector}, respectively. 
${z_i\coloneqq(x_i, y_i) \in D}$ is the~$i$th training instance and ${z_e \coloneqq(x_e, y_e) \in \mathcal{Z}}$ an arbitrary target instance. 

Let ${A: 2^{D} \rightarrow \mathcal{F}}$ be a \textit{learning algorithm}, where $2^D$ is the \emph{power set} of $D$ and $\mathcal{F}$ is the \emph{hypothesis class}.  ${f \coloneqq A(D)}$ is the \textit{model} $A$ yields when training on data set~$D$. A \emph{loss function} is denoted ${\ell : \mathcal{Y}\times\mathcal{Y} \rightarrow \mathbb{R}}$. 
}

We use notation~``$\sim$'' to denote a deep learning variable, to distinguish it from those corresponding to trees; for example, $\tilde{f}$~denotes a deep learning model; $\tilde{\ell}$~is a deep learning loss function, etc.

\subsection{Gradient-Boosted Decision Trees (GBDTs)}

Gradient boosting~\citep{friedman2000additive,friedman2001greedy} is a powerful machine-learning algorithm that iteratively adds weak learners to construct model $f: \mathcal{X} \rightarrow \mathbb{R}$ that minimizes some empirical risk (i.e.,~$f = \argmin_{\hat{f}} \sum_{i=1}^n \ell(y_i, \hat{f}(x_i))$). The model is defined recursively where:
\begin{align}
    f_{0}(x) &= \gamma \\
    &\hspace{5.5pt}\vdots \nonumber \\
    f_{t}(x) &= f_{t-1}(x) + m_{t}(x),
\end{align}
in which~$\gamma$ is an initial estimate~(typically the mean of the training targets: \revised{${\Bar{\textbf{y}} \coloneqq \frac{1}{n}\sum_i y_i}$}), $f_{t-1}$ is the model at the previous iteration, $m_{t}$ is a weak learner added during iteration $t$ to improve the model. Gradient-boosted decision trees~(GBDTs) use regression trees as the weak learners, with each tree typically chosen to approximate the negative gradient~\citep{malinin2021uncertainty}:
\begin{align}
    g_{t}(x, y) &= \fracpartial{\ell(y, \hat{y})}{\hat{y}}\Big|_{\hat{y}=f_{t-1}(x)},
    \label{eq:gbdt_gradient} \\
    \therefore m_{t} &= \argmin_{\hat{m}}\ \frac{1}{n} \sum_{i=1}^{n} (-g_{t}(x_i, y_i) - \hat{m}(x_i))^2.
    \label{eq:gbdt_weak_learner}
\end{align}
The weak learner at iteration~$t$ partitions the instance space into a set of~$M_t$ disjoint regions \revised{\{$r_{t,1}, \hdots, r_{t, M_{t}}$\}}. Each region~$r_{t,l}$ is called a \emph{leaf} whose parameter value $\theta_{t, l}$ is typically determined (given a fixed structure) using a one-step Newton-estimation method~\citep{chen2016xgboost,ke2017lightgbm}:
\begin{align}h_{t}(x, y) &= \fracpartialsq{\ell(y, \hat{y})}{\hat{y}}\Big|_{\hat{y}=f_{t-1}(x)},
\label{eq:gbdt_hessian} \\
\therefore \theta_{t, l} &=
-\left(
\frac
{\sum_{i \in I_{t, l}} g_{t}(x_i, y_i)}
{\sum_{i \in I_{t, l}} h_{t}(x_i, y_i) + \lambda}
\right)
\eta
\end{align}
in which $I_{t, l} = \{z_j~|~z_j \in r_{t, l}\}_{j=1}^{n}$ is the instance set of the $l$th leaf for the $t$th tree, $g_t(x_i, y_i)$ and $h_t(x_i, y_i)$ are the first and second derivatives of the $i$th training instance with respect to~$\hat{y}_i$,  respectively~(Eqs.~\ref{eq:gbdt_gradient} and~\ref{eq:gbdt_hessian}), $\lambda$ is a regularization constant, and~$\eta$ is the learning rate. Thus, $m_t$ can be written as
\begin{align}
m_{t}(x) = \sum_{l=1}^{M_t} \theta_{t, l} \mathbbm{1}[x \in r_{t, l}]
\end{align}
where~$\mathbbm{1}$ is the indicator function.
The final GBDT model~$f=f_T$ generates a prediction for a target example $x_e$ by summing the initial estimate with the values for the leaves $x_e$ is assigned to across all~$T$ trees: ${f(x_e) = f_0 + \sum_{t=1}^T m_t(x_e)}$.

\subsection{Influence Estimation}
\label{sec:influence_estimation}

Existing influence estimation methods attempt to compute the influence of each training example~$z_i$ on the prediction of a given target example~$z_e$. 
Informally, a training example is \emph{influential} if its inclusion in the training data impacts the learned model and its predictions. Leave-one-out (LOO) defines influence as the difference between training with the entire training data set and training with a specified example excluded. The Shapley value is similar to LOO, but computes an expectation of LOO over all (exponentially many) subsets of the original training data. In general, the impact of removing an example depends on which other examples are removed. Existing methods compute a single number for each individual training example in order to
generate a \emph{ranking} amongst the training data. Standard influence estimation metrics then evaluate this ordering by analyzing sequential subsets of the most influential examples~\citep{Hammoudeh:2022:InfluenceSurvey}.

Following previous work~\citep{koh2017understanding,pruthi2020estimating}, we analyze the influence of training examples on a given target example by computing their impact on the \emph{loss} of that target example.\footnote{A \emph{target example} is simply the example we compute influence values for,
either in the train or test set.}
\revised{%
To this end, we define an
\emph{influence-estimation method} ${\mathcal{I} : \mathcal{Z} \times \mathcal{Z} \rightarrow \mathbb{R}}$ as a function that quantifies the effect of training instance~$z_i$ on target example~$z_e$ given learning algorithm~$A$, training set~$D$, and loss function~$\ell$.  
Note that \emph{influence value} ${\mathcal{I}(z_i, z_e)}$ can be positive or negative, denoting that $z_i$ \emph{reduces} or \emph{increases} the model's loss on target~$z_e$, respectively}.%
\footnote{Following convention~\citep{pruthi2020estimating}, we refer to training examples that reduce the target example's loss as \emph{proponents} and those that increase loss as \emph{opponents}.}

\emph{LOO: Leave-One-Out.} The simplest and most intuitive approach to estimating the influence of a training example~$z_i$ on a target example~$z_e$ is to ignore the existing model and simply rerun $\revised{A}$ on a pruned data set without~$z_i$, and then measure the change in loss\footnote{When~$\revised{A}$ is non-deterministic, one would need to retrain \emph{multiple} times on the same data set to compute the \emph{expected} change in loss.} on~$z_e$:
\begin{align}
    \mathcal{I}_{LOO}(z_i, z_e) = \ell(y_e, \revised{A}(D \setminus \revised{\{z_i\}})(x_e)) - \ell(y_e, \revised{A}(D)(x_e)).
    \label{eq:loo_influence}
\end{align}
Repeating this naive approach for all training examples is known as \emph{leave-one-out}~(LOO) \citep{cook1982residuals}. LOO is agnostic to virtually all machine learning models, easy to understand, easy to implement, and is regularly described as a gold-standard influence-estimation method~\citep{ghorbani2019data}. Empirically, we find this approach performs relatively poorly for identifying a practical \emph{set} of influential examples for a given target prediction in GBDTs~(\S\ref{sec:loo_fragility}); furthermore, this approach becomes prohibitively expensive as the data set size or model complexity increases.

\emph{Data Shapley: Expected Marginal Influence.} A different way of determining the contributions of individual examples belonging to a group is via Shapley values, a game-theoretic method for distributing contributions among involved players~\citep{shapley1953value}. Data Shapley~\citep{ghorbani2019data} is a model-agnostic approach that applies the idea of Shapley values to influence estimation in which the marginal contribution of~$z_i$ on the loss of~$z_e$ can be written as:
\begin{align}
\mathcal{I}_{DShap}(z_i, z_e) = C \sum_{S \subseteq D \setminus \revised{\{z_i\}}} \frac{\ell(y_e, \revised{A}(S)(x_e)) - \ell(y_e, \revised{A}(S \cup \{z_i\})(x_e))}{\binom{n-1}{|S|}},
\label{eq:dshap_influence}
\end{align}
where~$C$ is a constant and~$S$ represents all possible subsets of the training data without~$z_i$.
Equation~\eqref{eq:dshap_influence} computes the \emph{expected} marginal impact of a single example given a subset of the training data, but is also far more intractable than LOO.

\emph{SubSample: \revised{Simplified Data Shapley}.} More recently, \citet{feldman2020neural} propose a method that quantifies the amount of memorization acquired by a deep learning model during training. Their approach -- \emph{SubSample} -- computes the memorization level of a training example as well as the influence of a training example on the accuracy of a learning algorithm~$\revised{A}$ by training~$\tau$ different models on random subsets of~$D$.
SubSample combines the predictions of these $\tau$~models to estimate each training instance's marginal influence. \revised{Thus, SubSample can be viewed as a simplified version of Data Shapley that considers a single pre-specified subset size instead of all possible subset sizes}.
\revised{The expected marginal-influence effect of~$z_i$ on the loss of~$z_e$ is then defined as}:
\begin{align}
\revised{%
\nonumber
\mathcal{I}_{Sub}(z_i, z_e) =~
&\mathbb{E}_{S \sim U({D} \setminus \{z_i\}, m - 1)}[\ell(y_e, \revised{A}(S \cup \{z_i\})(x_e))] \\
&- \mathbb{E}_{S \sim U({D \setminus \{z_i\}}, m)}[\ell(y_e, \revised{A}(S)(x_e))].
}
\label{eq:subsample_influence}
\end{align}
In Equation~\eqref{eq:subsample_influence}, ${U(\revised{D}, m)}$ represents the uniform distribution over all subsets of $\revised{D}$ with size ${m \coloneqq \lvert S \rvert < n}$. For this approach to be tractable and produce meaningful influence values, $m$ must be small enough to provide sufficient cases in which~$z_i \not\in S$, but large enough to train reasonable approximations to the original model. SubSample is much more \revised{efficient than Data Shapley~(Eq.~\ref{eq:dshap_influence}) and in most cases LOO~(Eq.~\ref{eq:loo_influence}), especially when $\tau \ll n$}.

The three influence estimators above work for \emph{any} model. A number of other influence-estimation methods have been developed in the context of deep learning and other differentiable, parametric models~\citep{koh2017understanding,pruthi2020estimating,yeh2018representer}. These parametric approaches cannot be applied directly to discrete, tree-structured, non-parametric models like GBDTs. In the next section, we introduce these deep learning-based influence estimators, and describe the adaptation of each method to estimate influence in GBDTs.

\section{Adapting Influence Methods to GBDTs}
\label{sec:adapting_influence_methods}

This section first reviews \emph{LeafRefit}~(\S\ref{sec:leaf_refit}) and \emph{LeafInfluence}~(\S\ref{sec:leafinf}), work by~\cite{sharchilev2018finding} adapting influence functions~\citep{koh2017understanding} to GBDTs. Next, we introduce \emph{BoostIn}~(\S\ref{sec:boostin}), our adaptation of TracIn~\citep{pruthi2020estimating}, and theoretically compare BoostIn to LeafInfluence. Lastly, we describe our adaptation of representer-point methods~\citep{yeh2018representer} to GDBTs---a method we denote~\emph{TREX}~(\S\ref{sec:trex}).
\revised{For a more detailed discussion and comparison of existing parametric influence analysis methods, see the survey of \citet{Hammoudeh:2022:InfluenceSurvey}}.

Before describing the influence estimation methods specific to GBDTs, we define an important assumption made by all the tree-based methods in this section.

\revised{%
\begin{simp_assumption}
\label{assumption:fixed_structure}
\textnormal{(Fixed Structure Approximation)}
    ${\forall S \subseteq D}$, GBDT models ${A(D)}$ and ${A(D \setminus S)}$ have the same structure.%
\end{simp_assumption}
}%

\revised{Assumption~\ref{assumption:fixed_structure} allows the influence of $z_i$ to be estimated while treating the structure of each tree as fixed, where feature splits are considered part of the structure of each tree, precluding the influence of~$z_i$ on any node split. Assumption~\ref{assumption:fixed_structure} thus enables the methods detailed in this section to efficiently estimate influence values for larger datasets. Without this assumption, estimating the effect of removing $z_i$ from a GBDT model would require considering all the possible ways in which the structure of each tree in the ensemble could change due to the removal of $z_i$, and how those changing structures would impact the loss on the target example $z_e$. 
}

\subsection{LeafRefit: LOO with Fixed Tree Structures}
\label{sec:leaf_refit}

To estimate the influence of a training example on a target example \emph{without} having to retrain from scratch,~\citet{sharchilev2018finding} develop \emph{LeafRefit}, a method that computes the influence of~$z_i$ on the loss of~$z_e$ in GBDTs by \emph{refitting} all leaf values without~$z_i$.
\revised{LeafRefit is a theoretically efficient alternative to LOO that assumes a fixed structure while refitting leaves and computing influence values~(Assumption~\ref{assumption:fixed_structure})}.

\revised{However, despite the theoretical advantage of LeafRefit avoiding recomputing node splits or retraining completely from scratch, LeafRefit still needs to recompute leaf values, which can be expensive}.

\subsection{LeafInfluence: Adapting Influence Functions}
\label{sec:leafinf}

Influence functions~\citep{jaeckel1972infinitesimaljackknife,hampel1974influencecurve} is a technique from robust statistics that analyzes how the continuous parameters~$\tilde{\theta}$ of a differentiable model change when a training example~$z_i$ is \revised{upweighted by a small amount~$w_i$, resulting in updated parameters {\small ${\tilde{\theta}_{w_i} \coloneqq \argmin_{\tilde{\theta} \in \tilde{\Theta}} \frac{1}{n} \sum_{j=1}^{n} \tilde{\ell}(z_j, \tilde{\theta}) + w_i \tilde{\ell}(z_i, \tilde{\theta})}$} where {\small $\tilde{\ell}(z_i, \tilde{\theta})$} is the loss of~$z_i$ when using parameters~{\small $\tilde{\theta}$}}. The influence of~$w_i$ on the model parameters can be approximated \emph{without} having to retrain from scratch:
\begin{align}
\frac{d \tilde{\theta}_{w_i}}{d w_i}\Big|_{w_i=0} = -H_{\tilde{\theta}}^{-1} \nabla_{\tilde{\theta}} \tilde{\ell}(z_i, \tilde{\theta}),
\end{align}
where~$H_{\tilde{\theta}} = \frac{1}{n} \sum_{i=1}^{n} \nabla_{\tilde{\theta}}^{2} \tilde{\ell}(z_i, \tilde{\theta})$~\citep{cook1982residuals}. \cite{koh2017understanding} use this result to develop an influence estimator for deep learning models by analyzing the changing parameters due to upweighting~$z_i$, and then analyzing the effect the changing model parameters have on the loss of a target example~$z_e$:
\begin{align}
\mathcal{I}_{IF}(z_i, z_e) &= \nabla_{\tilde{\theta}}\tilde{\ell}(z_e, \tilde{\theta})^{\intercal} \frac{d \tilde{\theta}_{w_i}}{d w_i} \Big|_{w_i=0} \nonumber \\
&= -\nabla_{\tilde{\theta}}\tilde{\ell}(z_e, \tilde{\theta})^{\intercal}H_{\tilde{\theta}}^{-1} \nabla_{\tilde{\theta}}\tilde{\ell}(z_i, \tilde{\theta}).
\end{align}
\citepos{koh2017understanding} method (henceforth referred to simply as ``influence functions'') provides an efficient approximation to LOO for shallow deep learning models with strictly convex and twice-differentiable loss functions; for larger models or non-convex losses, however, the approximation is typically poor~\citep{basu2020influence,bae2022influencefunctionsquestion,hammoudeh2022identifying}.

For GBDTs, \cite{sharchilev2018finding} develop \emph{LeafInfluence}, an adaptation of influence functions and an approximation of LeafRefit that analyzes the \emph{perturbation} of leaf values and the resulting effect on the loss of~$z_e$, where:
\begin{align}
\mathcal{I}_{LI}(z_i, z_e) &= \fracpartial{\ell(y_e, \hat{y}_e)}{w_i}\Big|_{\hat{y}_e=f(x_e)} \nonumber \\
&= \fracpartial{\ell(y_e, \hat{y}_e)}{\hat{y}_e}\Big|_{\hat{y}_e=f(x_e)} \cdot \fracpartial{f(x_e)}{w_i}.
\label{eq:leaf_influence}
\end{align}
Equation~\eqref{eq:leaf_influence} approximates the change in $z_e$'s loss as a result of upweighting $z_i$.\footnote{In our experiments, we negate Eq.~\eqref{eq:leaf_influence} to be consistent with the concept of proponents and opponents, defined in \S\ref{sec:influence_estimation}.} The effect of upweighting $z_i$ on the \emph{final} GBDT model~(second term in Eq.~\ref{eq:leaf_influence}) is defined as:
\begin{align}
\fracpartial{f(x_e)}{w_i} = \sum_{t=1}^{T} \fracpartial{\theta_{t, l_e}(f_{t-1}(\textbf{x}))}{w_i},
\end{align}
in which~$l_e$ is the index of the leaf assigned to~$x_e$ at iteration~$t$, $\theta_{t, l_e}$ is the corresponding leaf value, and $f_{t-1}(\textbf{x})$ are the intermediate predictions at iteration~$t-1$. The partial derivative of the $l_e$th leaf at iteration~$t$ with respect to the $i$th training instance is defined as:
\begin{align}
\fracpartial{\theta_{t, l_e}(f_{t-1}(\textbf{x}))}{w_i} = -\frac{\mathbbm{1}[i \in I_{t, l_e}] (h_{t, i} \theta_{t, l_e} + g_{t, i}) + \sum_{j \in I_{t, l_e}} (k_{t, j} \theta_{t, l_e} + h_{t, j}) J(f_{t-1}(\textbf{x}))_{i,j}}{\sum_{j \in I_{t, l_e}}  h_{t, j}},
\label{eq:leaf_influence_leaf_derivative}
\end{align}
in which~$g_{t, i}=g_t(x_i, y_i)$~(Eq.~\ref{eq:gbdt_gradient}), $h_{t, i}=h_t(x_i, y_i)$~(Eq.~\ref{eq:gbdt_hessian}), $k_{t, i}=\partial^3 \ell(y_i, \hat{y}_i) / \partial \hat{y}_i^3|_{\hat{y}_i=f_{t-1}(x_i)}$, and $J(f_{t-1}(\textbf{x}))_{i,j} = J(f_{t-2}(\textbf{x}))_{i,j} + \partial \theta_{t-1, R_{t-1}(x_j)}(f_{t-2}(\textbf{x})) / \partial w_i$ is a Jacobian matrix that accumulates the changing intermediate predictions of all training examples as a result of upweighting~$z_i$.
Thus, the estimated influence of training example~$z_i$ on the loss of~$z_e$ can be found by running~$x_e$ through a new ensemble with new leaf values defined by Eq.~\eqref{eq:leaf_influence_leaf_derivative} and multiplying the result by the derivative of the loss with respect to the original prediction~(Eq.~\ref{eq:leaf_influence}).

Approximating the influence of a \emph{single} training example via LeafInfluence~(Eq.~\ref{eq:leaf_influence}) may be more efficient than LeafRefit, but the computation is an expensive operation in general, mainly due to the \emph{cascade effect} of changing predictions. For example, upweighting~$z_i$ changes the second-iteration predictions of all examples in the same leaf as~$z_i$, which then changes the leaf values for all leaves containing those examples in the second iteration and the predictions of the examples in those leaves for the third iteration; this process repeats for subsequent iterations, potentially necessitating the tracking and updating of all intermediate training-example predictions throughout the ensemble. The runtime complexity of LeafInfluence for computing~$\mathcal{I}_{LI}(z_i, z_e)$ is~$O(Tn^2)$, and computing influence values for \emph{all} training examples is~$O(Tn^3)$.
Empirically, we find LeafInfluence to be only marginally faster than LeafRefit, and surprisingly even slower than simply retraining from scratch~(\S\ref{sec:runtime}).

\emph{LeafInfluence-SinglePoint: Mitigating the Cascade Effect.} By restricting LeafInfluence to analyze \emph{only} the intermediate predictions of~$z_i$ and the parameters~(leaf values) containing~$z_i$,~\cite{sharchilev2018finding} introduce LeafInfluence-SinglePoint~(which we henceforth call LeafInfSP), a rough approximation to the main proposed approach~(Eq.~\ref{eq:leaf_influence}):
\begin{align}
\mathcal{I}_{LI_{SP}}(z_i, z_e) = \fracpartial{\ell(y_e, \hat{y}_e)}{\hat{y}_e}\Big|_{\hat{y}_e=f_T(x_e)} \cdot \left(\sum_{t=1}^{T} \mathbbm{1}[R_t(x_i) = R_t(x_e)] \left( \fracpartial{\theta_{t, R_t(x_i)}(f_{t-1}(x_i))}{w_i} \right) \right)
\label{eq:leafinfsp_influence}
\end{align}
in which
\begin{align}
\fracpartial{\theta_{t, l=R_t(x_i)}(f_{t-1}(x_i))}{w_i} = -\frac{(g_{t, i} + h_{t, i} \theta_{t, l}) + (h_{t, i} + k_{t, i} \theta_{t, l}) J(f_{t-1}(x_i))}{\sum_{j \in I_{t, l}} w_j h_{t, j}},
\end{align}
\revised{${R_t : \mathcal{X} \rightarrow \{1, \ldots, M_t\}}$ is a function that maps an instance~$x$ to the $l$th leaf of the $t$th tree}, and~$J(f_{t-1}(x_i)) = J(f_{t-2}(x_i)) + \partial \theta_{t-1, R_{t-1}(x_i)}(f_{t-2}(x_i)) / \partial w_i$ analyzes the changing intermediate predictions of \emph{only}~$z_i$ throughout the ensemble, mitigating the problem of approximating the change in parameter values for leaves which do not contain~$z_i$. Both LeafInfSP and LeafInfluence approximate the total change in model parameters and estimate the effect of this change on the final model prediction. In the next section, however, we introduce BoostIn, a method that estimates the influence of~$z_i$ on~$z_e$ \emph{throughout} the training process. We then compare BoostIn to LeafInfSP and LeafInfluence, highlighting the similarities and differences.

\subsection{BoostIn: Adapting TracIn}
\label{sec:boostin}

\emph{TracIn}~\citep{pruthi2020estimating} is an influence-estimation method designed for deep learning models that analyzes the impact of~$z_i$ on~$z_e$ \emph{throughout} the training process. TracIn is based on the fundamental theorem of calculus that decomposes the difference between a function at two points using the gradients along the path between those two points.

The idealized version of TracIn assumes every model update uses one training example~$z^t$ at each iteration~$t$ and thus defines the influence of~$z_i$ as the total reduction in loss on~$z_e$ whenever~$z_i$ is used to update the model. More formally, the idealized version of TracIn can be defined as:
\begin{equation}
\mathcal{I}_{TracInIdeal}(z_i, z_e) = \sum_{t: z^t = z_i} \tilde{\ell}(z_e, \tilde{\theta}_{t}) - \tilde{\ell}(z_e, \tilde{\theta}_{t+1}).
\end{equation}
This notion of influence has the appealing property that the sum of influences for all training examples on~$z_e$ is \emph{exactly} the total reduction in loss during training:
\begin{equation}
\sum_{i=1}^{n} \mathcal{I}_{TracInIdeal}(z_i, z_e) = \tilde{\ell}(z_e, \tilde{\theta}_{0}) - \tilde{\ell}(z_e, \tilde{\theta}_{T}).
\end{equation}
However, deep learning models are almost never trained in this fashion, and are typically trained using a batch or minibatches. Furthermore, the target example would need to be known ahead of training, or the entire training process would need to be repeated, which is generally intractable. Thus, a reasonable heuristic approximation is to save the model at various \emph{checkpoints} throughout training in which it is assumed each training example has been processed exactly once between checkpoints;
the influence of~$z_i$ any target example~$z_e$'s loss can be estimated via a sum of first-order approximations:
\begin{align}
\mathcal{I}_{TracInCP}(z_i, z_e) = \sum_{t \in \mathcal{T}} \eta_t \nabla \tilde{\ell}(z_i, \tilde{\theta}_{t}) \cdot \tilde{\ell}(z_e, \tilde{\theta}_{t}),
\end{align}
\revised{%
where ${\mathcal{T} \subseteq \{1, \ldots, T\}}$ are the iteration numbers corresponding to the ${\lvert \mathcal{T} \rvert}$~checkpoints and $\eta_t$~is the learning rate at iteration~$t$.
}

\revised{\emph{BoostIn} is our adaptation of TracIn to GBDTs.} While TracIn computes influence by summing over gradient updates, the analog in a GBDT is to sum over trees, where each tree represents a functional gradient update~(Eq.~\ref{eq:gbdt_weak_learner}). \jonathan{Made explicit the connection between gradient steps in deep learning and gradient updates in GBDTs.} BoostIn first considers all intermediate GBDT models constructed during training as checkpoints~($f_0, f_1, \ldots, f_T$). Recall the difference between any two intermediate models~$f_{t}$ and~$f_{t-1}$ is the weak learner~$m_t$ multiplied by~$\eta_t$~(the learning rate at iteration~$t$), and that each training example is visited exactly once between checkpoints since~$m_t$ computes the gradients of \emph{all} training examples using the predictions from the previous iteration~(Eq.~\ref{eq:gbdt_weak_learner}). \revised{In contrast to TracIn, in which a checkpoint typically consists of updates from minibatches that are partitions of the randomly ordered training data, each checkpoint during GBDT training uses all the data at once; thus, training order for GBDTs is irrelevant}.

BoostIn processes each intermediate model using Assumption~\ref{assumption:fixed_structure} to analyze the effect of training example~$z_i$ on the leaf it belongs to at each iteration. For the $t$th iteration, the approximate change in leaf value due to~$z_i$ corresponds to an estimated change in prediction of~$z_e$ only if~$z_i$ and~$z_e$ are in the same leaf at that iteration. The estimated change in prediction then approximates the change in loss on~$z_e$. Finally, BoostIn aggregates these approximations across all iterations, simulating the effect of~$z_i$ on~$z_e$ throughout the training process.
More formally, BoostIn uses the chain rule to analyze the marginal effect each changing leaf value has on the loss of~$z_e$, and sums these marginal effects across checkpoints:
\begin{align}
\mathcal{I}_{BoostIn}(z_i, z_e)
&= \sum_{t=1}^{T} \mathbbm{1}[R_t(x_i) = R_t(x_e)] \left( -\frac{d \ell(y_e, \hat{y}_e)}{d w_i}\Big|_{\hat{y}_e=f_{t}(x_e)} \right) \nonumber \\
&= \sum_{t=1}^{T} \mathbbm{1}[R_t(x_i) = R_t(x_e)] \left( -\fracpartial{ \ell(y_e, \hat{y}_e)}{\hat{y}_e}\Big|_{\hat{y}_e=f_{t}(x_e)} \cdot \fracpartial{f_t(x_e)}{w_i} \right) \nonumber \\
&= \sum_{t=1}^{T} \mathbbm{1}[R_t(x_i) = R_t(x_e)] \left( \fracpartial{\ell(y_e, \hat{y_e})}{\hat{y}_e}\Big|_{\hat{y}_e=f_{t}(x_e)} \cdot \eta_t \fracpartial{\theta_{t, l}}{w_i} \right)
\label{eq:boostin_influence}
\end{align}
in which~$\fracpartial{f_t(x_e)}{w_i} \approx -\eta_t \fracpartial{\theta_{t, l}}{w_i}$, $l=R_{t}(x_e)$, and the partial derivative of~$\theta_{t, l}$ with respect to~$w_i$ is defined as:
\begin{align}
\fracpartial{\theta_{t, l}}{w_i} = \frac{g_{t, i} + h_{t, i}\ \theta_{t, l}}{\sum_{j \in I_{t, l}} h_{t, j} + \lambda}.
\end{align}

Again, note~Eq.~\eqref{eq:boostin_influence} enforces a constraint that attributes nonzero influence only when~$z_i$ and~$z_e$ are in the same leaf at a given iteration. Also, when processing an intermediate model at iteration~$t$, BoostIn avoids the cascade effect by approximating the change in only the $R_t(x_i) = l$th leaf for weak learner~$m_t$.

\emph{How BoostIn Relates to LeafInfluence-SinglePoint.} The main idea of LeafInfSP~(Eq.~\ref{eq:leafinfsp_influence}) is to \emph{accumulate} the changes in the leaf values affected by upweighting~$z_i$
and multiply this result by the \emph{final} prediction of the GBDT model on~$x_e$. In contrast, BoostIn~(Eq.~\ref{eq:boostin_influence}) multiplies each leaf-value derivative by the corresponding \emph{intermediate} prediction of~$x_e$ at each iteration~(only relevant when~$z_i$ and~$z_e$ are in the same leaf at a given iteration~$t$), then takes the sum over all intermediate results. This is an important distinction as BoostIn analyzes the cumulative change in loss \emph{throughout} the training process, not just on the final model prediction.

In terms of computation, BoostIn and LeafInfSP have the same runtime complexity for computing the influence of a \emph{single} training example~$O(T)$, and computing influence values for \emph{all} training examples~$O(Tn)$. Empirically, however, the original implementation of LeafInfSP\footnote{\url{https://github.com/bsharchilev/influence_boosting}} is not optimized to realize this lower runtime complexity as it is implemented in conjunction with the full LeafInfluence approach; thus, as a minor contribution, we implement an optimized version of LeafInfSP and demonstrate in~\S\ref{sec:runtime} that our implementation achieves similar runtime performance compared to BoostIn, as expected. Overall, BoostIn is a solid choice for influence estimation in GBDTs, providing  on par or better influence estimates than LeafInfSP and LeafInfluence while being orders of magnitude more efficient than LeafInfluence.

\subsection{TREX: Adapting Representer-Point Selection}
\label{sec:trex}

Representer theorems~\citep{scholkopf2001generalized} state the optimal solutions of many learning problems can be represented in terms of the training examples. In particular, the nonparametric representer theorem~\citep[Theorem 4]{scholkopf2001generalized} applies to empirical risk minimization within a reproducing kernel Hilbert space (RKHS); this theorem covers a wide range of linear and kernelized machine-learning methods.

\cite{yeh2018representer} apply representer theorems to deep learning models by using the layers~(except the final layer) of the network~$\tilde{\theta}_1$ as a feature map~$\revised{\tilde{\textbf{f}}}_i = \tilde{\phi}(x_i; \tilde{\theta}_1)$, and training a kernelized model with L2 regularization on the transformed features. Specifically, for surrogate \revised{linear} model~$f^*$, parameters~$\psi^*$ represent the regularized, empirical risk minimizer:
\begin{align}
\psi^* = \argmin_{\psi} \lambda ||\psi||^2 + \frac{1}{n} \sum_{i=1}^{n} \revised{\tilde{\ell}}(y_i, f^*(\revised{\tilde{\textbf{f}}}_i; \psi)).
\label{eq:rep_optimization}
\end{align}
Since $\psi^{*}$ is a stationary point, the gradient of the loss is zero:
\begin{align}
\frac{1}{n} \sum_{i=1}^{n} \fracpartial{\revised{\tilde{\ell}}(y_i, f^*(\revised{\tilde{\textbf{f}}}_i; \psi^*))}{\psi^*} + 2 \lambda \psi^* = 0, \\
\therefore \psi^* = -\frac{1}{2 \lambda n} \sum_{i=1}^{n} \fracpartial{\revised{\tilde{\ell}}(y_i, f^*(\revised{\tilde{\textbf{f}}}_i; \psi^*))}{\psi^*} = \sum_{i=1}^{n} \alpha_i \revised{\tilde{\textbf{f}}}_i,
\end{align}
in which~$\alpha_i = -\frac{1}{2 \lambda n} \fracpartial{\revised{\tilde{\ell}}(y_i, \hat{y}_i)}{\hat{y}_i} \big|_{\hat{y}_i=f^*(\revised{\tilde{\textbf{f}}}_i; \psi^*)}$ is the global importance of~$z_i$ to the overall surrogate model. Finally, the pre-activation prediction of an arbitrary target example~$z_e$ can be decomposed as a linear combination of the training examples:
\begin{align}
f^*(\revised{\tilde{\textbf{f}}}_e; \psi^*) = \sum_{i=1}^{n} \alpha_i \revised{\langle \tilde{\textbf{f}}_i, \tilde{\textbf{f}}_e \rangle},
\label{eq:rep_decompose}
\end{align}
in which~\revised{$\langle \tilde{\textbf{f}}_i, \tilde{\textbf{f}}_e \rangle$} represents the similarity between~$z_i$ and~$z_e$ in the transformed feature space;~$\alpha_i \revised{\langle \tilde{\textbf{f}}_i, \tilde{\textbf{f}}_e \rangle}$ is referred to as the \emph{representer value} for~$z_i$ given~$z_e$, and its magnitude is largest when the magnitudes of \emph{both} the training example weight~$\alpha_i$ and the similarity of~$z_i$ to~$z_e$ is large. The representer value of~$z_i$ can be positive or negative, representing the attribution of~$z_i$ towards or away from the predicted value of~$z_e$, respectively.

To adapt representer-point methods to GBDTs, we need a way of extracting the learned representation of a given GBDT model, similar to feature extraction in deep learning models. For this purpose, we use \emph{supervised tree kernels}~\citep{davies2014random,he2014practical,bloniarz2016supervised}, a general approach for extracting the learned representation from a tree ensemble by using the structure of the trees. 
\revised{
Tree kernels can be used to measure the similarity between two instances by analyzing how each example is processed by each tree in the ensemble~\citep{moosmann2006fast}. Tree kernels  can outperform traditional nearest-neighbor methods at identifying the training instances most relevant to a prediction -- in particular for large data dimensions.
}

Intuitively, two examples are predicted identically if they appear in the same leaf in every tree in the ensemble. Thus, we can define the degree of similarity between two data points by comparing the specific paths taken through each tree; we can incorporate additional flexibility into the similarity measure accounting for path overlap, node weights~(number of examples at a node), and leaf values.
More formally, we define tree kernels as dot products in an alternate feature representation defined by the feature mapping~$\phi$, i.e., {$\langle \textbf{f}_i, \textbf{f}_e \rangle = \langle \phi(x_i; f), \phi(x_e; f) \rangle$}. Note these supervised kernels are parameterized by the GBDT model~$f$, since the computation necessarily depends on the structure of the ensemble, similar to taking the output of the penultimate layer in deep learning models. Based on work by~\cite{plumb2018model} on local linear modeling, we define~{\small $\phi(x, f) = \revised{\bigcup}_{t=1}^{T} (\frac{1}{n_{t,l}} \cdot \mathbbm{1}[x \in r_{t, l}])_{l=1}^{M_t}$} as a sparse vector over all leaves in~$f$; i.e., for each tree, the element at $R_t(x) = l$th leaf is nonzero and inversely weighted by~$n_{t,l}$, the number of training examples assigned to the $l$th leaf.
\jonathan{Added description of WeightedLeafPath tree kernel here.} 

Given feature mapping~$\phi$, we use Eq.~\eqref{eq:rep_optimization} with \revised{loss function~$\ell$} to train a surrogate model to approximate our original GBDT model, enabling the decomposition of a target prediction using Eq.~\eqref{eq:rep_decompose}.
However, the resulting representer values only quantify the contribution of a training example~$z_i$ to the \emph{prediction} of~$z_e$, but we are interested in how~$z_i$ affects the loss of~$z_e$; thus, we measure the influence of~$z_i$ on the loss of~$z_e$ by subtracting the representer value for~$z_i$ from the prediction decomposition of~$z_e$ and computing the change in loss:
\begin{align}
    \mathcal{I}_{TREX}(z_i, z_e) = \ell\bigg(y_e, \upsilon \big( \hat{y}_e^*
    - \alpha_i \revised{\langle \textbf{f}_i, \textbf{f}_e \rangle} \big) \bigg)
    - \ell \big(y_e, \upsilon(\hat{y}_e^*) \big),
    \label{eq:trex_influence}
\end{align}
in which~$\upsilon$ is an activation function\footnote{Typically a sigmoid or softmax for binary and multiclass classification tasks, respectively.} and $\hat{y}_e^*=\sum_{j=1}^{n} \alpha_j \, \revised{\langle \textbf{f}_j, \textbf{f}_e \rangle}$ is the surrogate model pre-activation prediction for~$z_e$. We denote this method~\emph{TREX}~(\underline{\textbf{T}}ree-ensemble \underline{\textbf{R}}epresenter \underline{\textbf{Ex}}amples) and evaluate its influence-estimation quality in~\S\ref{sec:result_summary}.

\emph{Similarity-Based Influence.}
As defined above, TREX measures the influence of an example~$z_i$ on a target example~$z_e$ using two pieces of information: the weight of the training example~$\alpha_i$, and the similarity to the target instance~$\revised{\langle \textbf{f}_i, \textbf{f}_e \rangle}$. To better understand the marginal effect of each component, we define an additional influence estimation method that skips training a surrogate model and \emph{only} uses the tree-kernel similarity to quantify the influence of~$z_i$ on~$z_e$:
\begin{align}
    \mathcal{I}_{TreeSim}(z_i, z_e)
    = \mathbbm{1}[y_i = y_e]\revised{\langle \textbf{f}_i, \textbf{f}_e \rangle}
    - \mathbbm{1}[y_i \not = y_e]\revised{\langle \textbf{f}_i, \textbf{f}_e \rangle}.
\label{eq:treesim_influence}
\end{align}
This method, \emph{TreeSim}, attributes positive influence to examples with the \emph{same} label as~$z_e$, and negative otherwise;\footnote{For regression, TreeSim treats~$z_i$ and~$z_e$ as having the same label if~$y_i$ and~$y_e$ are on the same side of the target prediction~$\hat{y}_e$; that is,~$\revised{\text{sgn}}(\hat{y}_e - y_i) = \revised{\text{sgn}}(\hat{y}_e - y_e)$, in which~$sgn$ is the signum function.} this value is then scaled by the similarity between~$z_i$ and~$z_e$.

\subsection{Summary of Influence-Estimation Methods}

\daniel{This paragraph is somewhat redundant, but I like providing a high-level overview here. So I don't know if it should be trimmed.}\jonathan{I think it should stay, but if you have suggestions for it being more concise, I am open to it.}
Table~\ref{tab:method_summary} summarizes the influence-estimation methods discussed in this paper. In  short, LOO and SubSample are model-agnostic approaches that compute the influence of training examples by removing them and retraining one or multiple models on revised data sets. The rest of the methods are model specific and assume a fixed-structure (Assumption~\ref{assumption:fixed_structure}) while computing influence values. LeafRefit recomputes all leaf values without a particular training instance in order to estimate the influence of that instance. LeafInfluence and LeafInfSP approximate this process by measuring the total aggregated change in model parameters and analyzing how this change affects the final target prediction. BoostIn approximates the change in leaf values and their effects on the loss of a target prediction at each intermediate model, tracing the influence of a training instance on a target prediction throughout the training process. Both BoostIn and LeafInfSP are asymptotically much more tractable than LeafRefit and LeafInfluence. TREX trains an interpretable kernel surrogate model that decomposes any prediction into a sum of the training instances, and TreeSim identifies the most influential training examples by how similar they are to the target example.

\daniel{Can we also say something about speed in the table?}\jonathan{Perhaps, although there is hardly any space left in the table.}
\begin{table*}[h]
\newcommand{\header}[1]{\multirow{2}{*}{\textbf{#1}}}
\centering
\begin{tabular}{llcc}
\toprule
\header{Method} & \header{Source} & \header{Changes} & \textbf{Assumes} \\
& & & \textbf{Fixed Struct.} \\
\midrule
LOO            & -                            & -                            &            \\
SubSample      & \cite{feldman2020neural}     & -                            &            \\
LeafRefit      & \cite{sharchilev2018finding} & -                            & \checkmark \\
LeafInfluence  & \cite{koh2017understanding}  & \cite{sharchilev2018finding} & \checkmark \\
LeafInfSP      & \cite{koh2017understanding}  & \cite{sharchilev2018finding} & \checkmark \\
BoostIn        & \cite{pruthi2020estimating}  & \S\ref{sec:boostin} \revised{\tiny{(Ours)}}   & \checkmark \\
TREX           & \cite{yeh2018representer}    & \S\ref{sec:trex} \revised{\tiny{(Ours)}}      & \checkmark \\
TreeSim        & \cite{plumb2018model}        & \S\ref{sec:trex} \revised{\tiny{(Ours)}}      & \checkmark \\
\bottomrule
\end{tabular}
\caption{Summary of influence-estimation methods.}
\label{tab:method_summary}
\end{table*}

In the following sections, we evaluate the estimation quality of each method across a wide range of data sets and evaluation contexts.

\section{Methodology}
\label{sec:methodology}

In our experiments, we order the training data based on the influence values generated by each method for a \emph{single} test example~(\S\ref{sec:removal_single},~\S\ref{sec:targeted_edit_single}) or a \emph{set} of test examples~(\S\ref{sec:removal_set},~\S\ref{sec:adding_noise_set},~\S\ref{sec:noise_set}). We then evaluate these orderings in different contexts by removing~(\S\ref{sec:removal_single},~\S\ref{sec:removal_set}), performing targeted label editing~(\S\ref{sec:targeted_edit_single}), and adding label noise~(\S\ref{sec:adding_noise_set}) to the most influential examples and observing the effect on the model/predictions after \emph{retraining}; the more a method degrades the resulting model predictions,\footnote{We measure changes in model predictions via logistic loss for classification tasks and squared error for regression tasks.} the higher that method is ranked. We also evaluate the effectiveness of each method at detecting noisy or mislabelled training examples~(\S\ref{sec:noise_set}) as is often done in previous work~\citep{ghorbani2019data,pruthi2020estimating}. Overall, we want influence methods with high fidelity, that is, methods that accurately predict the \emph{behavior} of the model after some operation~(e.g., removing the most influential training examples for a test example should \emph{actually} increase the loss of the given test example after retraining without those` examples); in general, this evaluation protocol provides a quantitative measure of the effectiveness of each influence-estimation method. In the following subsections, we provide data set and method details, and then describe each experiment in detail; we present our results in~\S\ref{sec:result_summary}.

\subsection{Data Sets and Methods}

\emph{Data sets.} We evaluate on 22 real-world tabular data sets~(13~binary-classification tasks, 1~multiclass-classification task, and 8~regression tasks) well-suited for boosted tree-based models. For each data set, we generate one-hot encodings for any categorical attributes and leave all binary and numeric attributes as is. For any data set without a predefined train/test split, we sample 80\% of the data uniformly at random for training and use the rest for testing. All additional data set details are in the Appendix,~\S\ref{appendix_sec:datasets}.

\emph{Models.} We train GBDT models using the most popular and modern implementations:~LightGBM~\citep[LGB]{ke2017lightgbm}, XGBoost~\citep[XGB]{chen2016xgboost}, Scikit-Learn~\citep[SGB]{scikit-learn}, and CatBoost~\citep[CB]{prokhorenkova2018catboost}. Each model is tuned using 5-fold cross-validation; selected hyperparameters are in~\S\ref{appendix_sec:gbdt_predictive_performance}:~Table~\ref{tab:gbdt_hyperparams}, with predictive performance comparisons in~\S\ref{appendix_sec:gbdt_predictive_performance}:~Table~\ref{tab:predictive_performance}.

\emph{Influence Methods and Baselines.} We include the following influence-estimation methods in our evaluation: LOO~(Eq.~\ref{eq:loo_influence}), SubSample~(Eq.~\ref{eq:subsample_influence}; we set~$\tau=4,000$ and~$m= \lfloor 0.7n \rfloor$ as recommended by \citet{feldman2020neural}), LeafRefit, LeafInfluence~(Eq.~\ref{eq:leaf_influence}),
BoostIn~(Eq.~\ref{eq:boostin_influence}), LeafInfSP~(Eq.~\ref{eq:leafinfsp_influence}), TREX~(Eq.~\ref{eq:trex_influence}), and TreeSim~(Eq.~\ref{eq:treesim_influence}).
We also include~\emph{Random} as an additional baseline, which assigns influence values via a standard normal distribution:
$\mathcal{I}_{Random}(z_i, z_e) \sim \mathcal{N}(0, 1)$. Due to the limited scalability of LeafRefit and LeafInfluence~(see~\S\ref{sec:runtime}), we evaluate these methods on a subset of the data sets consisting of 13 smaller data sets and present results on this group, which we denote \emph{small data subset~(SDS)}~(exactly which data sets are part of SDS is given in~\S\ref{appendix_sec:datasets}:~Table~\ref{tab:datasets}); however, we observe the same trends when including all data sets in our analysis, which are in~\S\ref{app_sec:resul_summary_all_datasets}.

\emph{Implementation.} Code containing all influence-estimation implementations and experiments is available at~\url{https://github.com/jjbrophy47/tree\_influence}; our implementations also include an optimized version of LeafInfSP. Supplemental implementation details as well as hardware details used for the experiments are in~\S\ref{appendix_sec:experiment_details}.

\subsection{Single Test Instance: Removing Influential Training Examples}
\label{sec:removal_single}

Inspired by the remove and retrain~(ROAR) framework for measuring the impact of different features~\citep{hooker2019benchmark}, we evaluate the influence of training examples on a given test example by measuring the change in loss on the test example using a model retrained after removing the most influential training examples. In theory, the training examples with the most positive influence values decrease the loss of the test example the most; thus, removing them and retraining the model should increase the loss of the test example.

For this experiment, we generate influence values for a given test example~$z_e$, and order the training instances from most positive~(i.e.,~instances that decrease the loss of~$z_e$ the most) to most negative. Using this ordering, we remove 0.1\%, 0.5\%, 1\%, 1.5\%, and 2\% of the training data, retraining a separate GBDT model on each modified version of the data set. We then measure the change in loss on~$z_e$ using the original and retrained models. We repeat this experiment for 100 randomly chosen test examples and compute the average increase in test loss per example. We then rank the influence-estimation methods by how much they increase the test loss on average; then, we average these rankings over removal percentages~(0.1\%, 0.5\%, etc.), GBDT types, and data sets.

\subsection{Single Test Instance: Targeted Training-Label Edits}
\label{sec:targeted_edit_single}

In this section, instead of removing examples, we ask the counterfactual~\citep{byrne2019counterfactuals,keane2020good}, ``how would this prediction change if I were to edit the \emph{label(s)} of the most influential training examples?'' 
\daniel{Next sentence is mostly filler?}\jonathan{Trimmed.}
However, since most of the influence-estimation methods simulate the removal of a training example, we slightly adapt the estimators to simulate a changing training label for this experiment. We modify LOO to compute the influence of~$z_i$ on~$z_e$ via training-label edit by changing~$z_i$~to~$z_i^*$~($z_i$ with a new label~$y_i^*$) and retraining the model. We similarly modify LeafRefit~(LOO with a fixed structure) to refit leaf values with~$z_i^*$ instead~$z_i$. Note these operations are equivalent to \emph{removing}~$z_i$ and \emph{adding}~$z_i^*$. However, LeafInfluence, LeafInfSP, and BoostIn only \emph{approximate} the removal of~$z_i$, but these influence methods are able to estimate the influence of an instance that does not actually exist in the training data~(see influence definitions in~\S\ref{sec:adapting_influence_methods}).
Thus, we can approximate the influence of a changing training label by simulating the removal of~$z_i$ and the addition of~$z_i^*$; that is,~$\mathcal{I}(z_i \rightarrow z_i^*, z_e) \approx \mathcal{I}(z_i, z_e) - \mathcal{I}(z_i^*, z_e)$.

In this experiment, we use the same setup as~\S\ref{sec:removal_single}, ordering the training examples from most positive to most negative. Using this ordering, we change 0.1\%, 0.5\%, 1\%, 1.5\%, and 2\% of the training labels to the \emph{same} chosen target label~$y^*$.\footnote{For binary-classification tasks, we choose $y^*$ to be opposite~$\hat{y}_e$~(the predicted label of $z_e$). For multiclass classification, $y^*$ is randomly sampled from~$\mathcal{Y} \setminus \revised{\{\hat{y}_e\}}$. For regression, $y^* = \Bar{\textbf{y}} - (\Bar{\textbf{y}} / 2)$ if $\hat{y}_e > \Bar{\textbf{y}}$, otherwise $y^* = \Bar{\textbf{y}} + (\Bar{\textbf{y}} / 2)$.} We retrain the model after each batch modification, and measure the change in loss on the test example; we then rank each method by how much the loss increases, and average these results over modification percentages, GBDT types, and data sets.

\subsection{Multiple Test Instances: Removing Influential Training Examples}
\label{sec:removal_set}

We now analyze the effect of influential examples on a \emph{set} of test instances. We sample 10\% of the test examples uniformly at random to use as a validation set and generate influence values for each example. We then aggregate the influence values via a sum over the validation examples to get a single influence value per training example, and then rank the training examples from most positive~(decreases the loss of the validation examples the most on aggregate) to most negative. Using this ordering, we remove examples in batches of 5\%, removing up to a maximum 50\% of the training data. We retrain a separate GBDT model after each batch removal and measure the change in loss to the original model on a held-out test set~(that is,~the test examples not used for the validation set). We then rank each influence-estimation method by how well it degrades the resulting model at each level of removal~(5\%, 10\%, etc.), and average these rankings over removal percentages, GBDT types, and data sets.

\subsection{Multiple Test Instances: Adding Training-Label Noise}
\label{sec:adding_noise_set}

For this experiment, we use the same setup as~\S\ref{sec:removal_set}, including the orderings produced by each influence-estimation method. Then, instead of removing the most positively-influential examples, we add noise to them by randomly changing each of their labels:~we flip labels for binary-classification tasks, sample new labels uniformly at random for multiclass-classification tasks, and sample new target values between the minimum and maximum values of~\textbf{y} uniformly at random for regression tasks. This procedure can be viewed as an availability data poisoning attack~\citep{steinhardt2017certified,levine2020dpa,hammoudeh2023certifiedregression}. We then retrain, remeasure, and rank the influence methods in the same way as~\S\ref{sec:removal_set}.

\subsection{Multiple Test Instances: Fixing Mislabelled Training Examples}
\label{sec:noise_set}

Another way of evaluating the influence of training examples is via detecting and fixing noisy or mislabelled training instances. Intuitively, very \emph{negatively-influential} training examples may signify outliers, mislabelled/noisy examples, etc., and possibly warrant manual inspection. We conduct this experiment in a similar fashion to those in the literature~\citep{koh2017understanding,yeh2018representer,ghorbani2019data,pruthi2020estimating,pleiss2020aum}, sampling 40\% of the training data uniformly at random and flipping their labels in the same way as~\S\ref{sec:adding_noise_set}.
We then generate influence values in the same way as~\S\ref{sec:removal_set}~and~\S\ref{sec:adding_noise_set}, but then rank the training examples from most negative~(that is,~examples that increase the loss of the validation examples the most on aggregate) to most positive. Using this ordering, we manually inspect~5\%, 10\%, 15\%, 20\%, 25\%, and 30\% of the training data, fixing the training examples whose labels had been flipped. We rank each method by how many mislabelled examples it detects at each level of manual inspection~(5\%, 10\%, etc.); we then average these rankings over all inspection levels, GBDT types, and data sets. We also add a simple but effective baseline called~\emph{Loss} that orders the training examples to be checked based on their loss~(high-loss training examples are checked first).

\section{Results and Analyses}
\label{sec:result_summary}

\begin{figure*}[t]
\begin{subfigure}{\textwidth}
    \centering
    \includegraphics[width=\textwidth,clip,trim=0 0 0 0]{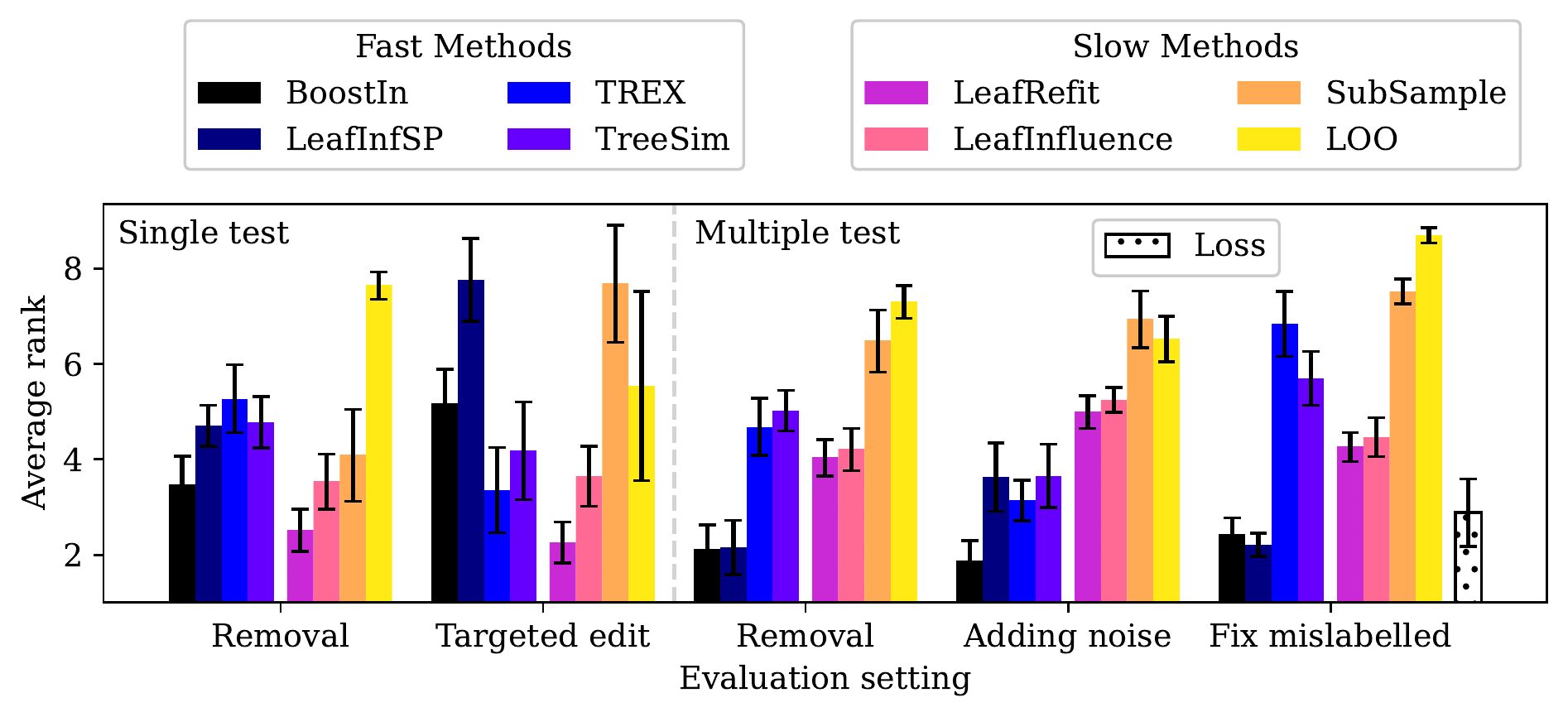}
    \caption{Average rank of each method. For each evaluation setting, results are averaged over all checkpoints, tree types, and data sets; error bars represent 95\% confidence intervals and are computed over data sets. Lower is better. We exclude Random since it performed consistently worse than all other methods in each setting.}
    \vskip 2.75mm
    \label{fig:rank_loss_li}
\end{subfigure}
\begin{subfigure}{\textwidth}
    \centering
    \includegraphics[width=\textwidth,clip,trim=0 0 0 0]{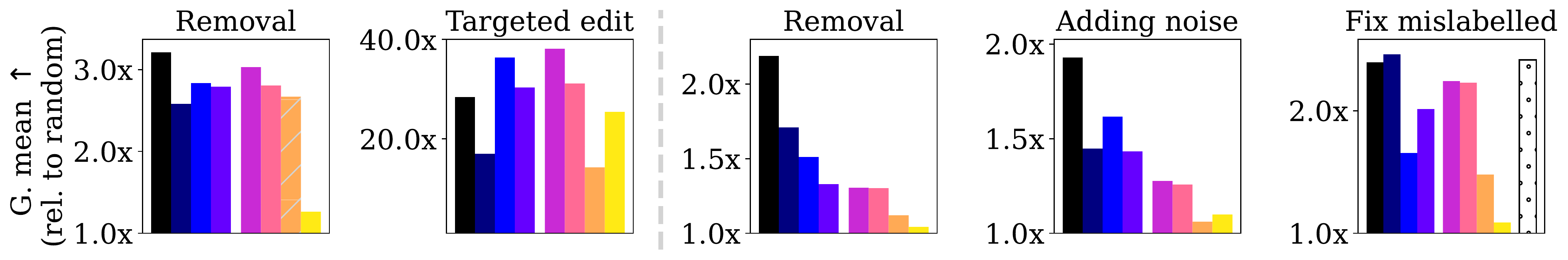}
    \caption{Average \emph{loss} increase~(except for ``fix mislabelled'', which shows average increase in \emph{mislabelled detection}) relative to random. For each evaluation setting, results are averaged over all checkpoints and tree types, then the geometric mean is computed over all data sets. Higher is better.}
    \label{fig:magnitude_loss_li}
\end{subfigure}
\caption{High-level overview of results showing~(a)~average rank and~(b)~relative impact of each method. Methods are grouped based on their relative efficiency~(Figure~\ref{fig:runtime}); for both subfigures, evaluation settings left of the gray-dashed line represent experiments that compute influence values for a \emph{single} test instance and measure the predictive impact on that instance (this is then repeated and averaged over 100 randomly-chosen test instances), while experiments right of the dashed line compute aggregate influence values for a \emph{set} of test examples and measure the predictive impact on a held-out test set.}
\label{fig:result_summary}
\end{figure*}

Figure~\ref{fig:result_summary} shows a high-level overview of our results across evaluation settings; we partition the influence techniques into ``fast'' and ``slow'' methods~(see~\S\ref{sec:runtime} for a runtime comparison) to give readers a sense of how much performance can be increased~(if any) given more computational effort.
Overall, BoostIn tends to perform best or equally best, except for the ``targeted edit'' experiment, in which LeafRefit and methods based on tree-kernel similarity such as TREX and TreeSim are more effective.
Surprisingly, LOO performs consistently poorly; thus, we investigate this method further in~\S\ref{sec:loo_fragility}.

\subsection{Summary of Results}

We present an overview of the results for each experiment in this section, with additional analyses in the Appendix,~\S\ref{app_sec:single_test_removal}-\S\ref{app_sec:multi_test_fix_noise}.

\daniel{Consider making these paragraphs instead of subsections.}\jonathan{Done.}

\emph{Single Test Instance: Removing Influential Training Examples.}
Figure~\ref{fig:rank_loss_li}~(left) shows LeafRefit ranking highest, with BoostIn and LeafInfluence performing roughly the same. SubSample also performs well, although we observe a noticeable drop in its relative performance when adding larger data sets into our analysis~(\S\ref{app_sec:resul_summary_all_datasets}); increasing~$\tau$ for larger data sets may improve performance, but may also significantly increase the running time. In terms of magnitude, all methods outperform random removal, with BoostIn having a slight advantage over all other methods~(Figure~\ref{fig:magnitude_loss_li}:~left); surprisingly, LOO does only marginally better than random.

\emph{Single Test Instance: Targeted Training-Label Edits.}
Figure~\ref{fig:rank_loss_li}~(middle-left) shows LeafRefit ranking higher than all other methods; LeafInfluence, TREX, and TreeSim also rank highly. Although BoostIn is not ranked as highly as these methods, it is significantly more effective than LeafInfSP and SubSample, and its relative magnitude in terms of loss increase is similar to that of LeafInfluence and TreeSim~(Figure~\ref{fig:magnitude_loss_li}:~middle-left).

\emph{Multiple Test Instances: Removing Influential Training Examples.}
Figure~\ref{fig:rank_loss_li}~(middle) shows BoostIn and LeafInfSP ranking significantly higher than all other methods; however, BoostIn tends to choose examples that increase the loss more than LeafInfSP, on average~(Figure~\ref{fig:magnitude_loss_li}-middle shows a 2.2x and 1.7x increase in loss relative to Random for BoostIn and LeafInfSP, respectively). Additional analyses for other predictive performance metrics such as accuracy are in \S\ref{app_sec:multi_test_remove}.

\emph{Multiple Test Instances: Adding Training-Label Noise.}
Our results show BoostIn clearly outperforms all other methods in terms of rank~(Figure~\ref{fig:rank_loss_li}:~middle-right) and relative loss increase~(Figure~\ref{fig:magnitude_loss_li}:~middle-right), and suggest BoostIn may be an effective tool for providing untargeted data set poisoning attacks. Somewhat surprisingly, LeafInfSP does not perform as well on this task as compared to removing examples. Additional analyses for other predictive performance metrics are in \S\ref{app_sec:multi_test_add_noise}.

\emph{Multiple Test Instances: Fixing Mislabelled Training Examples.}
Figure~\ref{fig:rank_loss_li}~(right) shows BoostIn and LeafInfSP performing best, slightly outranking Loss; however, all three methods perform comparably in terms of relative magnitude, on average~(Figure~\ref{fig:magnitude_loss_li}:~right). Predictive-performance improvements on the held-out test set after retraining the model on partially fixed versions of the data sets are in~\S\ref{app_sec:multi_test_fix_noise}.

\subsection{Runtime Comparison}
\label{sec:runtime}

To better understand the relative efficiency of each method, we measure the time it takes to compute all influences values for a single random test example for each dataset.\footnote{For a clearer comparison, no parallelization is used for any method.} We repeat each experiment 5 times, and average the results over GBDT types.

\begin{figure}[ht]
\centering
\includegraphics[width=1.0\textwidth]{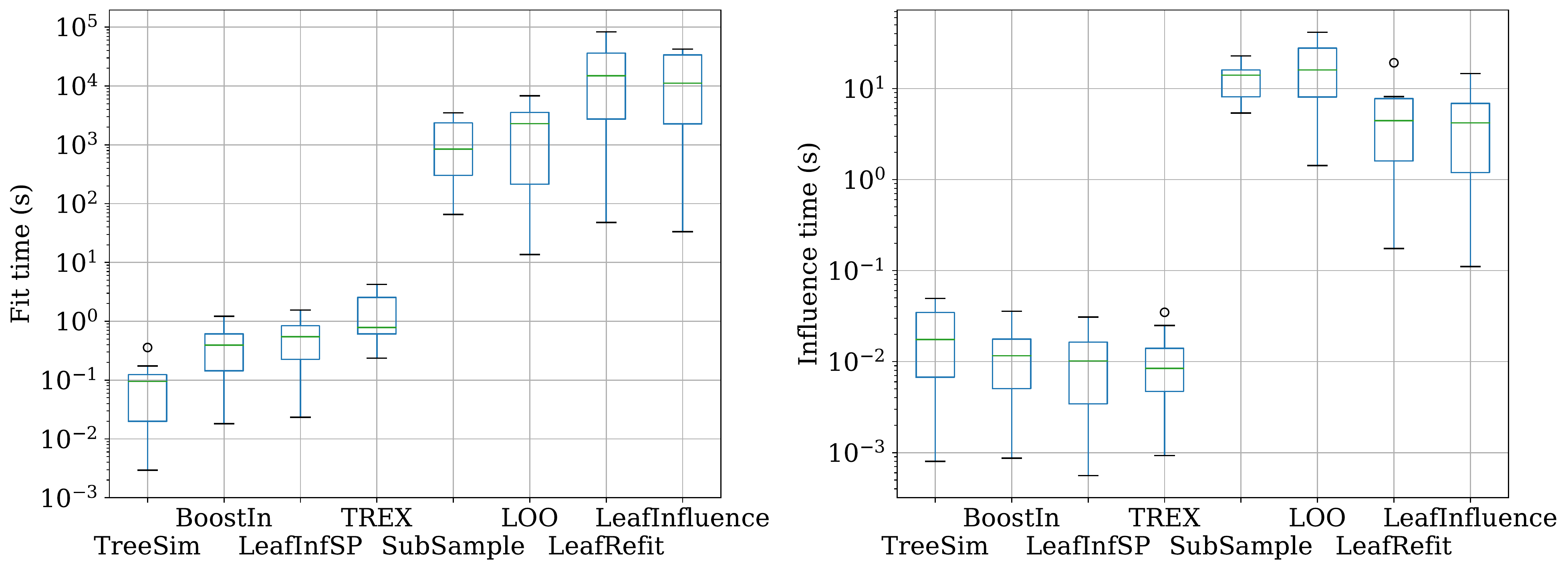}
\caption{\emph{Left}: Average setup time for each explainer. \emph{Right}: Average time to compute influence values of all training examples for one test example. Results are averaged over 5 folds and GBDT types; each box plot represents average running times across all SDS datasets. TreeSim, BoostIn, LeafInfSP, and TREX represent ``fast'' methods with low setup and influence times, and are separated from the remaining ``slow'' methods by orders of magnitude efficiency.} 
\label{fig:runtime}
\end{figure}

Figure~\ref{fig:runtime} shows the runtime of each approach broken down into two components: ``fit time'' and ``influence time''. Fit time is the time to initialize and set up the explainer, and influence time is the time to compute the influence of all training examples for one test example; note the log scale. TreeSim has the fastest setup time overall; however, TreeSim, BoostIn, LeafInfSP, and TREX all have low initialization and influence times compared to SubSample, LOO, LeafRefit, and LeafInfluence. We semantically group the former and latter methods into ``fast'' and ``slow'' groups, separated by orders of magnitude efficiency.

All methods in the ``slow'' group must train or approximate~(and store) a separate model for each training example during setup, which becomes unwieldy as~$n$ increases~(SubSample is the exception that trains a fixed~$\tau=4000$ models; however, this is still a significant number of models to train and store). When computing influence values, the ``slow'' methods predict using all models obtained during setup. In contrast, methods in the ``fast'' group are able to compute influence values using only one model instead of~$n$ or~$\tau$ models.

Surprisingly, LeafInfluence and LeafRefit take significantly more time to set up than than LOO and have similar influence times, on average. Additionally, the relative efficiency of SubSample is very similar to LOO; this is mainly due to our choice of~$\tau$ and the small data set sizes in the SDS. For larger data sets~(assuming~$\tau$ remains fixed) or smaller choices of~$\tau$, one can expect SubSample to be more efficient than LOO in general.

\subsection{Correlation Between Influence Methods}
\label{sec:correlation}

To better understand the relationships between different influence methods, we compute the Spearman rank~\citep{zar2005spearman} and Pearson correlation coefficients between every pair of influence methods using the values generated for each test example in~\S\ref{sec:removal_single}. We then average these correlations over all test examples, GBDT types, and data sets.

\begin{figure}[ht]
\centering
\begin{subfigure}{0.5\textwidth}
  \centering
  \includegraphics[width=1.0\linewidth]{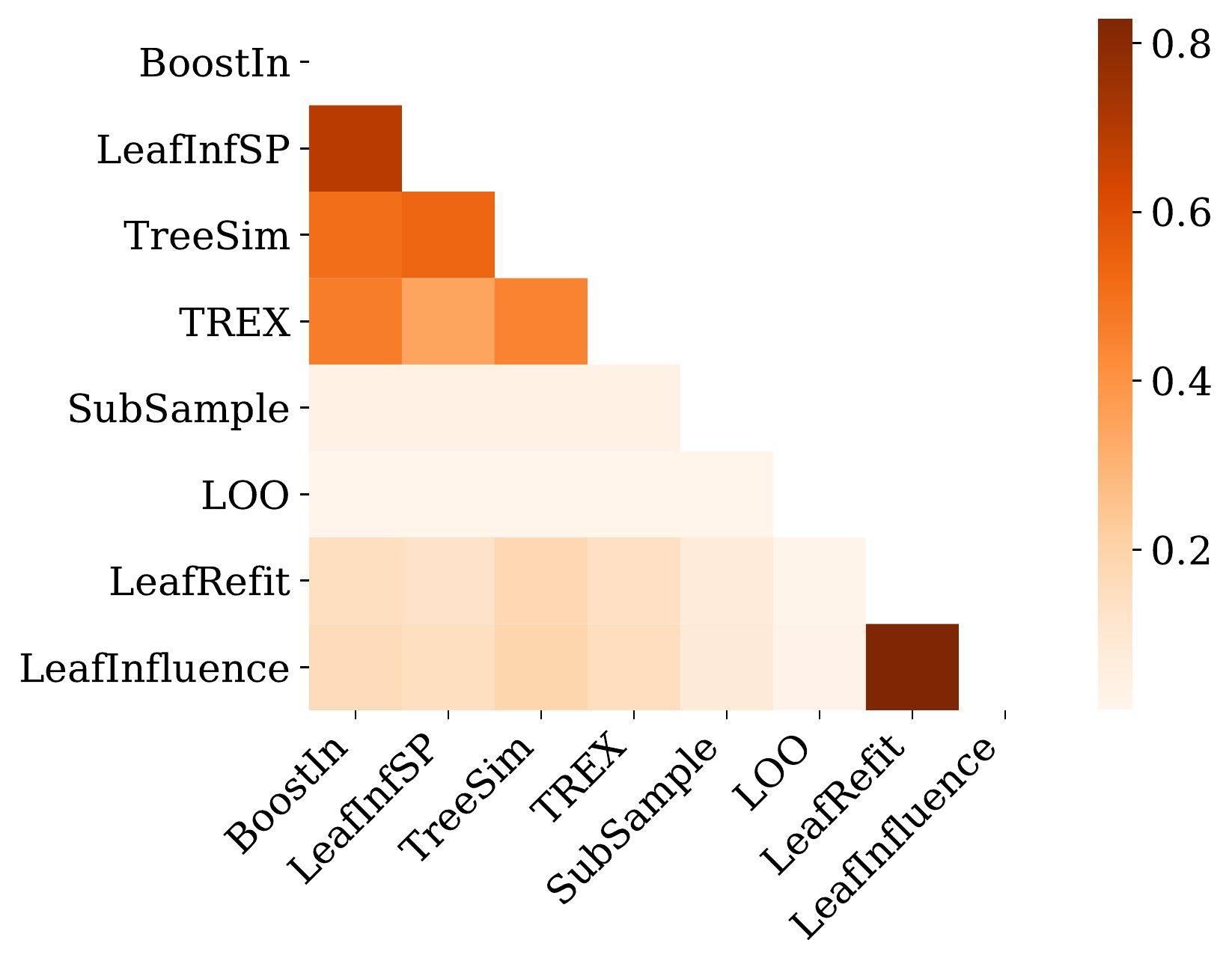}
  \caption{Spearman}
  \label{fig:sub1}
\end{subfigure}%
\begin{subfigure}{0.5\textwidth}
  \centering
  \includegraphics[width=1.0\linewidth]{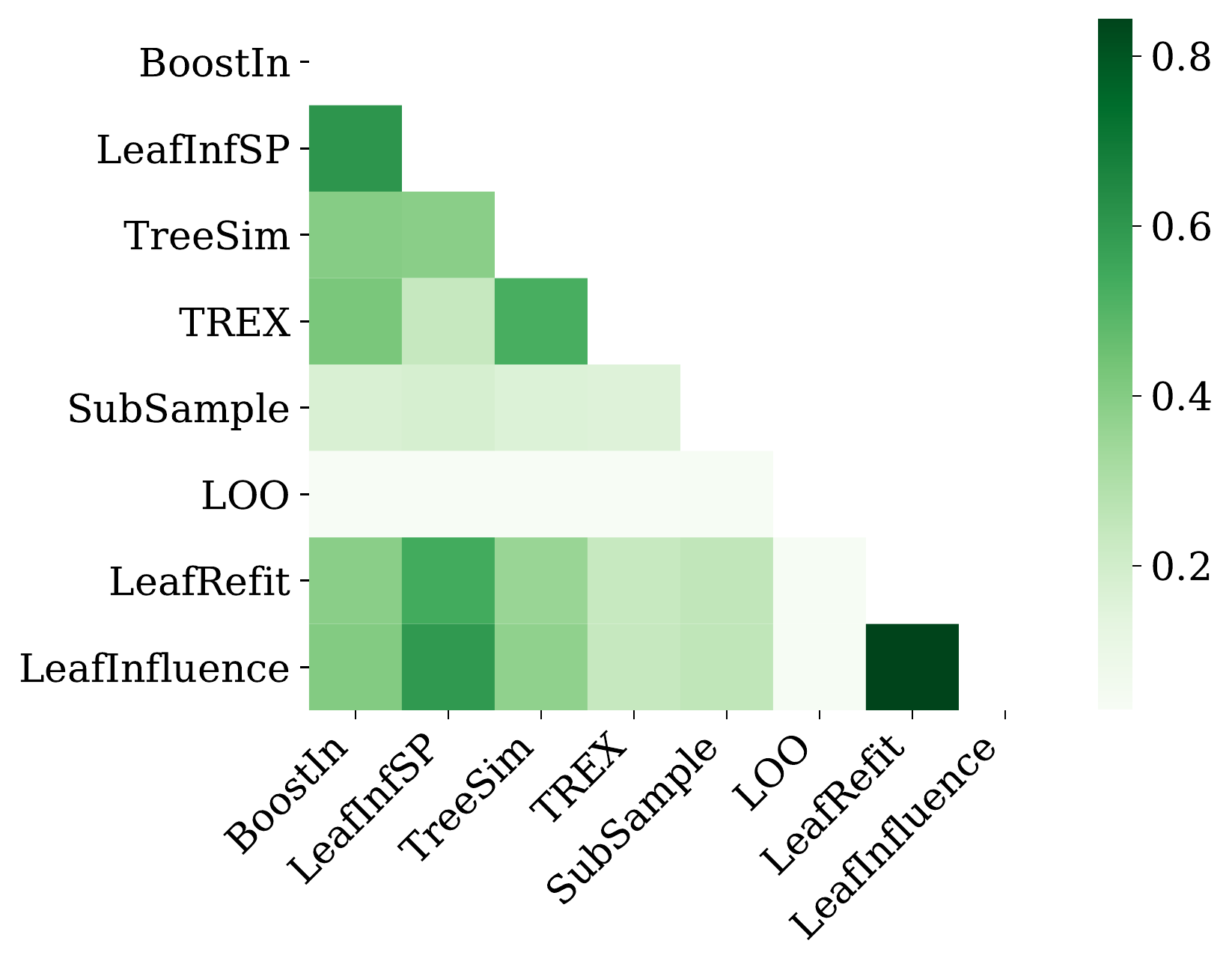}
  \caption{Pearson}
  \label{fig:sub2}
\end{subfigure}
\caption{Average Spearman and Pearson correlation coefficients between every pair of influence methods; results are based on the rankings generated via the influence values for each test example, averaged over 100 test examples and then over tree types and data sets.}
\label{fig:spearman}
\end{figure}

Figure~\ref{fig:spearman} shows a high correlation between LeafRefit and LeafInfluence, which is expected since LeafInfluence is an approximation of LeafRefit. BoostIn is highly correlated with LeafInfSP, which we theoretically analyzed as relatively similar in~\S\ref{sec:boostin}. We also observe a cluster of similar methods: BoostIn, LeafInfSP, TreeSim, and TREX. Surprisingly, LOO is not highly correlated with any other method; the low correlation and relatively poor performance~(Figure~\ref{fig:result_summary}) prompts us to investigate LOO further in~\S\ref{sec:loo_fragility}. This result also provides evidence that the previous state-of-the-art, LeafInfluence, is a poor approximation of LOO. Additional analysis regarding the correlation between influence methods is in~\S\ref{app_sec:correlation}.

\subsection{The Structural Fragility of LOO}
\label{sec:loo_fragility}

Throughout our experiments, LOO performs consistently worse than many of the other methods, often only doing marginally better than random. These results thus warrant further investigation.

\begin{figure}[ht]
\centering
\begin{subfigure}{0.5\textwidth}
  \centering
  \includegraphics[width=1.0\linewidth]{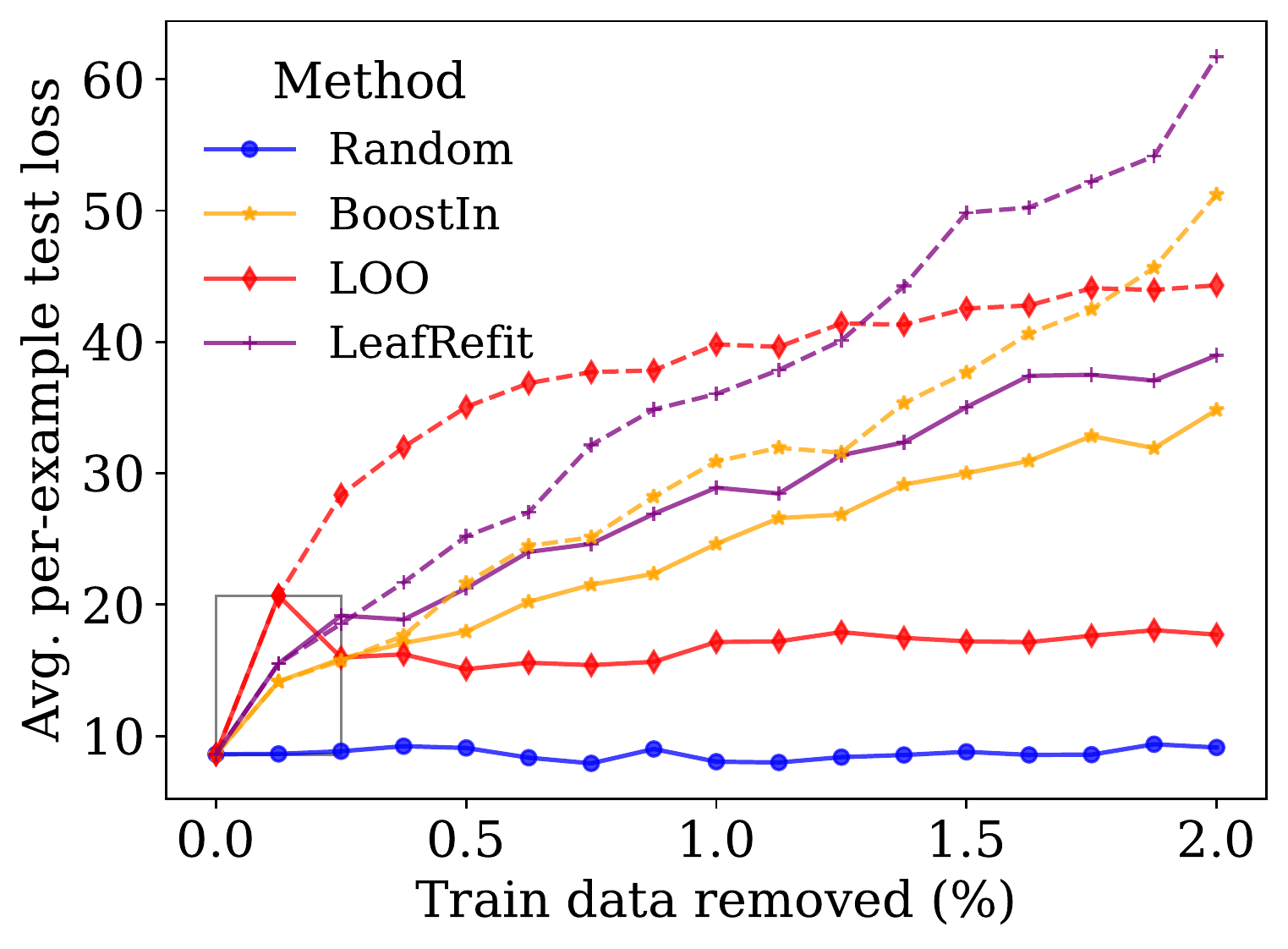}
  \caption{Concrete data set}
  \label{fig:concrete_reinf}
\end{subfigure}%
\begin{subfigure}{0.5\textwidth}
  \centering
  \includegraphics[width=1.0\linewidth]{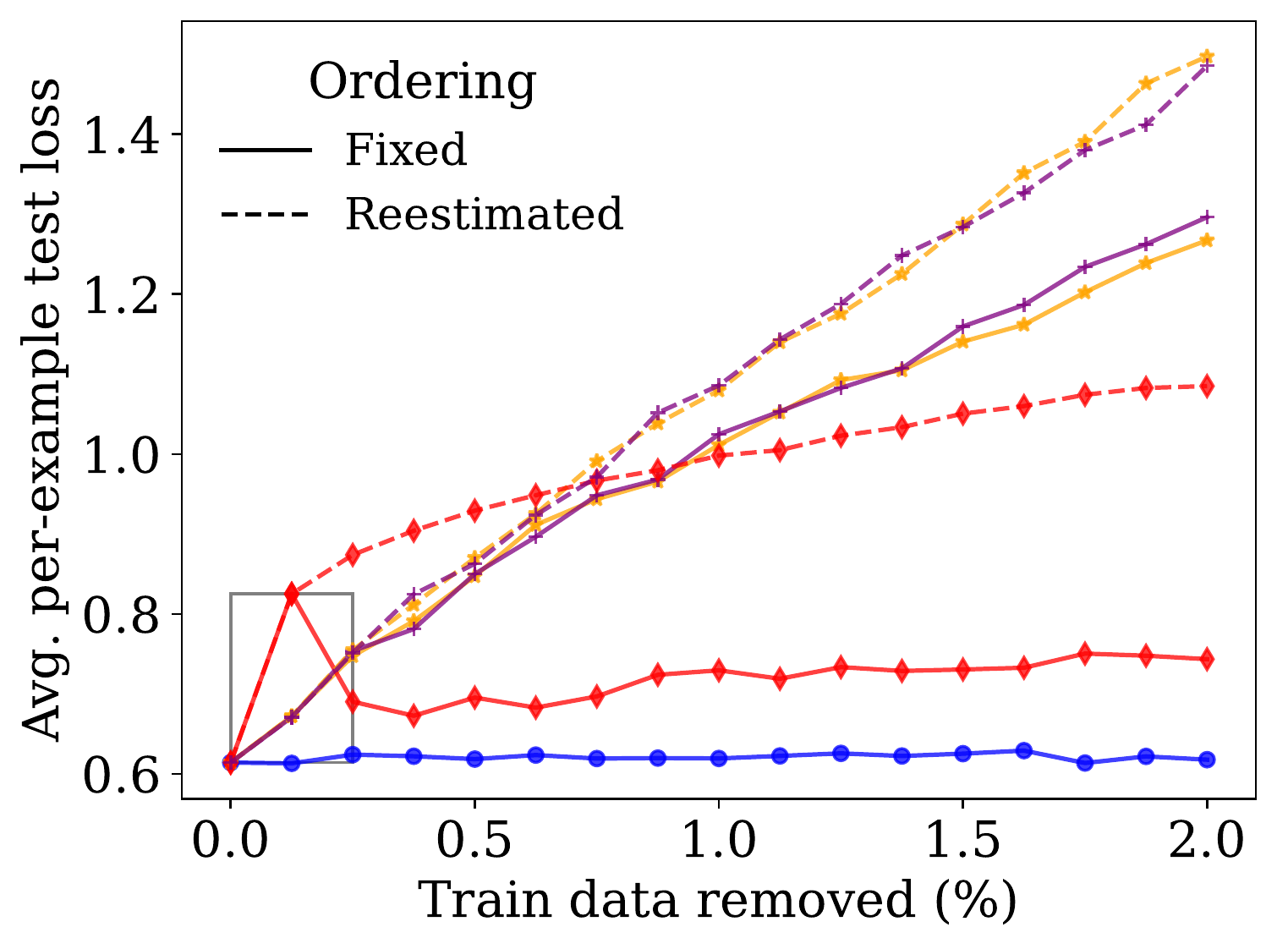}
  \caption{German Credit data set}
  \label{fig:german_credit_reinf}
\end{subfigure}
\caption{Change in test-example loss~(averaged over 100 test examples using LGB) after removing the most positively-influential training examples \emph{one at a time} using a fixed ordering as well as a dynamic ordering that \emph{reestimates} influence values for the remaining training data after each removal. The gray box highlights the large increase in test loss by LOO after removing only a single example. Additional examples are in the Appendix,~\S\ref{app_sec:loo_structural_fragility}.}
\label{fig:reinfluence}
\end{figure}

To assess the performance of LOO in more depth, we use the same setup as~\S\ref{sec:removal_single}, except we remove examples in increments of \emph{one} instead of specified percentages~(0.1\%, 0.5\%, etc.); we do the same for Random, BoostIn, and LeafRefit for additional context. Figure~\ref{fig:reinfluence} shows the results, and we immediately notice a spike in test loss~(outlined by a gray box) after the \emph{first} removal for LOO higher than any of the other methods, followed by a drop and general plateau of the test loss for subsequent removals. We consistently see this spike across data sets and GBDT types~(additional examples are in~\S\ref{app_sec:loo_structural_fragility}). 
\revised{This shows that LOO's top-ranked example is indeed the most influential, but using LOO's ordering to pick a \emph{group} of examples to remove is much less effective. In fact, on average, removing the top two examples (as ranked by LOO) has \emph{less} impact on test loss than just removing one! In other words, \emph{the fixed ordering produced by LOO means that highly-ranked examples can counteract the effects of other highly-ranked examples when removed}.}

To quantify how much this \revised{fixed} ordering makes a difference, we \emph{reestimate} influence values on the remaining training data after each deletion, dynamically reordering the training examples to be removed~(Figure~\ref{fig:reinfluence}). We make two observations from this variation; first, we notice a significant improvement in the performance of not only LOO, but all methods~(except for random). Second, even with improved performance, LOO tends to plateau after a certain point, being surpassed by both BoostIn and LeafRefit. %

\revised{

The contrast between LOO and LeafRefit is particularly illuminating, because as detailed in~\S\ref{sec:leaf_refit}, the \emph{only} difference between LOO and LeafRefit is that LeafRefit makes a fixed-structure assumption (i.e., $z_i$ is ``deleted'' from the leaves where it appears, but the tree structure is left intact).
Since LOO measures each single deletion's effect exactly, LOO outperforms all other methods in the \emph{single deletion case} (shown by the ``spike'' in test loss).  %
However, as the size of the deletion group increases beyond one, LOO \emph{greedily} selects subsequent training examples to add to the deletion group.
LOO's poor results provides evidence that considering structural changes while making greedy selections is counterproductive, often leading to suboptimal deletion groups, especially for larger group sizes.
In contrast, LeafRefit ignores structural changes, instead selecting instances based on their affinity to the target instance.
Hence, as more instances are added to the deletion group, LeafRefit quickly outperforms LOO. 
This phenomenon holds regardless of whether each training instance's influence is measured once or when the remaining instances' influences are reestimated and reordered after each deletion.

The key takeaway of these results is that \textit{the fixed structure assumption yields better group influence estimates}.
In short, the assumption allows an estimator to ignore the noisy effects of structural changes and instead focus on the training instances' similarity to the target.

}

\begin{figure}[t]
\centering
\includegraphics[width=1.0\textwidth]{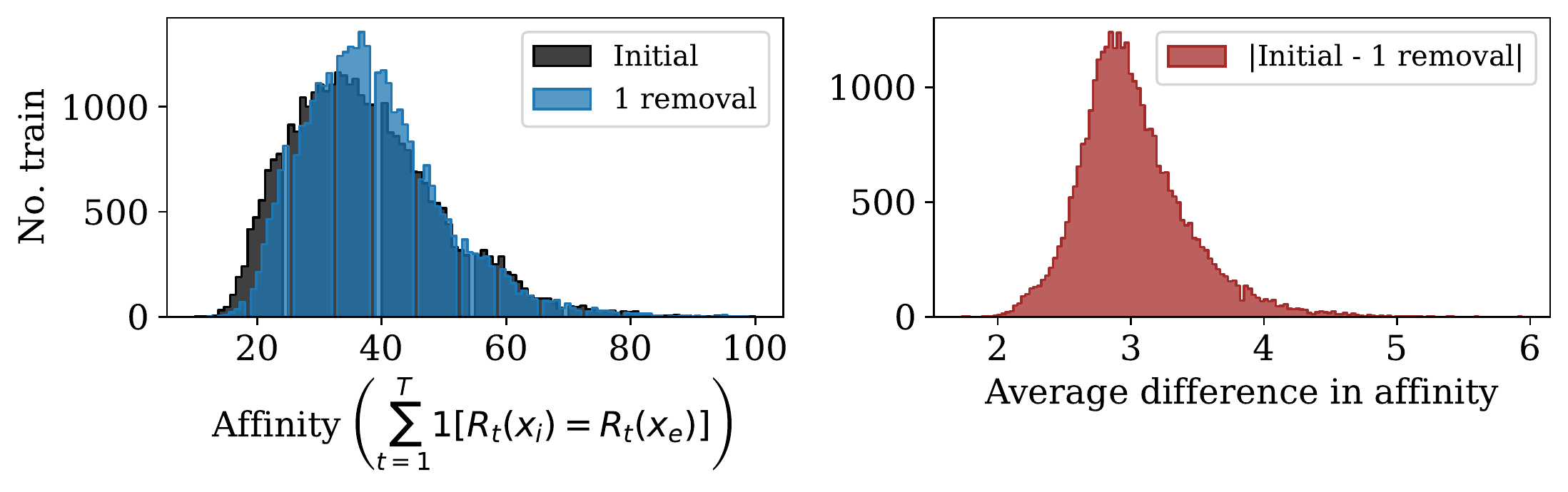}
\caption{\emph{Left}: Distribution of training-example affinities to a randomly selected test example~$z_e$ using an LGB model trained on the Adult data set before~(initial) and after the removal of a single training example~(1 removal) ordered using LOO. \emph{Right}: Average change in affinity over 100 randomly selected test examples. The changing distribution of affinity values signals structural changes to the tree structures after only a single removal.}
\label{fig:structure}
\end{figure}

Figure~\ref{fig:structure} validates structural changes are induced by a single removal ordered by LOO; specifically, the figure reports the training-example affinities (i.e.,  the number of times the training examples are assigned to the same leaf as the test example across all trees in the ensemble).

\section{Additional Related Work}
\label{sec:related_work}

\cite{kong2021understanding} adapt TracIn to variational autoencoders while~\cite{zhang2021sample} adapt influence functions and TracIn to NLP models with transformer architectures, improving the understanding of influence estimation in the context of unsupervised and large deep learning models, respectively. In a similar vein,~\cite{terashita2021influence} adapt existing influence-estimation methods to generative adversarial networks~(GANs).

HyDRA~\citep{chen2021hydra} and SGDCleanse~\citep{hara2019data} are two additional influence-estimation methods designed for deep learning models. They both compute the influence of~$z_i$ on~$z_e$ by estimating the effect of~$z_i$ on the \emph{entire} model~(including the optimization process), and then estimating the effect the changing model has on the loss of~$z_e$. These approaches are similar but significantly different from influence functions, which measures the change in the model with respect to the \emph{final} model parameters; this is also different from TracIn, which treats each checkpoint as independent and aggregates the marginal \emph{local} changes on the loss of~$z_e$ at each one. Translating these approaches to GBDTs would require a way of estimating the effect of~$z_i$ on the model \emph{structure}, of which there are an exponential number of potential structures; thus, it is not immediately obvious how one would approximate the effect of~$z_i$ on all relevant models. In our experiments, LOO and SubSample substitute as reasonable approximations for estimating structural changes; however, it may be valuable future work to explore this direction in greater depth.

A tangentially-related direction of research is \emph{prototypes}~(and their complement:~\emph{criticisms}); prototypes attempt to summarize a data set by identifying training examples in high- and low-density regions of the input space~\citep{bien2011prototype,kim2016examples,gurumoorthy2019efficient}. Although typically model-agnostic, model-specific versions exist such as TreeProto~\citep{tan2020tree}. These approaches tend to work well at providing a \emph{global} perspective of a given data set; however, these approaches differ significantly from the methods described in this paper, which attempt to return the most influential training examples for a \emph{given} test example or \emph{set} of test examples. Similar to prototypes, data set cartography maps~\citep{swayamdipta2020dataset} provide a global perspective of the training data set, characterizing training examples as easy, hard, or ambiguous to learn by measuring the confidence and variability of their predictions throughout the training process.

A highly related but significantly different body of research is \emph{feature-based} influence estimation. There is a plethora of work in this area~\citep{ribeiro2016should,selvaraju2017grad,lundberg2018consistent,ghorbani2019towards,koh2020concept,sundararajan2020many} which estimates the loss of~$z_e$ with respect to each \emph{feature}. Although feature-based approaches are currently more prevalent than instance-attribution methods, instance-based influence techniques are becoming increasingly popular as machine-learning practitioners and researchers are shifting their focus from solely analyzing the quality of their models to analyzing the quality of their data and the effect their data has on their models~\citep{barshan2020relatif,brophy2021machine,hammoudeh2022identifying}. Both influence approaches are not mutually exclusive, and using a combination of feature- and instance-based influence estimation may provide the most informative context for a given prediction.a

\daniel{Consider cutting? This mostly applies to all influence estimation methods.}\jonathan{Trimmed.}

\section{Conclusions and Future Work}
\label{sec:conclusion}

In this work, we adapt recent popular influence-estimation methods designed for deep learning models to GBDTs, identify theoretical similarities between BoostIn and LeafInfluence, and provide a comprehensive evaluation of each influence method across many data sets using multiple GBDT implementations.

Overall, we find LeafRefit typically works best at finding influential examples for a \emph{single} test instance; however, this method is extremely slow and intractable in most cases. Thus, we believe BoostIn is a viable alternative that performs comparably for the single test case, and significantly outperforms LeafRefit, LeafInfluence, and most other methods when identifying influential instances for a \emph{set} of test instances while being orders of magnitude more efficient, providing an effective and scalable solution for influence estimation in GBDTs.

Our findings also suggest that LOO consistently identifies the \emph{single-most} influential example to a given test prediction, but performs poorly at finding the most influential \emph{set} of examples due to small but significant structural changes in response to removing one or very few examples. We find methods assuming a fixed structure when computing influence values generally perform better than model-agnostic approaches that do not. These structural changes also help explain why LOO is not correlated with any other methods, and why the previous state-of-the-art, LeafInfluence, is actually a poor approximation of LOO.

For future work, further investigation regarding the structural sensitivity of GBDTs as it relates to influence estimation may lead to influence methods that better approximate the changing GBDT model resulting from one or more data removals. Another potential avenue is trying to determine the \emph{exact} number of training examples responsible for a given prediction. Finally, testing the effectiveness of these methods for different applications mentioned in the introduction would also be valuable future work.

\acks{This work was supported by a grant from the Air Force Research Laboratory and the Defense Advanced Research Projects Agency (DARPA) --- agreement number FA8750-16-C-0166, subcontract K001892-00-S05, as well as a second grant from DARPA, agreement number HR00112090135. This work benefited from access to the University of Oregon high-performance computer, Talapas.}

\appendix

\section{Implementation and Experiment Details}
\label{appendix_sec:experiment_details}

Experiments are run on an Intel(R) Xeon(R) CPU E5-2690 v4 @ 2.6GHz with 100GB of DDR4 RAM @ 2.4GHZ. We use Cython~\citep{behnel2011cython}, a Python package allowing the development of C extensions, to store a unified representation of the model structure to which we can then apply the specified influence-estimation method. Experiments are run using Python 3.9.6, and source code for all influence-estimation implementations and all experiments is available at \url{https://github.com/jjbrophy47/tree_influence}. The library currently supports all major modern gradient boosting frameworks including Scikit-learn\footnote{For our experiments, we use \texttt{HistGradientBoostingRegressor} and \texttt{HistGradientBoostingClassifier}.}~\citep{scikit-learn}, XGBoost~\citep{chen2016xgboost}, LightGBM~\citep{ke2017lightgbm}, and CatBoost~\citep{prokhorenkova2018catboost}.

\subsection{Data Sets}
\label{appendix_sec:datasets}

We perform experiments on 22 real-world data sets that represent problems well-suited for tree-based models. For each data set, we generate one-hot encodings for any categorical variable and leave all numeric and binary variables as is. For any data set without a designated train and test split, we randomly sample 80\% of the data for training and use the rest for testing. Table~\ref{tab:datasets} summarizes the data sets after preprocessing.

\begin{itemize}

    \item \textbf{Adult}~\citep{Dua:2019} contains 48,842 instances (11,687 positive) of 14 demographic attributes to determine if a person's personal income level is more than \$50K per year~(binary classification).
    
    \item \textbf{Bank}~\citep{moro2014data,Dua:2019} consists of 41,188 marketing phone calls (4,640 positive) from a Portuguese banking institution. There are 20 attributes, and the aim is to figure out if a client will subscribe~(binary classification).
    
    \item \textbf{Bean}~\citep{koklu2020multiclass,Dua:2019} consists of 13,611 images of grains. The aim is to classify each image into one of 7 different types of registered dry beans based on 16 features extracted from the image~(multiclass classification).
    
    \item \textbf{COMPAS}~\citep{compas1,compas2} is a recidivism data set consisting of 6,172 defendants~(2,751 deemed ``high-risk'') characterized by 11 attributes. The aim is to decide whether or not the defendant is at `high-risk'' to be a repeat offender~(binary classification).
    
    \item \textbf{Concrete}~\citep{yeh1998modeling,Dua:2019} consists of 1,030 instances of concrete characterized by 8 attributes. The aim is to predict the compressive strength of the concrete~(regression).
    
    \item \textbf{Credit}~\citep{yeh2009comparisons,Dua:2019} consists of the payment credibility of 30,000 people in Taiwan~(6,636 people with bad credibility). Each person is characterized by 23 attributes relating to default payments. The aim is to predict the credibility of the client~(binary classification).

    \item \textbf{Diabetes}~\citep{strack2014impact,Dua:2019} consists of 101,766 instances of patient and hospital readmission outcomes (46,902 readmitted) characterized by 55 attributes~(binary classification).
    
    \item \textbf{Energy}~\citep{tsanas2012accurate,Dua:2019} consists of 768 buildings in which each building is one of 12 different shapes and is characterized by 8 features. The aim is to predict the cooling load associated with the building~(regression).
    
    \item \textbf{Flight}~\citep{flight_delays} consists of 100,000 actual arrival and departure times of flights by certified U.S. air carriers; the data was collected by the Bureau of Transportation Statistics' (BTS) Office of Airline Information. The data contains 8 attributes and 19,044 delays. The task is to predict if a flight will be delayed~(binary classification).
    
    \item \textbf{German}~\citep{Dua:2019} consists of 1,000 credit applicants characterized by 20 attributes. The aim is to predict whether the person is a good or bad credit risk~(binary classification).
    
    \item \textbf{HTRU2}~\citep{lyon2016fifty,Dua:2019} consists of 17,898 pulsar candidates characterized by 8 attributes. The aim is to predict whether the pulsar is legitimate or a spurious example~(binary classification).
    
    \item \textbf{Life}~\citep{life} consists of 2,928 instances of life expectancy estimates for various countries during a specific year. Each instance is characterized by 20 attributes, and the aim is to predict the life expectancy of the country during a specific year~(regression).
    
    \item \textbf{Naval}~\citep{coraddu2016machine,Dua:2019} consists of 11,934 instances extracted from a high-performing gas turbine simulation. Each instance is characterized by 16 features. The aim is to predict the gas turbine decay coefficient~(regression).
    
    \item \textbf{No Show}~\citep{no_show} contains 110,527 instances of patient attendances for doctors' appointments  (22,319 no shows) characterized by 14 attributes. The aim is to predict whether or not a patient shows up to their doctors' appointment~(binary classification).
    
    \item \textbf{Obesity}~\citep{obesity} contains 48,346 instances of obesity rates for different states and regions with differing socioeconomic backgrounds. Each instance is characterized by 32 attributes. The aim is to predict the obesity rate of the region~(regression).
    
    \item \textbf{Power}~\citep{kaya2012local,tufekci2014prediction,Dua:2019} contains 9,568 readings of a Combined Cycle Power Plant~(CCPP) at full work load. Each reading is characterized by 4 features. The aim is to predict the net hourly electrical energy output~(regression).
    
    \item \textbf{Protein}~\citep{Dua:2019} contains 45,730 tertiary-protein-structure instances characterized by 9 attributes. The aim is to predict the armstrong coefficient of the protein structure~(regression).
    
    \item \textbf{Spambase}~\citep{Dua:2019} consists of 4,601 emails~(1,813 spam) characterized by 57 attributes. The aim is to predict whether or not the email is spam~(binary classification).
    
    \item \textbf{Surgical}~\citep{surgical} consists of 14,635 medical patient surgeries (3,690 surgeries with complications), characterized by 25 attributes; the goal is to predict whether or not a patient had a complication from their surgery~(binary classification).
    
    \item \textbf{Twitter} uses the first 250,000 tweets (33,843 spam) of the HSpam14 data set~\citep{sedhai2015hspam14}. Each instance contains the tweet ID and label. After retrieving the text and user ID for each tweet, we derive the following attributes: no.\ chars, no.\ hashtags, no.\ mentions, no.\ links, no.\ retweets, no.\ unicode chars., and no.\ messages per user. The aim is to predict whether a tweet is spam or not~(binary classification).
    
    \item \textbf{Vaccine}~\citep{bull2016harnessing,vaccine} consists of 26,707 survey responses collected between October 2009 and June 2010 asking people a range of 36 behavioral and personal questions, and ultimately asking whether or not they got an H1N1 and/or seasonal flu vaccine. Our aim is to predict whether or not a person received a seasonal flu vaccine~(binary classification).
    
    \item \textbf{Wine}~\citep{cortez2009modeling,Dua:2019} consists of 6,497 instances of Portuguese ``Vinho Verde'' red and white wine. Each instance is characterized by 11 features. The aim is to predict the quality of the wine from 0-10~(regression).

\end{itemize}

\begin{table*}[h]
\centering
\begin{tabular}{lrrrrrrrrr}
\toprule
\textbf{Data set} & \textbf{Task} & \textbf{Metric} & \textbf{No. instances} & \textbf{Pos. \%} & \textbf{No.\ attr.} & \textbf{SDS?} \\
\midrule
Bank            & binary     & AUC  & 41,188    & 11.3 &  63 &            \\
Flight          & binary     & AUC  & 100,000   & 19.0 & 650 &            \\
HTRU2           & binary     & AUC  & 17,898    &  9.2 &   8 & \checkmark \\
No Show         & binary     & AUC  & 110,527   & 20.2 &  89 &            \\
Twitter         & binary     & AUC  & 250,000   & 13.5 &  14 &            \\
\midrule
Adult           & binary     & Acc. & 48,842    & 23.9 & 108 &            \\
COMPAS          & binary     & Acc. & 6,172     & 44.6 &  10 & \checkmark \\
Credit Card     & binary     & Acc. & 30,000    & 22.1 &  23 & \checkmark \\
Diabetes        & binary     & Acc. & 101,766   & 46.1 & 255 &            \\
German          & binary     & Acc. & 1,000     & 30.0 &  27 & \checkmark \\
Spambase        & binary     & Acc. & 4,601     & 39.4 &  57 & \checkmark \\
Surgical        & binary     & Acc. & 14,635    & 25.2 &  90 & \checkmark \\
Vaccine         & binary     & Acc. & 26,707    & 46.6 & 155 &            \\
\midrule
Bean            & multiclass & Acc. & 13,611    & -    &  16 & \checkmark \\
\midrule
Concrete        & regression & MSE  & 1,030     & -    &   8 & \checkmark \\
Energy          & regression & MSE  & 768       & -    &  16 & \checkmark \\
Life            & regression & MSE  & 2,928     & -    & 204 & \checkmark \\
Naval           & regression & MSE  & 11,934    & -    &  17 & \checkmark \\
Obesity         & regression & MSE  & 48,346    & -    & 100 &            \\
Power           & regression & MSE  & 9,568     & -    &   4 & \checkmark \\
Protein         & regression & MSE  & 45,730    & -    &   9 &            \\
Wine            & regression & MSE  & 6,497     & -    &  11 & \checkmark \\
\bottomrule
\end{tabular}
\caption{Data set summary after preprocessing. AUC = area under the ROC curve, Acc. = Accuracy, MSE = mean squared error, No. attr. = number of attributes, SDS = small data subset (data sets for which LeafRefit and LeafInfluence are tractable).}
\label{tab:datasets}
\end{table*}

\newpage
\hphantom{dummy}
\newpage

\subsection{Predictive Performance of GBDTs}
\label{appendix_sec:gbdt_predictive_performance}

This section evaluates the predictive performance of the four most-popular modern gradient-boosting frameworks: LightGBM~(LGB), XGBoost~(XGB), CatBoost~(CB), and Scikit-learn boosting~(SGB).

\begin{table*}[h]
\small
\centering
\begin{tabular}{lcccc|cccccc}
\toprule
& \multicolumn{4}{c}{\textbf{GBDT}} & \multicolumn{5}{c}{\textbf{Non-GBDT}} \\
\cmidrule(lr){2-5}\cmidrule(lr){6-11}
\textbf{Data set} & \textbf{LGB} & \textbf{XGB} & \textbf{CB} & \textbf{SGB} & \textbf{LR} & \textbf{DT} & \textbf{KNN} & \textbf{SVM} & \textbf{RF} & \textbf{MLP} \\
\midrule
\multicolumn{11}{c}{\textbf{AUC (binary classification)~($\uparrow$)}} \\
Bank            & \textBF{0.951}   & 0.947          & 0.948             & 0.949           
                & 0.932            & 0.930          & 0.930             & 0.930           & 0.924 & 0.934   \\
Flight          & 0.748            & \textBF{0.749} & 0.745             & 0.747               
                & 0.707            & 0.696          & 0.687             & 0.662           & 0.688 & 0.720   \\
HTRU2           & \textBF{0.982}   & 0.981          & 0.981             & 0.981           
                & 0.978            & 0.966          & 0.964             & 0.955           & 0.977 & 0.972   \\
No Show         & 0.621            & \textBF{0.622} & 0.621             & 0.621           
                & 0.601            & 0.603          & 0.592             & 0.538           & 0.612 & 0.609   \\
Twitter         & \textBF{0.927}   & 0.924          & 0.917             & \textBF{0.927}  
                & 0.808            & 0.893          & 0.859             & 0.836           & 0.884 & 0.897   \\
\midrule
\multicolumn{11}{c}{\textbf{Accuracy (binary classification)~($\uparrow$)}} \\
Adult           & \textBF{0.874}   & \textBF{0.874} & \textBF{0.874}    & 0.873           
                & 0.853            & 0.861          & 0.803             & 0.852           & 0.852 & 0.818   \\
COMPAS          & 0.752            & 0.770          & \textBF{0.777}    & 0.770           
                & 0.768            & 0.747          & 0.746             & 0.760           & 0.768 & 0.769   \\
Credit          & \textBF{0.822}   & \textBF{0.822} & 0.820             & 0.821           
                & 0.810            & \textBF{0.822} & 0.780             & 0.819           & 0.821 & 0.760   \\
Diabetes        & 0.648            & 0.648          & \textBF{0.650}    & 0.648           
                & 0.637            & 0.631          & 0.602             & 0.643           & 0.628 & 0.626   \\
German          & 0.735            & 0.710          & 0.705             & \textBF{0.745}  
                & 0.730            & 0.730          & 0.720             & 0.720           & 0.720 & 0.645   \\
Spambase        & \textBF{0.957}   & 0.952          & 0.955             & \textBF{0.957}  
                & 0.933            & 0.941          & 0.810             & 0.940           & 0.932 & 0.941   \\
Surgical        & \textBF{0.909}   & \textBF{0.909} & \textBF{0.909}    & 0.908           
                & 0.800            & 0.894          & 0.886             & 0.803           & 0.821 & 0.788   \\
Vaccine         & 0.811            & 0.811          & 0.807             & \textBF{0.813}  
                & 0.807            & 0.780          & 0.771             & 0.805           & 0.785 & 0.750   \\
\midrule
\multicolumn{11}{c}{\textbf{Accuracy (multiclass classification)~($\uparrow$)}} \\
Bean            & 0.930            & \textBF{0.931} & \textBF{0.931}    & 0.930           
                & 0.927            & 0.908          & 0.737             & \textBF{0.931}  & 0.918 & 0.505   \\
\midrule
\multicolumn{11}{c}{\textbf{MSE (regression)~($\downarrow$)}} \\
Concrete        & 20.1             & 21.6           & 23.9              & \textBF{18.8}   
                & 124.9            & 69.4           & 98.9              & 102.4           & 54.7 & 50.6    \\
Energy          & 0.28             & \textBF{0.10}  & 0.13              & 0.26           
                & 0.97             & 0.36           & 2.10              & 10.03           & 0.36 & 20.18   \\
Life            & \textBF{3.21}    & 3.30           & 3.22              & 3.40           
                & 3.72             & 6.98           & 69.36             & 8.44           & 6.99 & 6e4   \\
Naval           & \textBF{4e-7}    & 8e-7           & 6e-6               & \textBF{4e-7} 
                & 3e-6             & 1e-6           & 6e-6               & 6e-5           & 2e-5 & 1e1   \\
Obesity         & \textBF{0.027}   & 0.043          & 0.033             & 0.038           
                & 0.115            & 0.043          & 5.191             & 0.786           & 0.676 & 1.0e4   \\
Power           & 8.5              & \textBF{8.4}   & 8.8               & 8.6           
                & 20.3             & 16.1           & 15.1              & 17.1           & 16.5 & 24.7   \\
Protein         & 13.4             & 13.6           & 14.9              & \textBF{13.3}  
                & 26.6             & 20.9           & 33.2              & 23.9            & 23.9 & 329.1   \\
Wine            & \textBF{0.384}   & 0.422          & 0.427             & 0.389           
                & 0.528            & 0.524          & 0.626             & 0.446           & 0.524 & 0.512   \\
\bottomrule
\end{tabular}
\caption{Predictive performance of GBDTs against alternative methods: logistic regression~(LR), decision tree~(DT), $k$-nearest neighbor~(KNN), support vector machine with an RBF kernel~(SVM), random forest~(RF), and a multilayer peceptron~(MLP), all evaluated on the test set of each data set. We use MSE to evaluate regression models, accuracy~(acc.) for models trained on multiclass data sets or binary data sets with a positive label percentage $> 20\%$, and AUC for the rest; see Table~\ref{tab:datasets} for reference.}
\label{tab:predictive_performance}
\end{table*}

We compare LGB, XGB, CB, and SGB against alternative methods that are arguably more interpretable:

\begin{itemize}

    \item Logistic regression~(LR): Logistic regression implementation from Scikit-Learn~\citep{scikit-learn}; we tune the regularization hyperparameter using values~[l1, l2], and the penalty hyperparameter C using values~[0.01, 0.1, 1.0].
    
    \item Decision tree~(DT): Single decision tree implementation from Scikit-Learn~\citep{scikit-learn}; we tune the split criterion hyperparameter using values~[gini, entropy], the decision node splitter using values~[best, random], and the maximum depth of the tree using values~[3, 5, 10, no limit].
    
    \item $k$-nearest neighbor~(KNN): $k$-nearest neighbor implementation from Scikit-Learn~\citep{scikit-learn}; we tune $k$ using values~[3, 5, 7, 11, 15, 31, 61].
    
    \item Support vector machine~(SVM): Support vector machine implementation from Scikit-Learn~\citep{scikit-learn}; we use a radial basis function~(RBF) kernel and tune the penalty hyperparameter $C$ using values~[0.01, 0.1, 1.0].
    
    \item Random forest~(RF): Random forest implementation from Scikit-Learn~\citep{scikit-learn}; we tune the number of trees using values~[10, 25, 50, 100, 200], and maximum depth using values~[2, 3, 4, 5, 6, 7].
    
    \item Multilayer perceptron~(MLP): Multilayer perceptron implementation from Scikit-Learn~\citep{scikit-learn}; we tune the number of layers and number of nodes per layer using values~[(100,), (100, 100)].

\end{itemize}

\begin{table*}[t]
\centering
\begin{tabular}{lrrrrrrrrrr}
\toprule
& \multicolumn{2}{c}{\textbf{LGB}} & \multicolumn{2}{c}{\textbf{XGB}} & \multicolumn{3}{c}{\textbf{CB}} & \multicolumn{3}{c}{\textbf{SGB}} \\
\cmidrule(lr){2-3}\cmidrule(lr){4-5}\cmidrule(lr){6-8}\cmidrule(lr){9-11}
\textbf{Data set} & $T$ & $l_{\max}$ & $T$ & $d_{\max}$ & $T$ & $d_{\max}$ & $\eta$ & $T$ & $l_{\max}$ & $b_{max}$ \\
\midrule
Bank            & 50  & 31 & 100 & 4 & 200 & 7 & 0.1 & 50  & 31 & 100 \\
Flight          & 200 & 91 & 200 & 7 & 200 & 7 & 0.3 & 200 & 61 & 250 \\
HTRU2           & 100 & 15 & 100 & 2 & 200 & 3 & 0.3 & 50  & 15 & 50  \\
No Show         & 50  & 61 & 100 & 5 & 200 & 7 & 0.3 & 50  & 61 & 100 \\
Twitter         & 200 & 91 & 200 & 7 & 200 & 7 & 0.6 & 200 & 91 & 250 \\
\midrule
Adult           & 100 & 31 & 200 & 3 & 200 & 4 & 0.6 & 200 & 15 & 250 \\
COMPAS          & 25  & 91 & 50  & 3 & 50  & 4 & 0.3 & 50  & 15 & 50  \\
Credit          & 50  & 15 & 10  & 3 & 50  & 5 & 0.1 & 25  & 15 & 100 \\
Diabetes        & 200 & 31 & 200 & 3 & 200 & 5 & 0.3 & 200 & 31 & 100 \\
German          & 25  & 15 & 10  & 4 & 100 & 5 & 0.1 & 25  & 15 & 100 \\
Spambase        & 200 & 31 & 200 & 4 & 200 & 5 & 0.3 & 200 & 91 & 250 \\
Surgical        & 200 & 15 & 50  & 5 & 200 & 4 & 0.1 & 100 & 31 & 250 \\
Vaccine         & 100 & 15 & 100 & 3 & 200 & 4 & 0.1 & 100 & 15 & 50  \\
\midrule
Bean            & 25  & 15 & 25  & 6 & 200 & 3 & 0.3 & 25  & 15 & 50  \\
\midrule
Concrete        & 200 & 15 & 200 & 4 & 200 & 4 & 0.3 & 200 & 15 & 50  \\
Energy          & 200 & 15 & 200 & 5 & 200 & 4 & 0.9 & 200 & 15 & 50  \\
Life            & 200 & 61 & 200 & 5 & 200 & 6 & 0.3 & 200 & 31 & 250 \\
Naval           & 200 & 91 & 100 & 7 & 200 & 7 & 0.6 & 200 & 91 & 250 \\
Obesity         & 200 & 91 & 200 & 7 & 200 & 6 & 0.6 & 200 & 91 & 250 \\
Power           & 200 & 61 & 200 & 7 & 200 & 6 & 0.6 & 200 & 61 & 250 \\
Protein         & 200 & 91 & 200 & 7 & 200 & 7 & 0.3 & 200 & 91 & 250 \\
Wine            & 200 & 91 & 100 & 7 & 200 & 7 & 0.3 & 200 & 91 & 100 \\
\bottomrule
\end{tabular}
\caption{Hyperparameters selected for the GBDT models. The number of trees/boosting iterations~($T$), maximum number of leaves~($l_{\max}$), maximum depth~($d_{\max})$, learning rate~($\eta$), and maximum number of bins~($b_{\max}$) is found using 5-fold cross-validation. Data sets are grouped based on their task and metric used for evaluation; see Table~\ref{tab:datasets} for reference.}
\label{tab:gbdt_hyperparams}
\end{table*}

For the LGB, XGB, CB, and SGB models, we tune the number of trees/boosting iterations~($T$) using values~[10, 25, 50, 100, 200]. Since the LGB and SGB models grow trees in a leaf-wise~(depth-first) manner, we tune the maximum number of leaves~($l_{\max})$ for LGB and SGB using values~[15, 31, 61, 91]. In contrast, we tune the the maximum depth~($d_{\max})$ for XGB and CB using values~[2, 3, 4, 5, 6, 7]. We also tune the learning rate~($\eta$) for CB using values~[0.1, 0.3, 0.6, 0.9], and the maximum number of bins~($b_{\max}$) for SGB using values~[50, 100, 250].

We tune all hyperparameters using 5-fold cross-validation. We use mean squared error~(MSE) to tune hyperparameters for regression tasks, accuracy for multiclass tasks, and accuracy for binary tasks with a positive label percentage $> 20\%$, otherwise we use area under the ROC curve~(AUC). The associated task and metric used to tune hyperparameters for each data set is in Table~\ref{tab:datasets}, and the selected hyperparameters for the LGB, XGB, CB, and SGB models are in Table~\ref{tab:gbdt_hyperparams}.

Table~\ref{tab:predictive_performance} shows the GBDT models consistently outperform the alternative models in terms of predictive performance. These results reaffirm the notion that GBDT models generally outperform more traditional machine-learning algorithms on tabular data and motivate the need for tailored influence-estimation methods for GBDT models to better understand their decision-making processes.

\newpage
\hphantom{dummy}
\newpage

\subsection{Summary of Results: All Data Sets}
\label{app_sec:resul_summary_all_datasets}

\begin{figure*}[ht]
\begin{subfigure}{\textwidth}
    \centering
    \includegraphics[width=\textwidth,clip,trim=0 0 0 0]{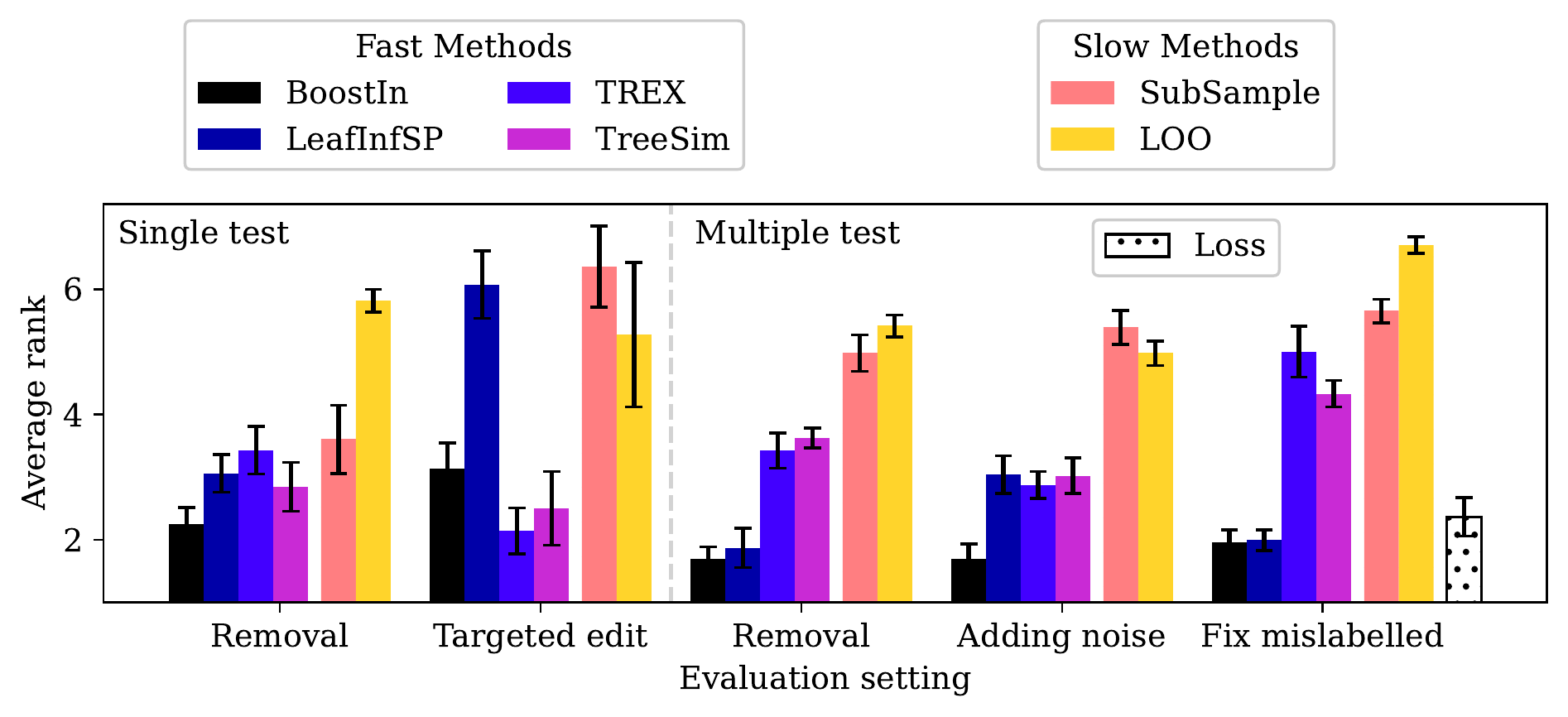}
    \caption{Average rank of each method. For each evaluation setting, results are averaged over all checkpoints, tree types, and data sets; error bars represent 95\% confidence intervals and are computed over data sets. Lower is better.}
    \vskip 2.75mm
    \label{app_fig:rank_loss}
\end{subfigure}
\begin{subfigure}{\textwidth}
    \centering
    \includegraphics[width=\textwidth,clip,trim=0 0 0 0]{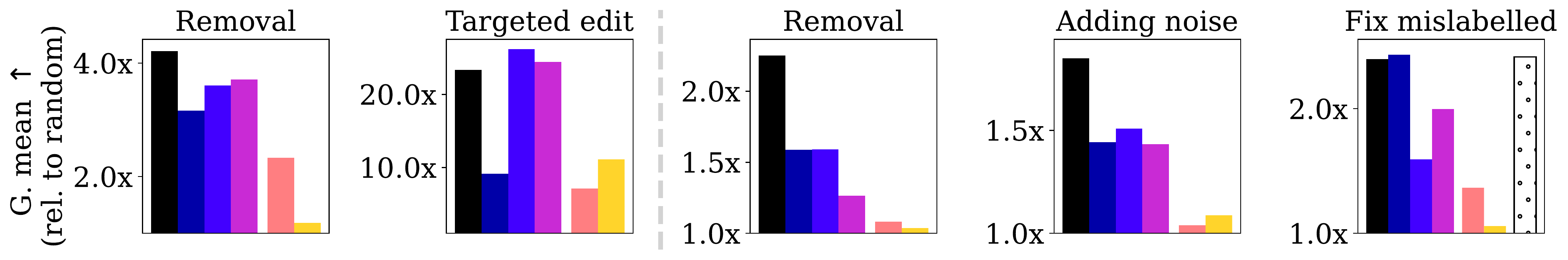}
    \caption{Average \emph{loss} increase~(except for ``fix mislabelled'', which shows average increase in \emph{mislabelled detection}) relative to random. For each evaluation setting, results are averaged over all checkpoints and tree types, then the geometric mean is computed over all data sets. Higher is better.}
    \label{app_fig:magnitude_loss}
\end{subfigure}
\caption{High-level overview of results including \emph{all} data sets; thus, LeafRefit and LeafInfluence are not included in this analysis.}
\label{app_fig:result_summary}
\end{figure*}

Figure~\ref{app_fig:result_summary} shows a high-level summary of the results including \emph{all} data sets. Overall, we observe very similar trends as when analyzing only the SDS data sets. However, we do notice a decrease in relative performance for SubSample; better performance may be achieved by increasing~$\tau$, but this may also significantly increase its running time.

\newpage

\subsection{Removing Examples (Single Test): Additional Analysis}
\label{app_sec:single_test_removal}

Figure~\ref{app_fig:single_remove_loss_rank} shows a more fine-grained analysis for the removal experiment involving a single test instance; it also
includes an additional baseline: \emph{RandomSL}, which assigns a randomly-sampled positive value for training examples with the same label as the test example, and negative otherwise:\footnote{For regression tasks,~$\mathcal{I}_{RandomSL}(z_i, z_e) = \mathcal{N}(\mu_i, \sigma_i)$ in which~$\mu_i = 1 / |y_i - y_e|$ and $\sigma_i = s.d.(|y_i - y_e|)$.}
\begin{align*}
    \mathcal{I}_{RandomSL}(z_i, z_e) &= \mathbbm{1}[y_i = y_e]U - \mathbbm{1}[y_i \not = y_e]U,\ U \sim \mathcal{U}(0, 1)
\end{align*}
Overall, the trends are relatively consistent across GBDT types; however, we observe TreeSim tends to perform better on XGB and CB than LGB and SGB.

\newcommand{\tw}{0.24}

\begin{figure}[h]
\centering
\begin{subfigure}{\tw\textwidth}
  \centering
  \includegraphics[width=1.0\linewidth]{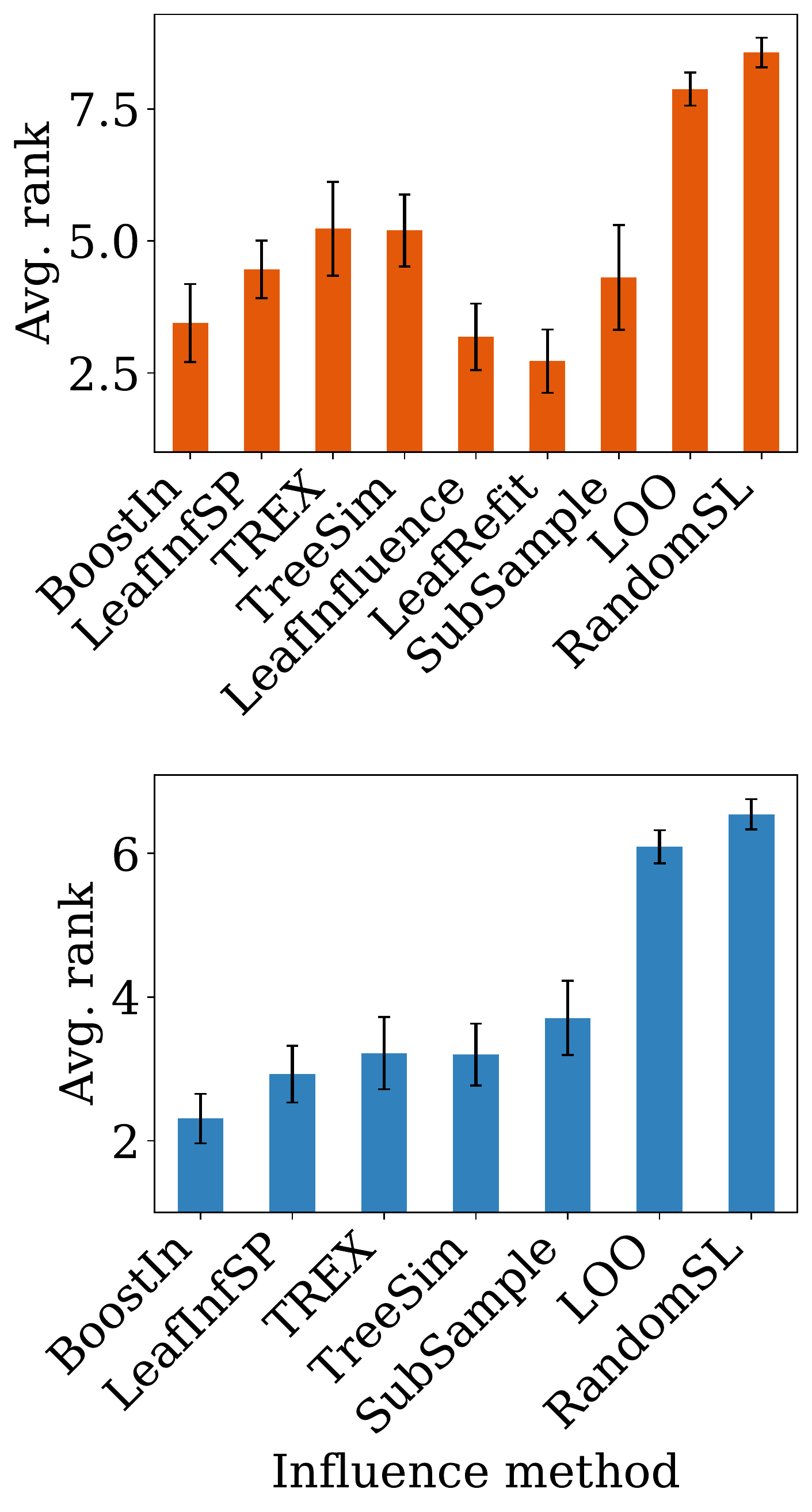}
  \caption{LGB}
  \label{app_fig:remove_loss_rank_lgb}
\end{subfigure}
\begin{subfigure}{\tw\textwidth}
  \centering
  \includegraphics[width=1.0\linewidth]{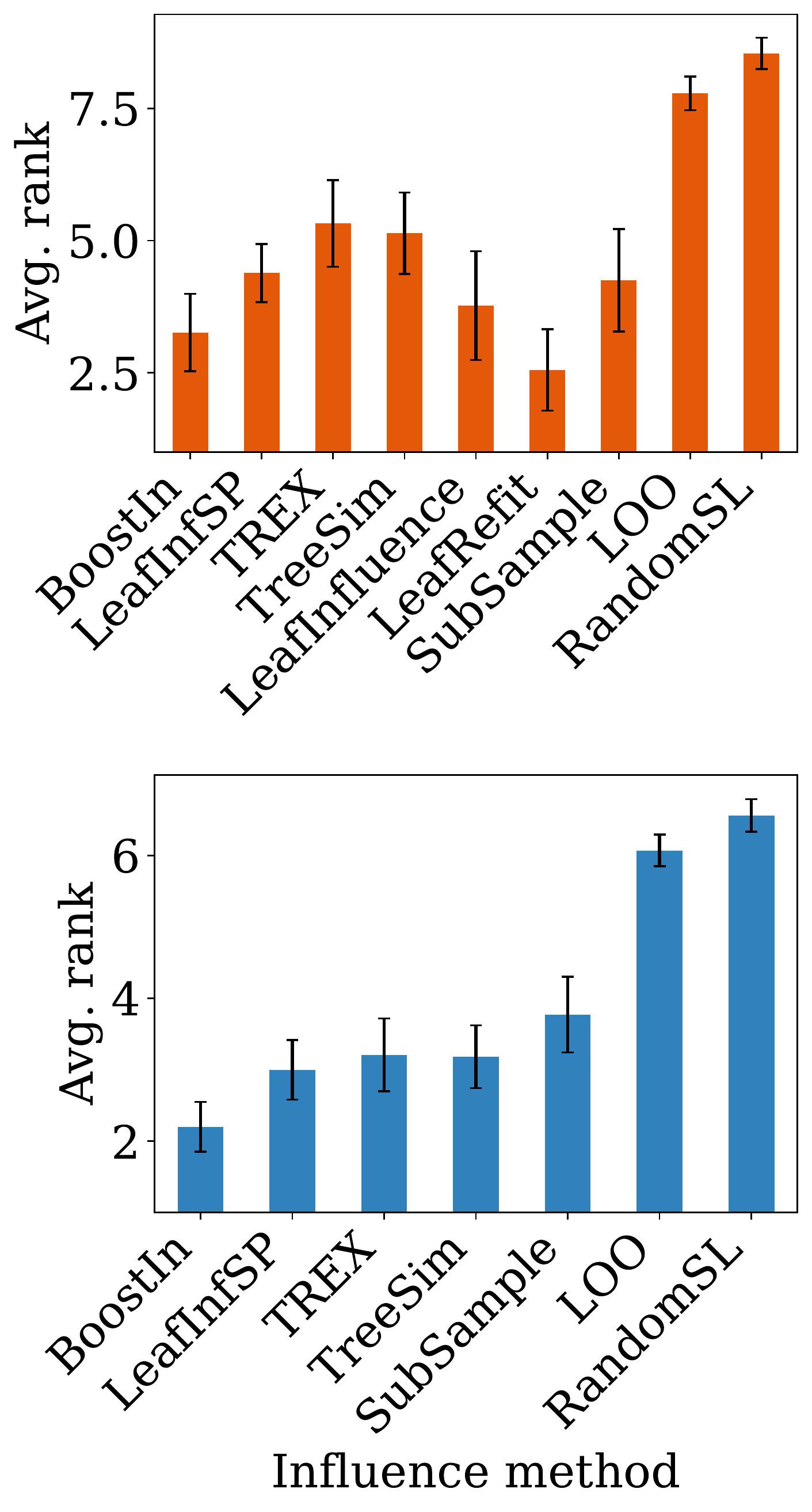}
  \caption{SGB}
  \label{app_fig:remove_loss_rank_sgb}
\end{subfigure}
\begin{subfigure}{\tw\textwidth}
  \centering
  \includegraphics[width=1.0\linewidth]{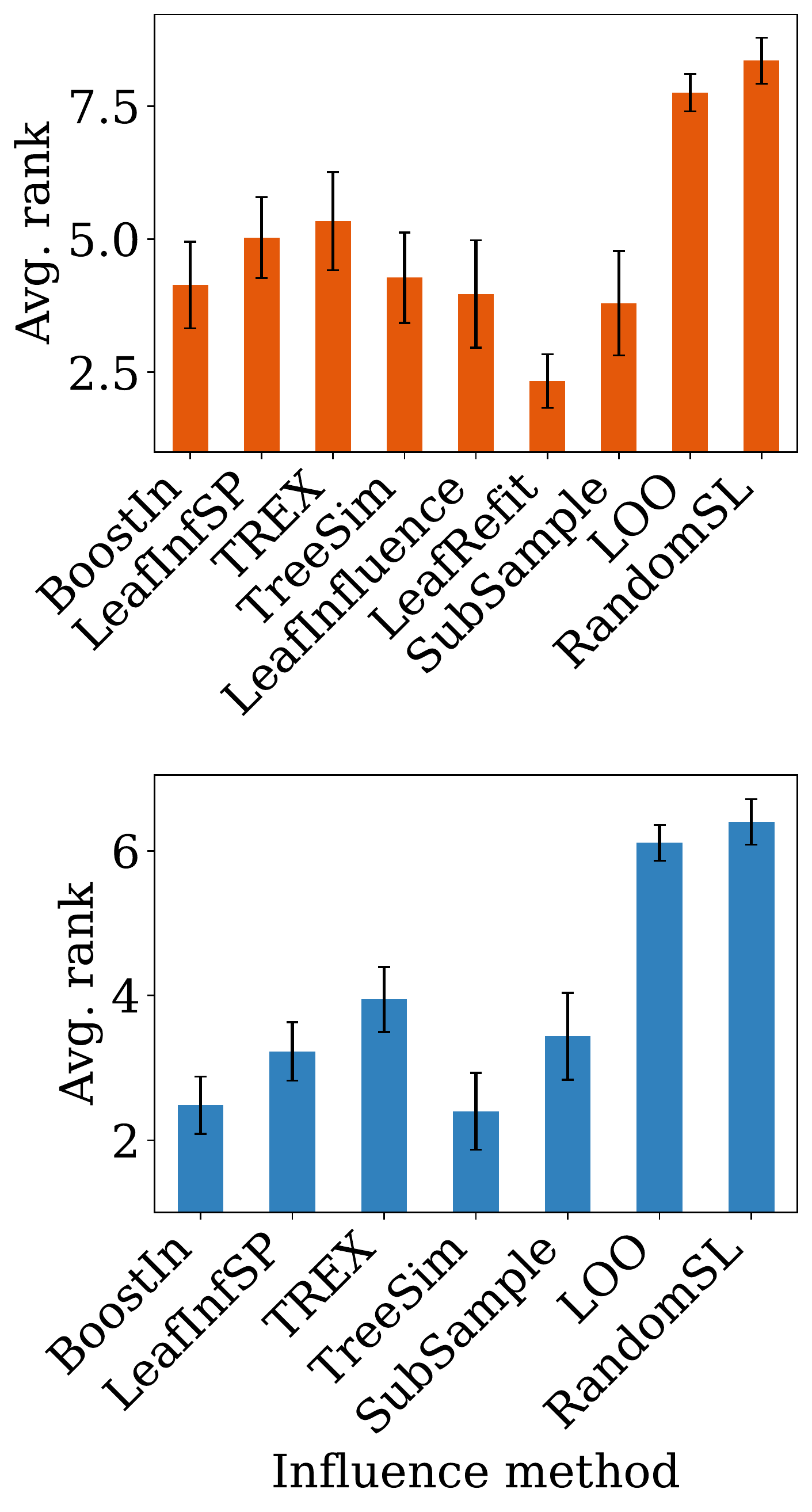}
  \caption{XGB}
  \label{app_fig:remove_loss_rank_xgb}
\end{subfigure}
\begin{subfigure}{\tw\textwidth}
  \centering
  \includegraphics[width=1.0\linewidth]{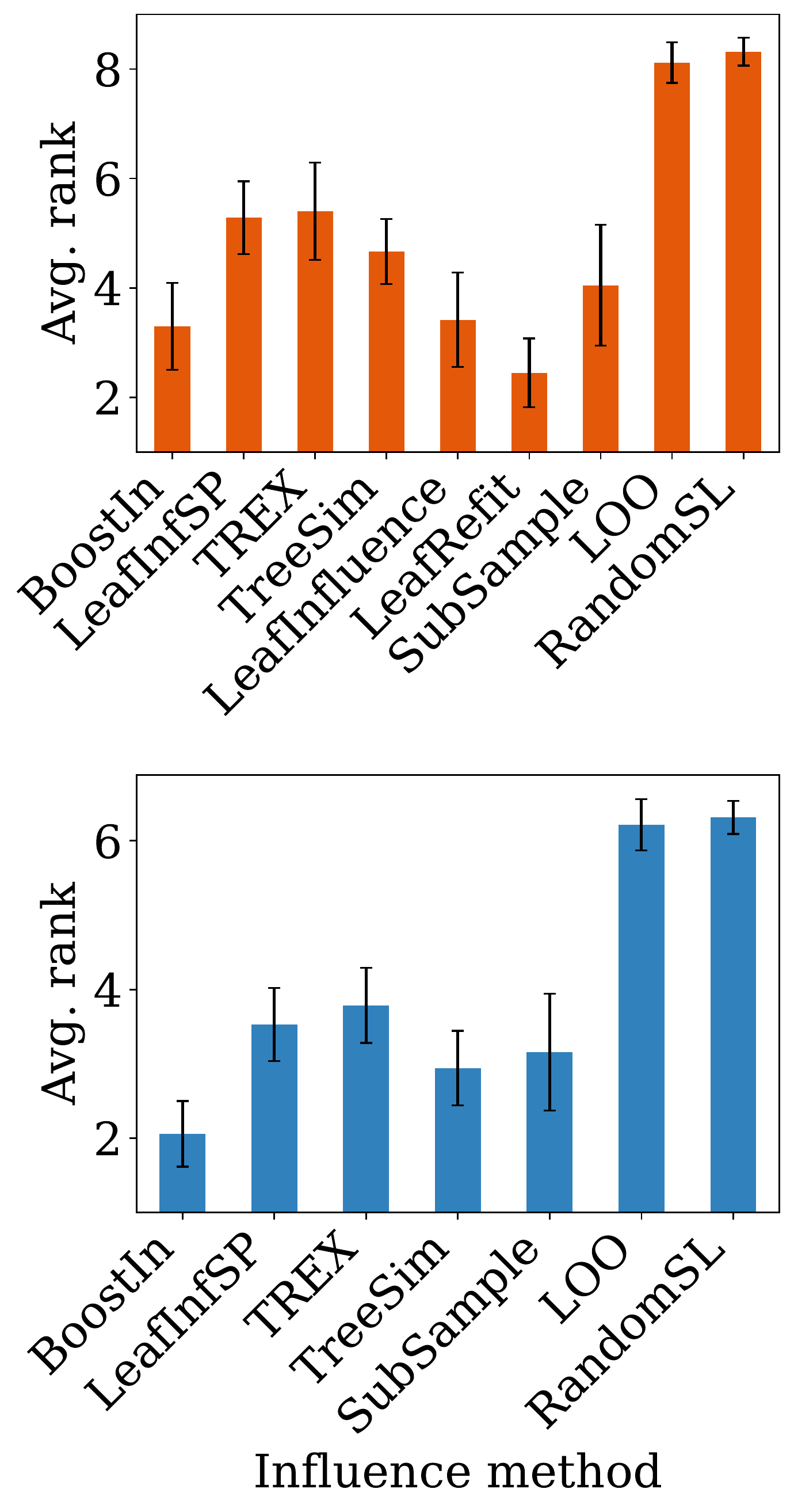}
  \caption{CB}
  \label{app_fig:remove_loss_rank_cb}
\end{subfigure}
\caption{Average ranks when removing examples for a single test instance, shown for each GBDT type. \emph{Top row}: SDS data sets; \emph{Bottom row}: all data sets. Results are averaged over checkpoints and data sets. Error bars represent 95\% confidence intervals computed over data sets. Lower is better.}
\label{app_fig:single_remove_loss_rank}
\end{figure}

Figure~\ref{app_fig:single_remove_individual} shows the change in loss for different GBDT types and data sets. Methods using the fixed-structure assumption tend to perform best, with BoostIn, LeafRefit, and LeafInfluence performing slightly better than the rest on average.

\renewcommand{\tw}{0.325}

\begin{figure}[ht]
\centering
\begin{subfigure}{\tw\textwidth}
  \centering
  \includegraphics[width=1.0\linewidth]{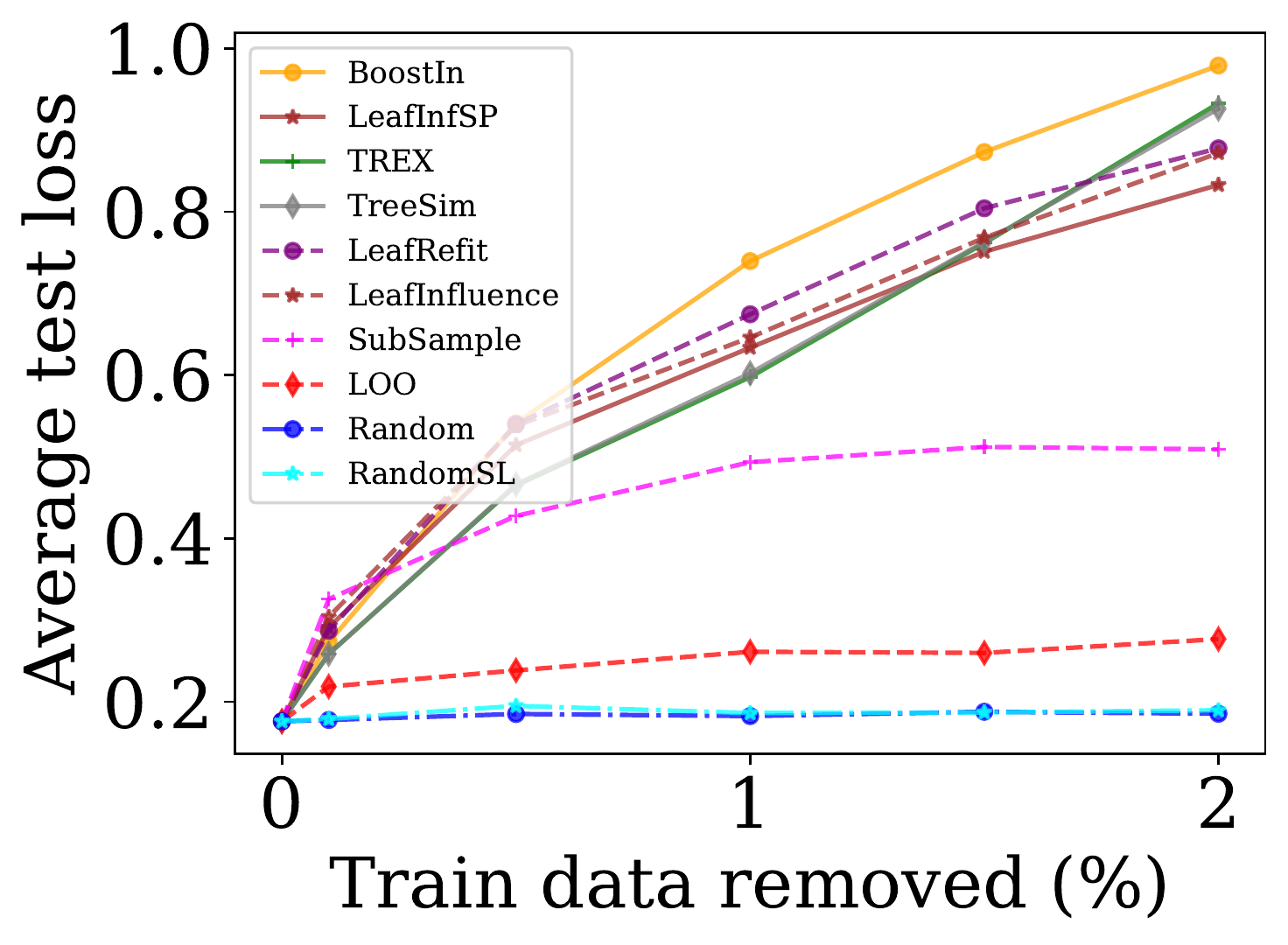}
  \caption{LGB: Bean}
\end{subfigure}
\begin{subfigure}{\tw\textwidth}
  \centering
  \includegraphics[width=1.0\linewidth]{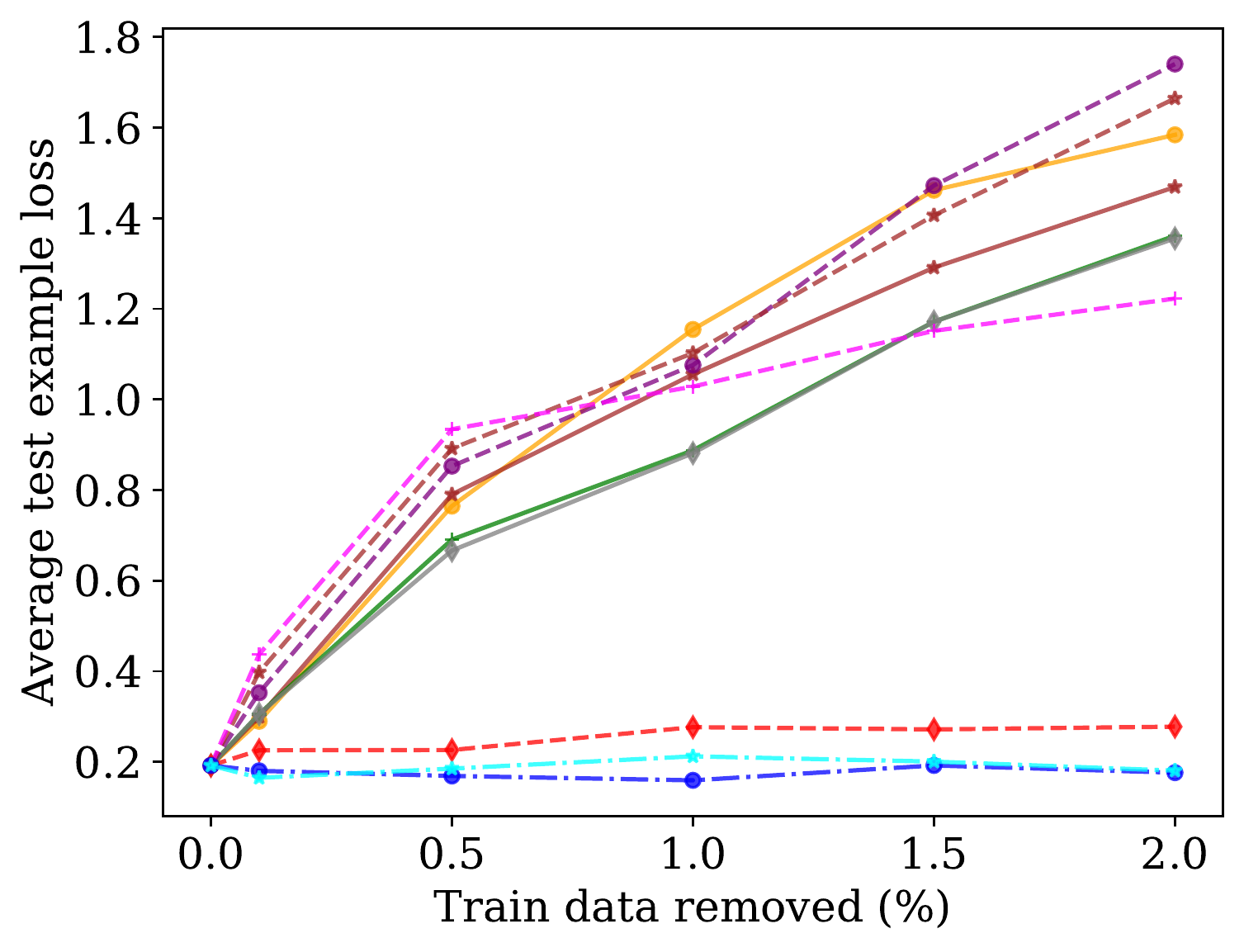}
  \caption{LGB: Spambase}
\end{subfigure}
\begin{subfigure}{\tw\textwidth}
  \centering
  \includegraphics[width=1.0\linewidth]{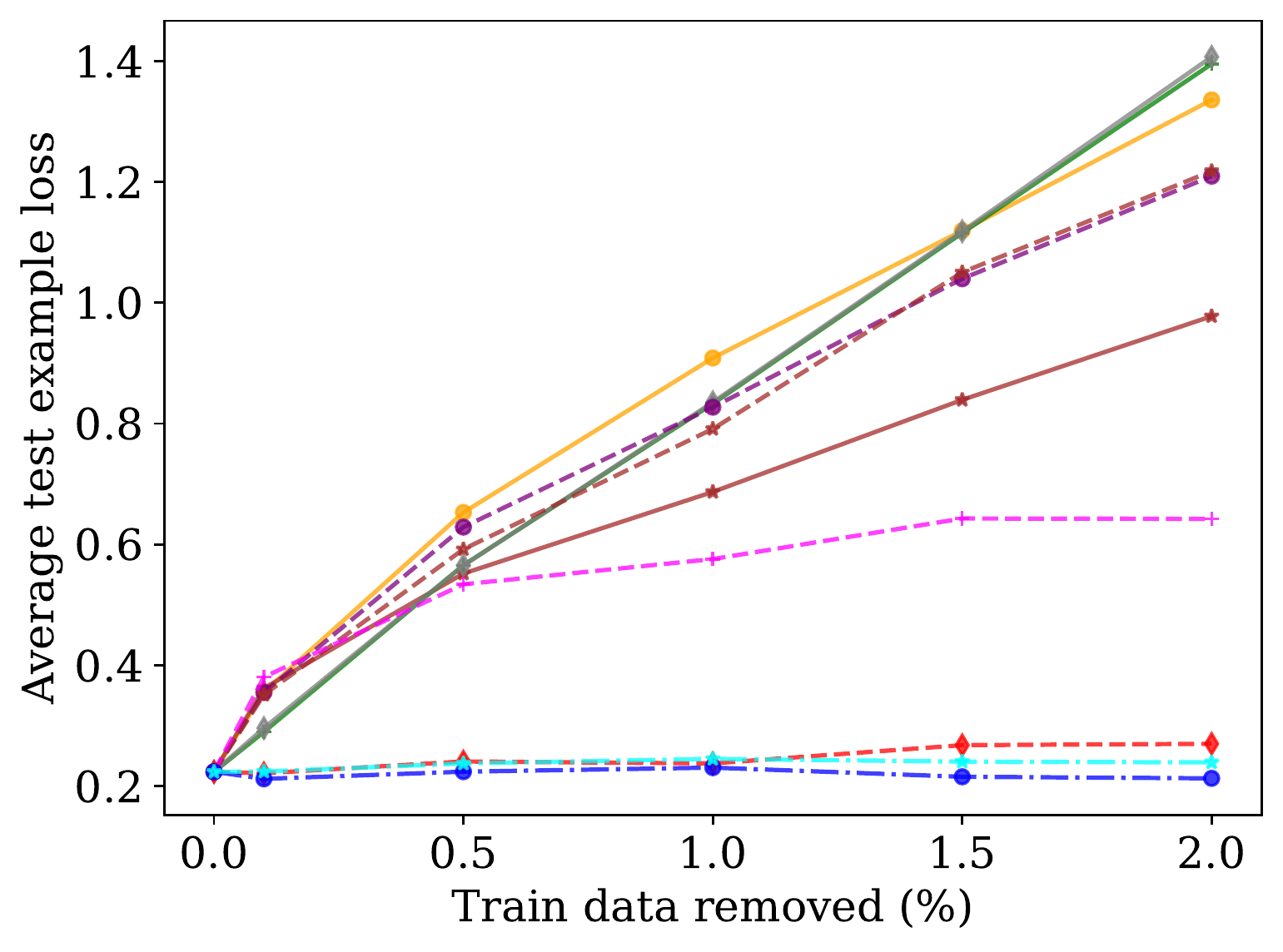}
  \caption{LGB: Surgical}
\end{subfigure}
\\
\begin{subfigure}{\tw\textwidth}
  \centering
  \includegraphics[width=1.0\linewidth]{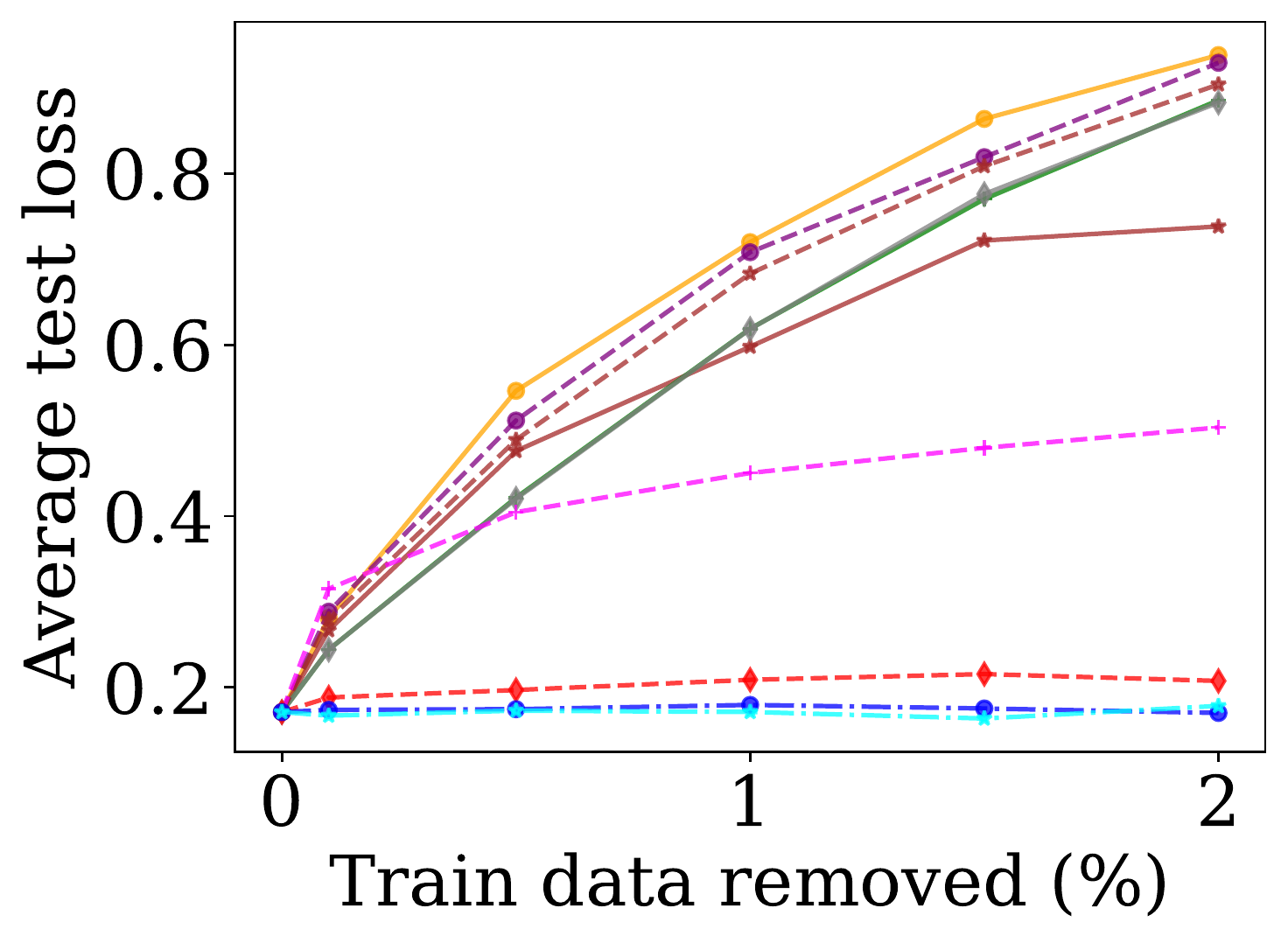}
  \caption{SGB: Bean}
\end{subfigure}
\begin{subfigure}{\tw\textwidth}
  \centering
  \includegraphics[width=1.0\linewidth]{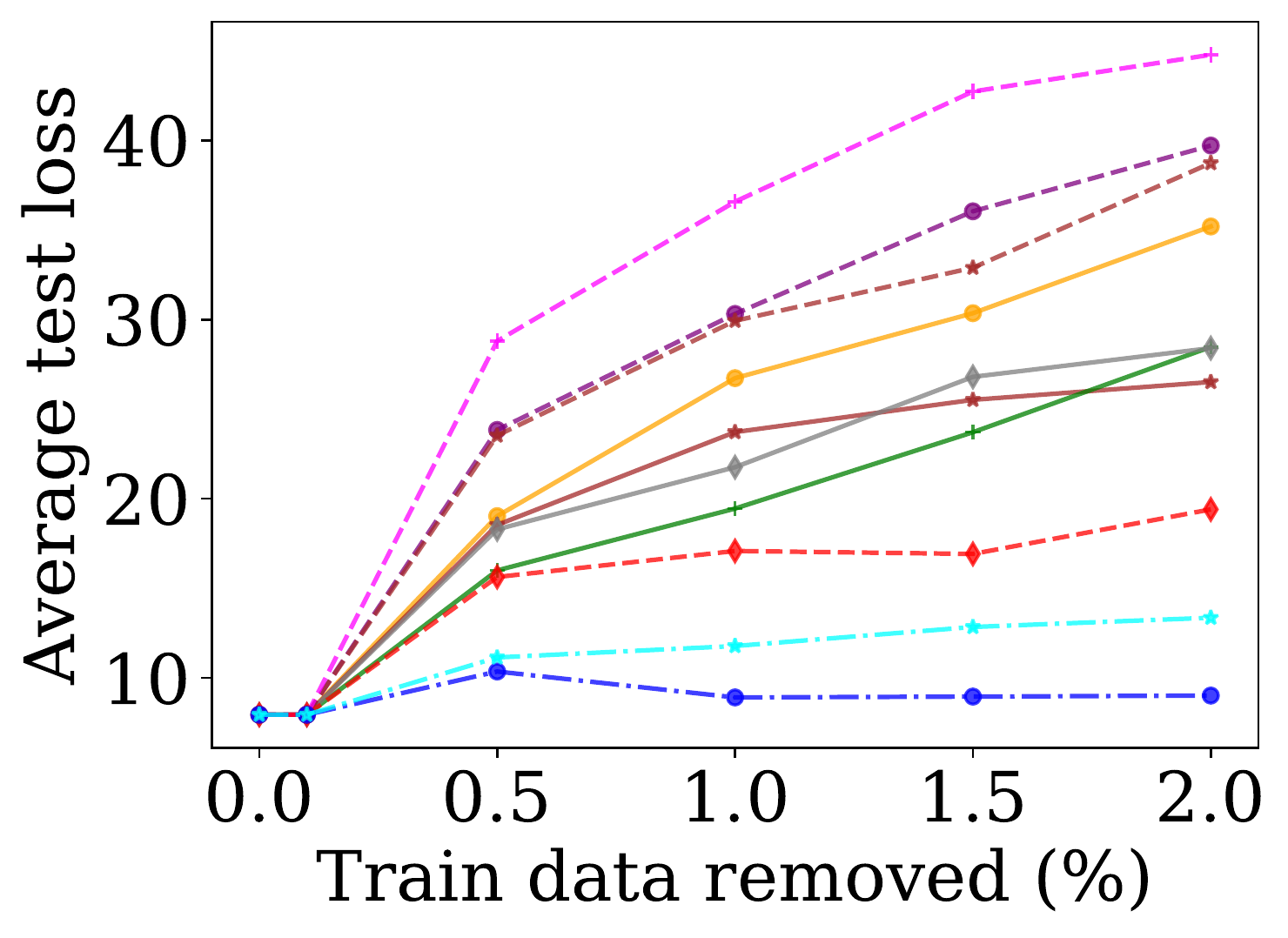}
  \caption{SGB: Concrete}
\end{subfigure}
\begin{subfigure}{\tw\textwidth}
  \centering
  \includegraphics[width=1.0\linewidth]{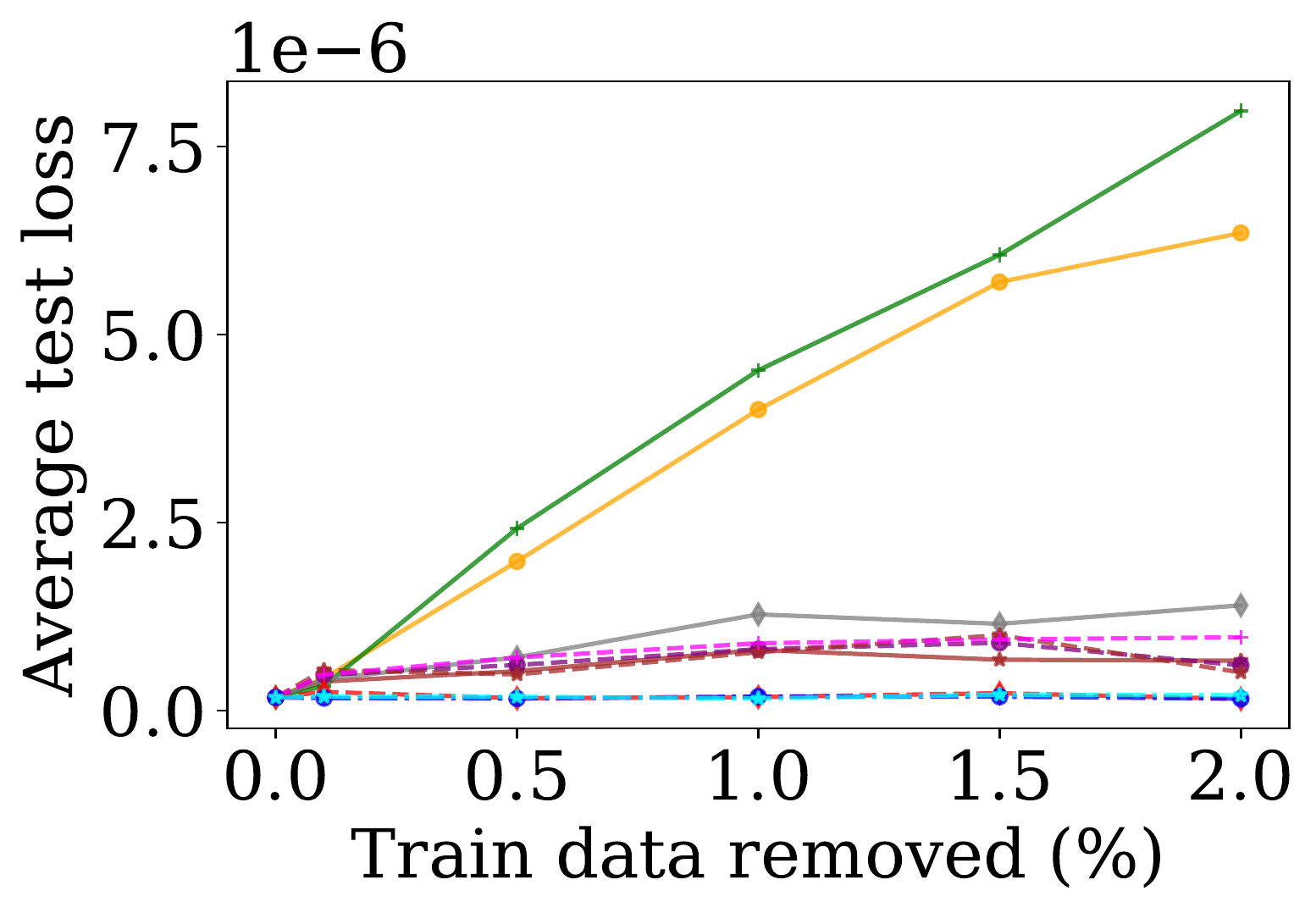}
  \caption{SGB: Naval}
\end{subfigure}
\\
\begin{subfigure}{\tw\textwidth}
  \centering
  \includegraphics[width=1.0\linewidth]{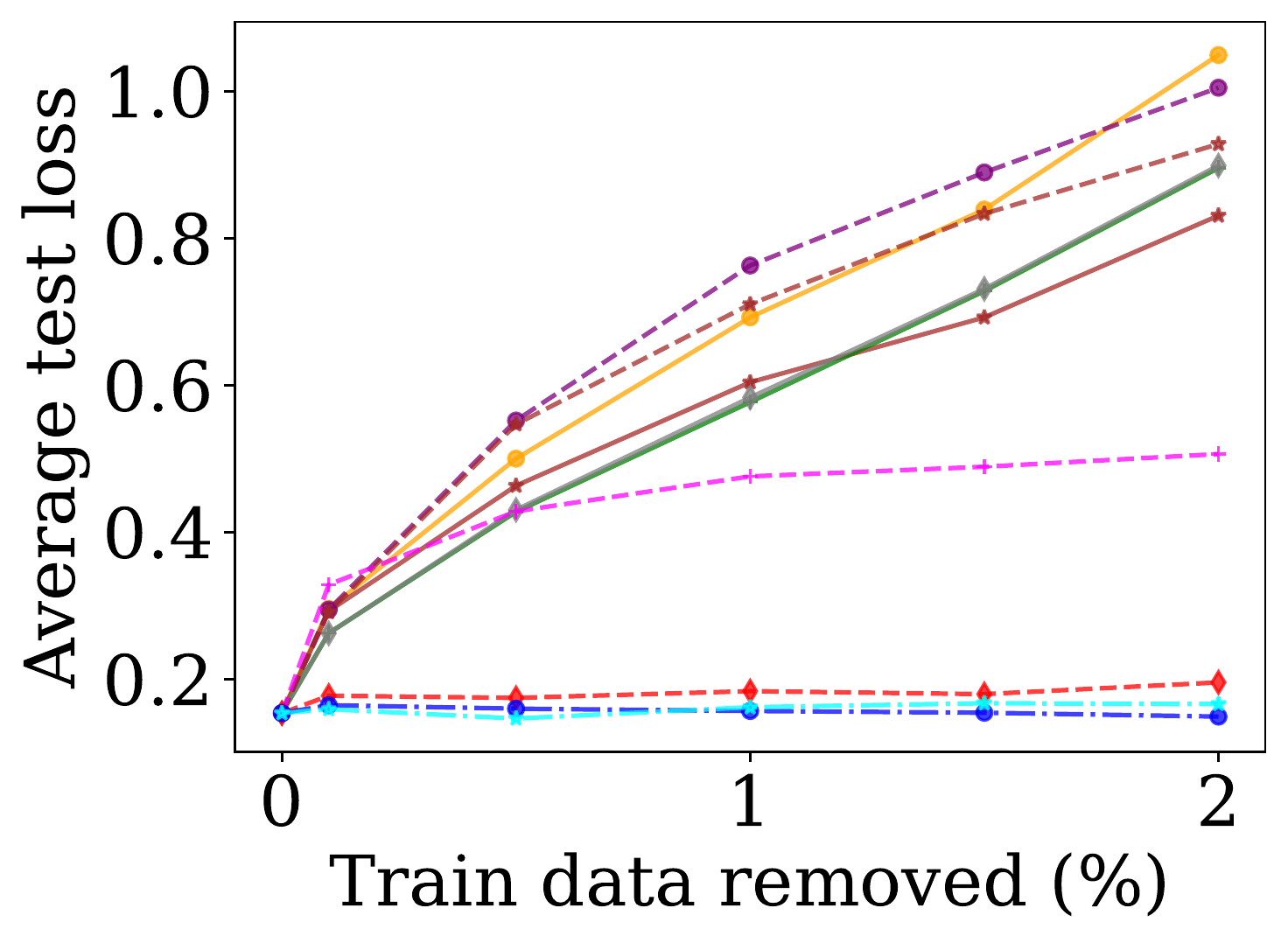}
  \caption{XGB: Bean}
\end{subfigure}
\begin{subfigure}{\tw\textwidth}
  \centering
  \includegraphics[width=1.0\linewidth]{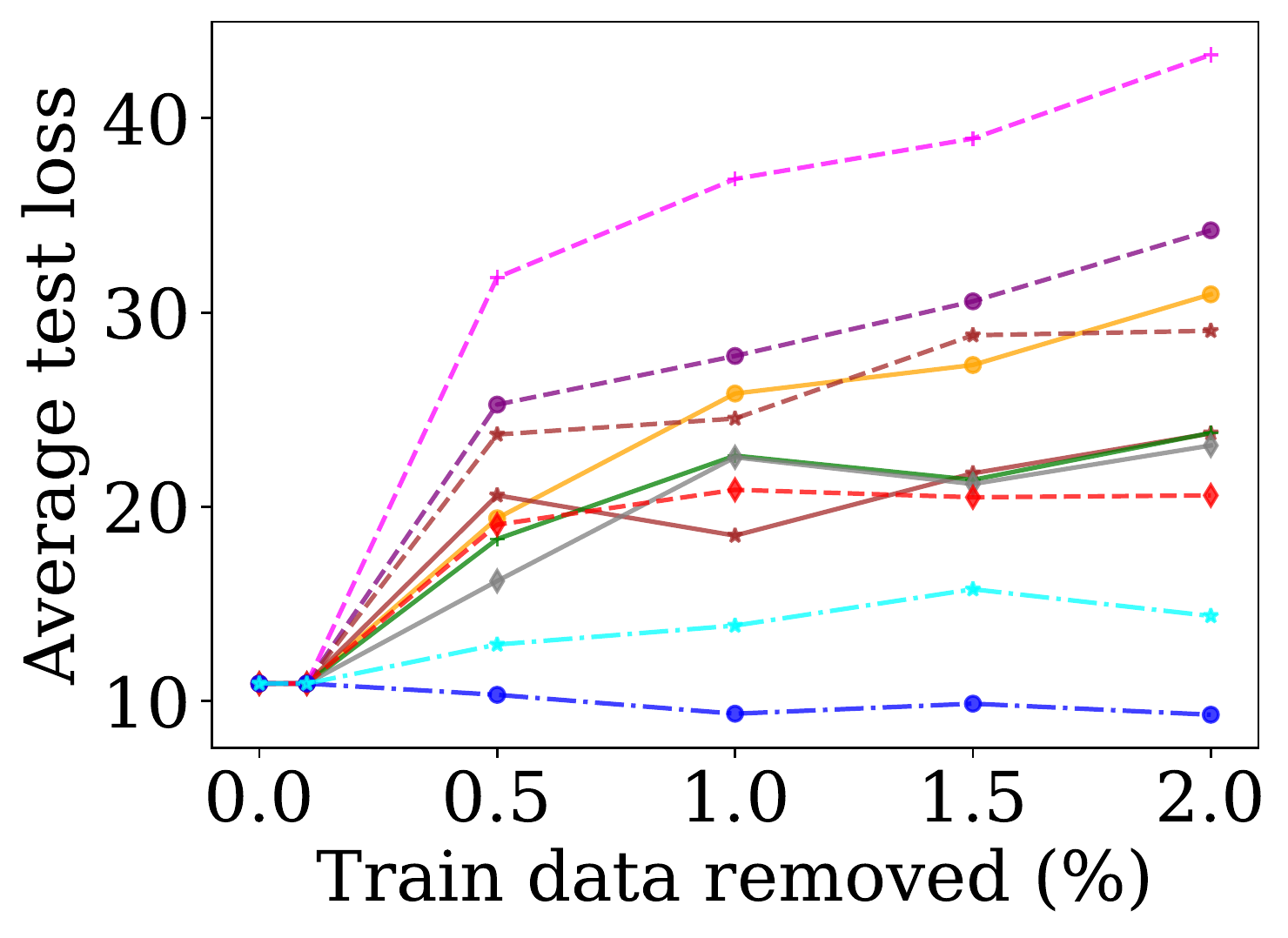}
  \caption{XGB: Concrete}
\end{subfigure}
\begin{subfigure}{\tw\textwidth}
  \centering
  \includegraphics[width=1.0\linewidth]{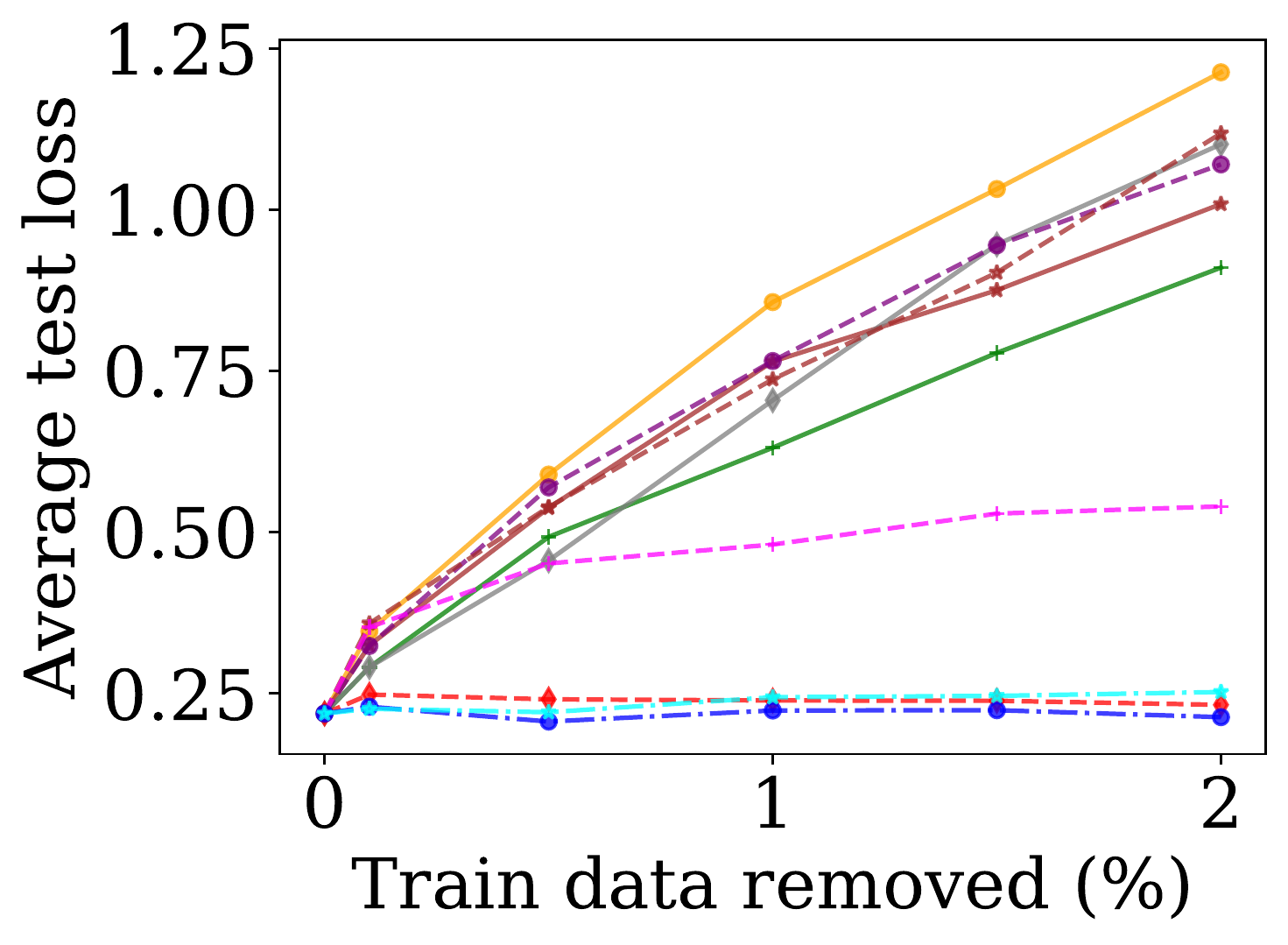}
  \caption{XGB: Surgical}
\end{subfigure}
\\
\begin{subfigure}{\tw\textwidth}
  \centering
  \includegraphics[width=1.0\linewidth]{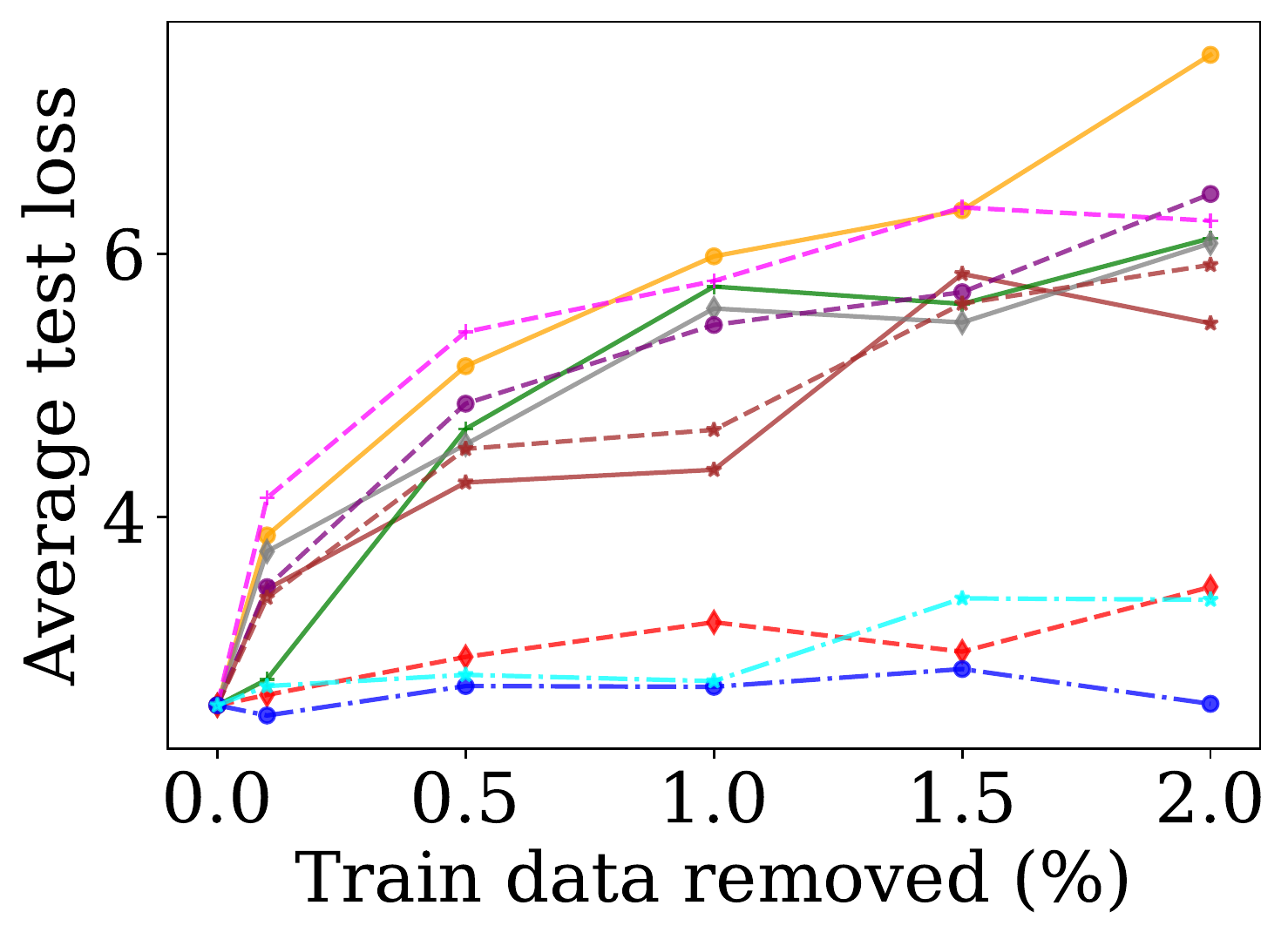}
  \caption{CB: Life}
\end{subfigure}
\begin{subfigure}{\tw\textwidth}
  \centering
  \includegraphics[width=1.0\linewidth]{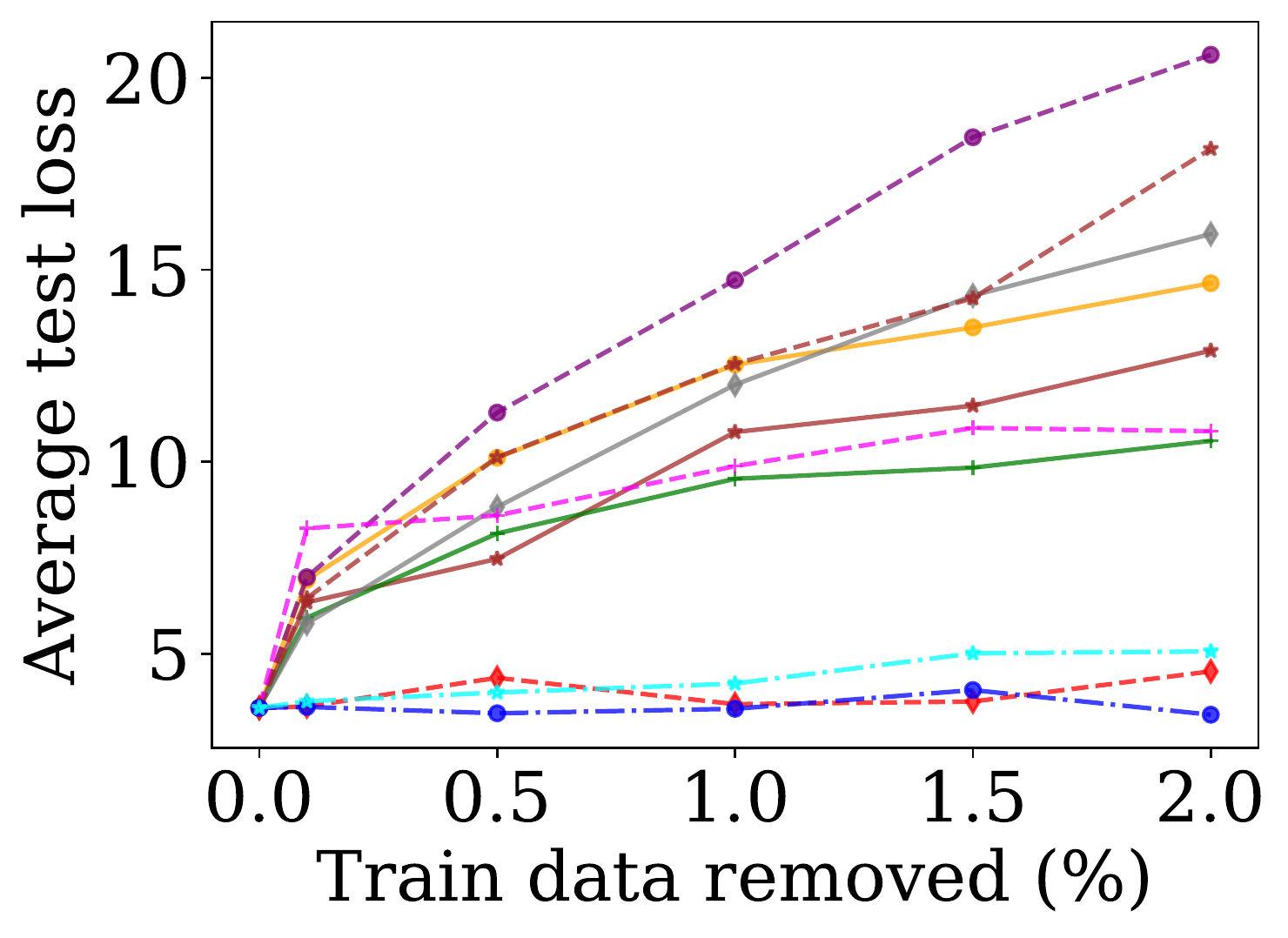}
  \caption{CB: Power}
\end{subfigure}
\begin{subfigure}{\tw\textwidth}
  \centering
  \includegraphics[width=1.0\linewidth]{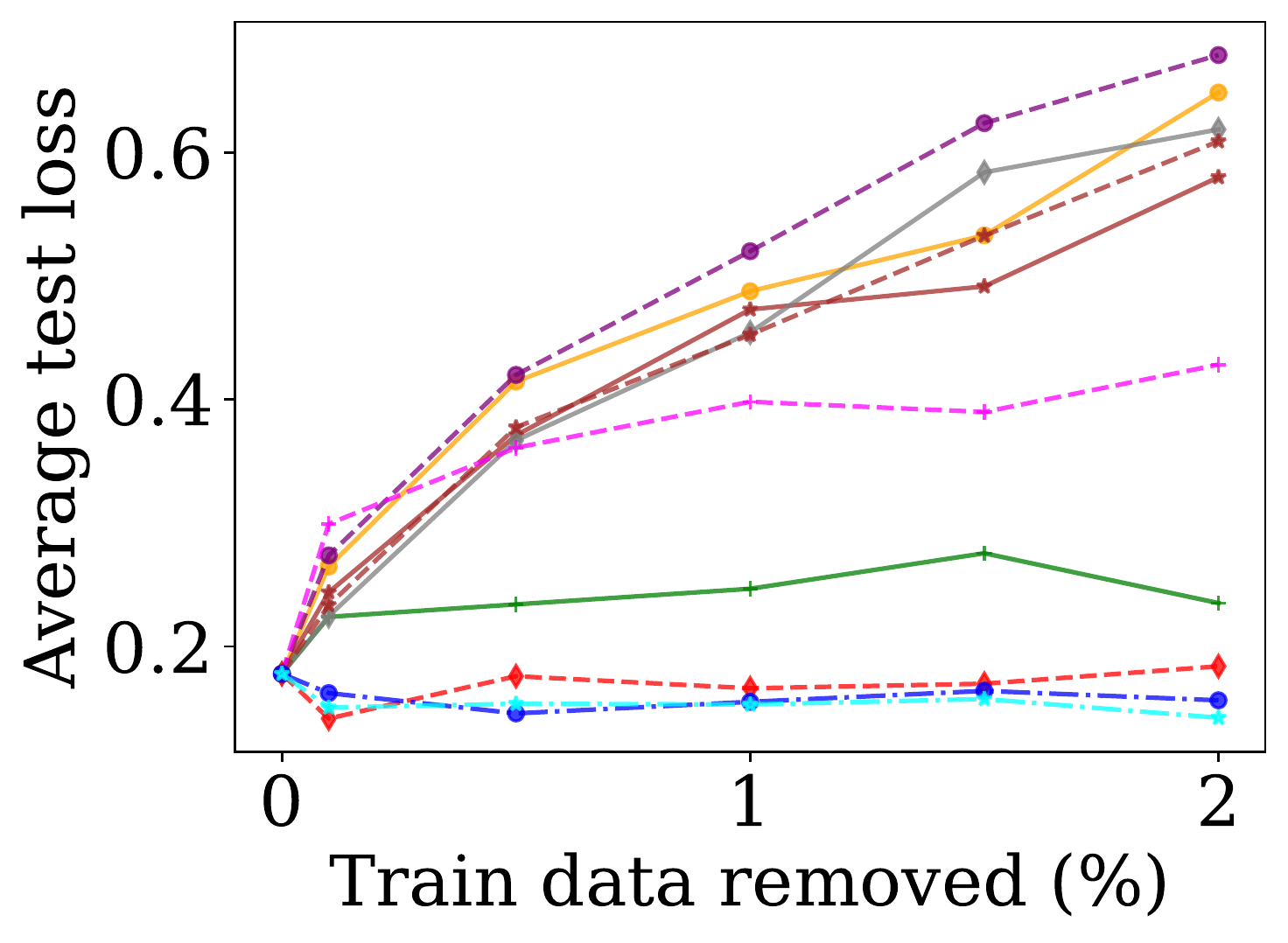}
  \caption{CB: Wine}
\end{subfigure}
\caption{Change in loss for a random test example~(averaged over 100~runs) as training examples are removed. Higher is better.}
\label{app_fig:single_remove_individual}
\end{figure}

\newpage
\hphantom{dummy}
\newpage

\subsection{Targeted Label Edits (Single Test): Additional Analysis}
\label{app_sec:single_test_targeted_edit}

Figure~\ref{app_fig:single_target_edit_loss_rank} shows more fine-grained ranking analysis for the targeted-label-edit experiment involving a single test instance. The trends are relatively consistent across GBDT types with LeafRefit consistently performing well for the SDS data sets, especially on XGB and CB; for larger data sets, TREX or TreeSim are solid alternative choices to LeafRefit.

\renewcommand{\tw}{0.24}

\begin{figure}[h]
\centering
\begin{subfigure}{\tw\textwidth}
  \centering
  \includegraphics[width=1.0\linewidth]{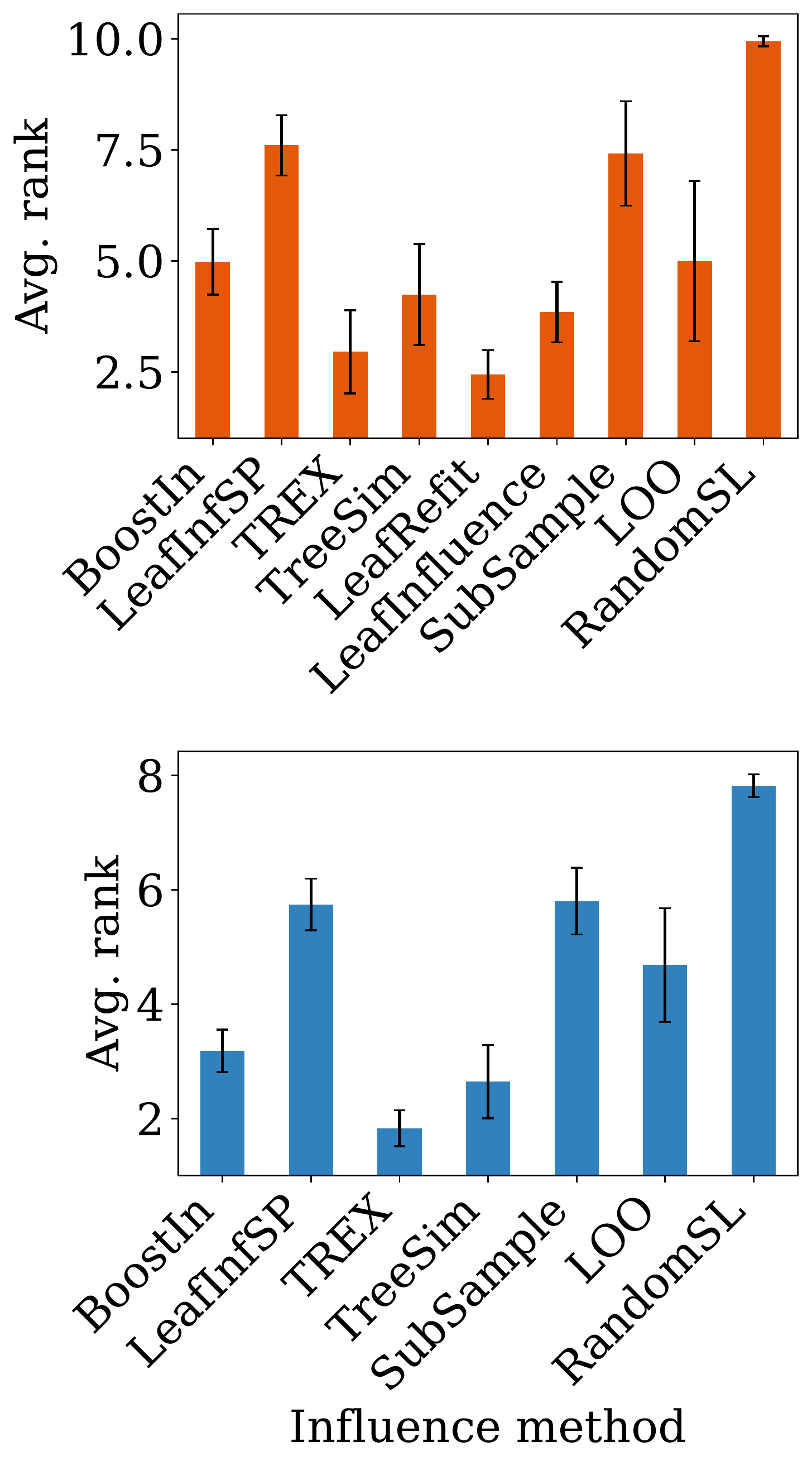}
  \caption{LGB}
  \label{app_fig:target_loss_rank_lgb}
\end{subfigure}
\begin{subfigure}{\tw\textwidth}
  \centering
  \includegraphics[width=1.0\linewidth]{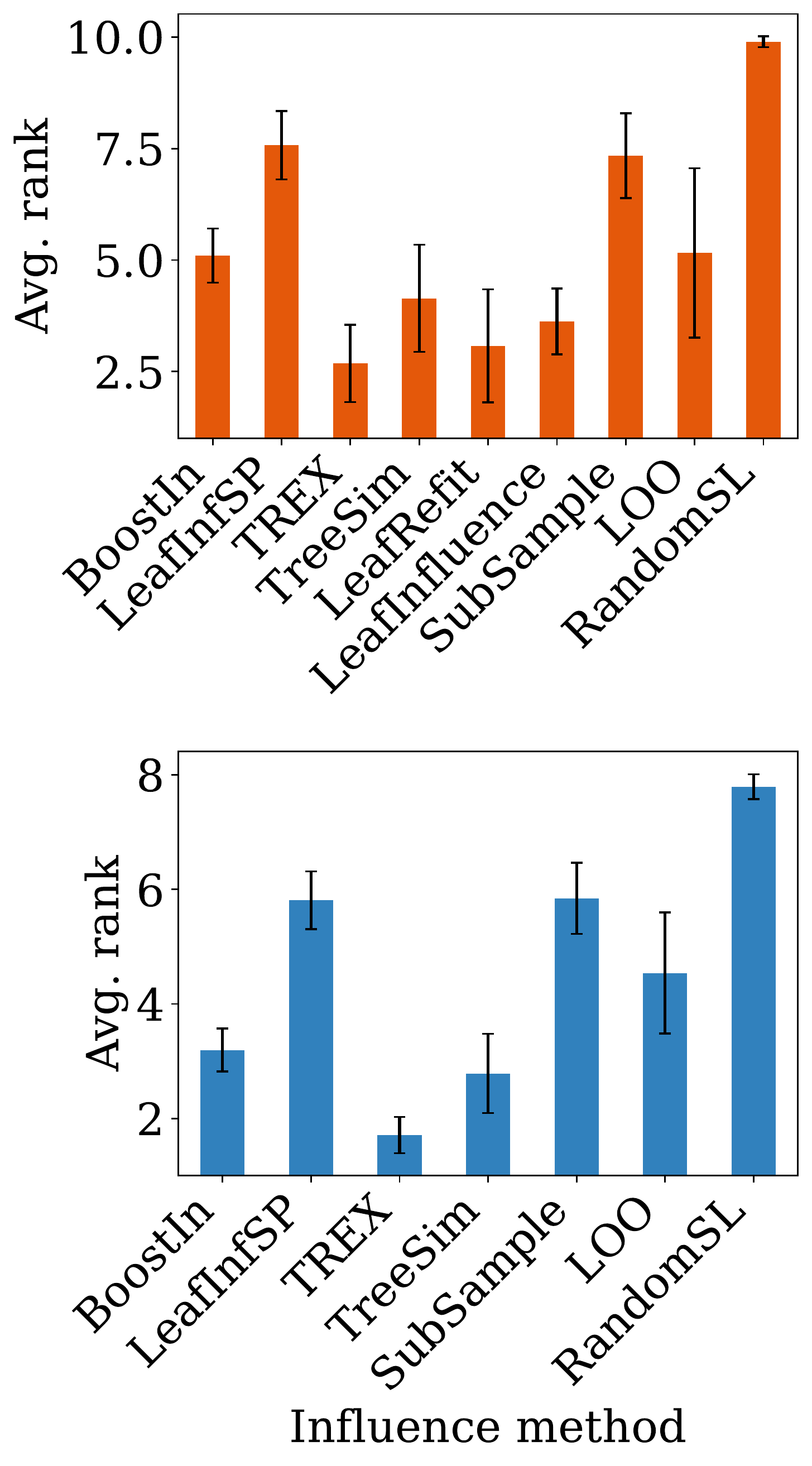}
  \caption{SGB}
  \label{app_fig:target_loss_rank_sgb}
\end{subfigure}
\begin{subfigure}{\tw\textwidth}
  \centering
  \includegraphics[width=1.0\linewidth]{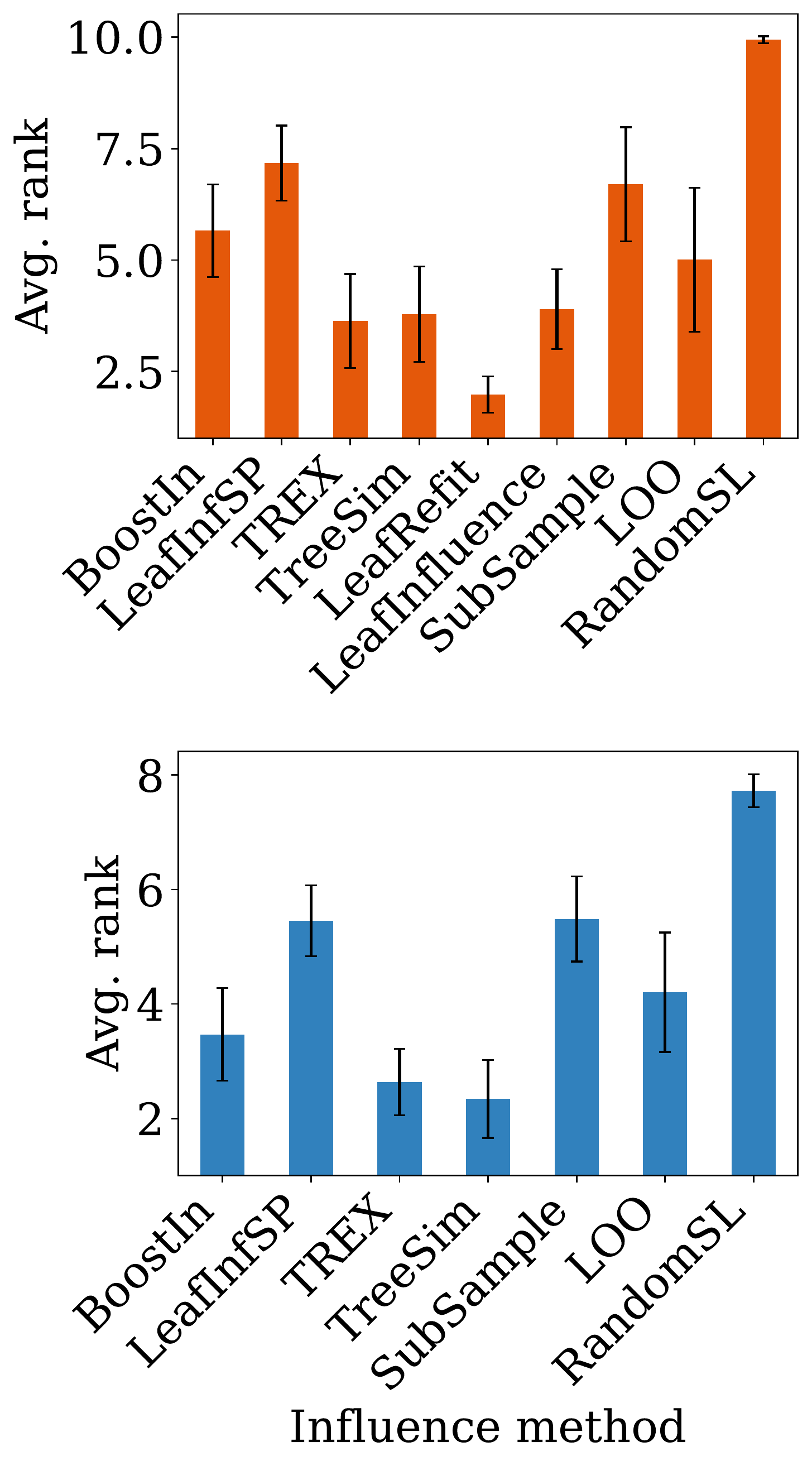}
  \caption{XGB}
  \label{app_fig:target_loss_rank_xgb}
\end{subfigure}
\begin{subfigure}{\tw\textwidth}
  \centering
  \includegraphics[width=1.0\linewidth]{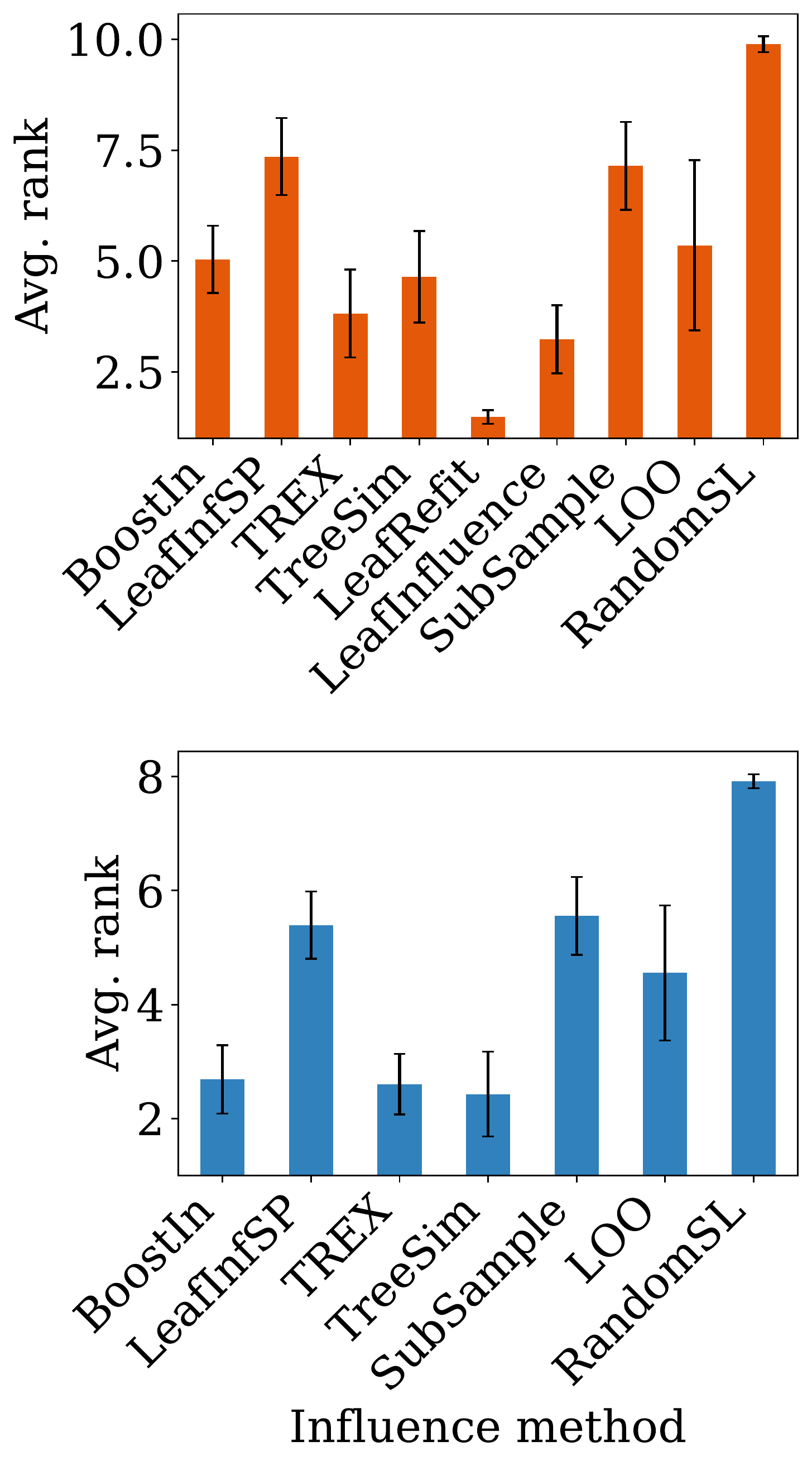}
  \caption{CB}
  \label{app_fig:target_loss_rank_cb}
\end{subfigure}
\caption{Average ranks when editing training labels to a target label for a single test instance, shown for each GBDT type. \emph{Top row}: SDS data sets; \emph{Bottom row}: all data sets. Results are averaged over checkpoints and data sets. Error bars represent 95\% confidence intervals computed over data sets. Lower is better.}
\label{app_fig:single_target_edit_loss_rank}
\end{figure}

\newpage
\subsection{Removing Examples (Multiple Test): Additional Analysis}
\label{app_sec:multi_test_remove}

Figure~\ref{app_fig:multi_remove_rank} shows the rankings when using different predictive performance measures~(e.g., accuracy or AUC) when computing ranks. In contrast to loss, methods rank higher the more they decrease accuracy or AUC as training examples are removed. Overall, BoostIn and LeafInfSP are clear favorites across all three performance metrics.

\renewcommand{\tw}{0.325}

\begin{figure}[h]
\centering
\begin{subfigure}{\tw\textwidth}
  \centering
  \includegraphics[width=1.0\linewidth]{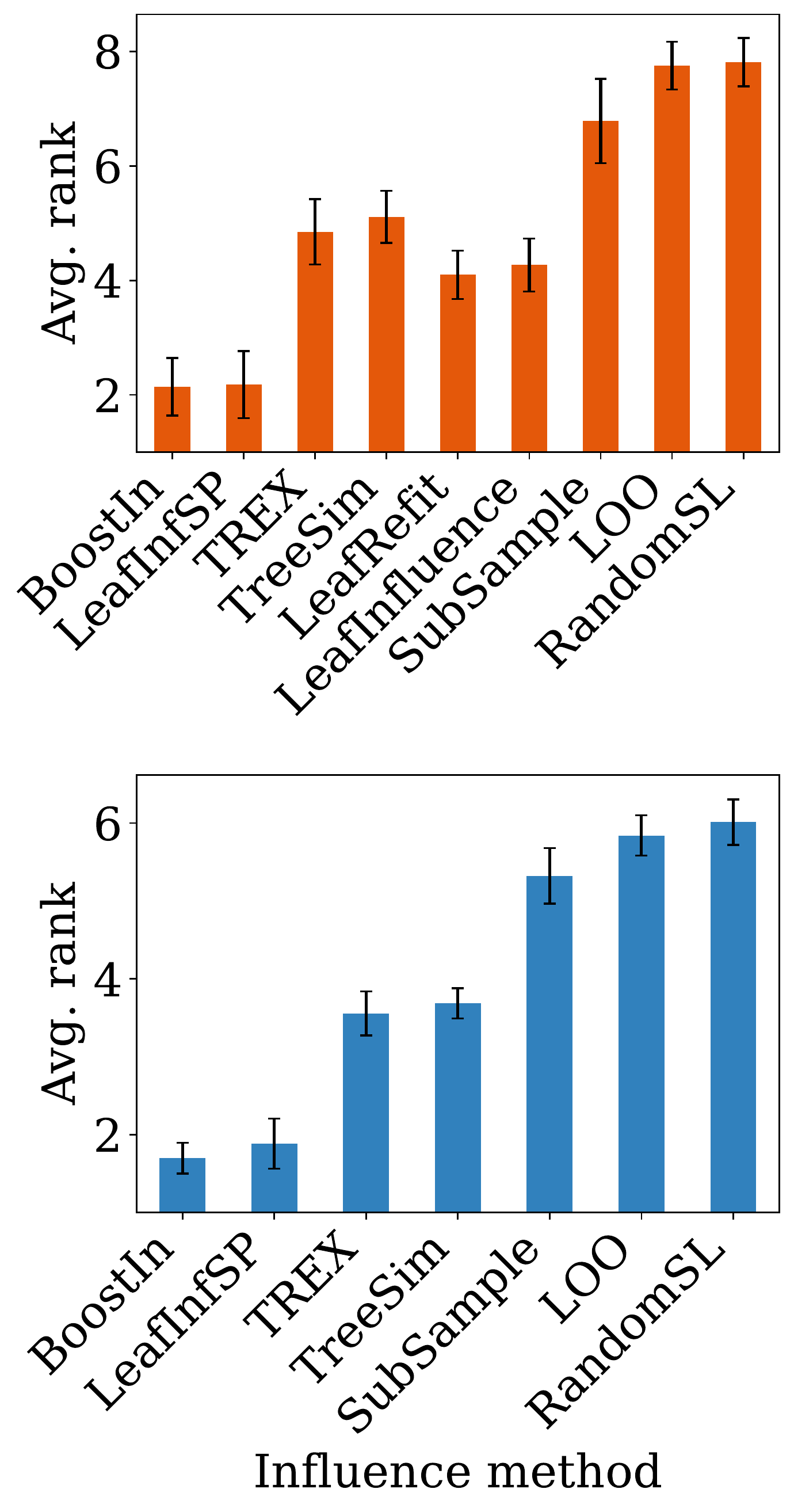}
  \caption{Loss}
  \label{app_fig:multi_remove_loss_rank}
\end{subfigure}
\begin{subfigure}{\tw\textwidth}
  \centering
  \includegraphics[width=1.0\linewidth]{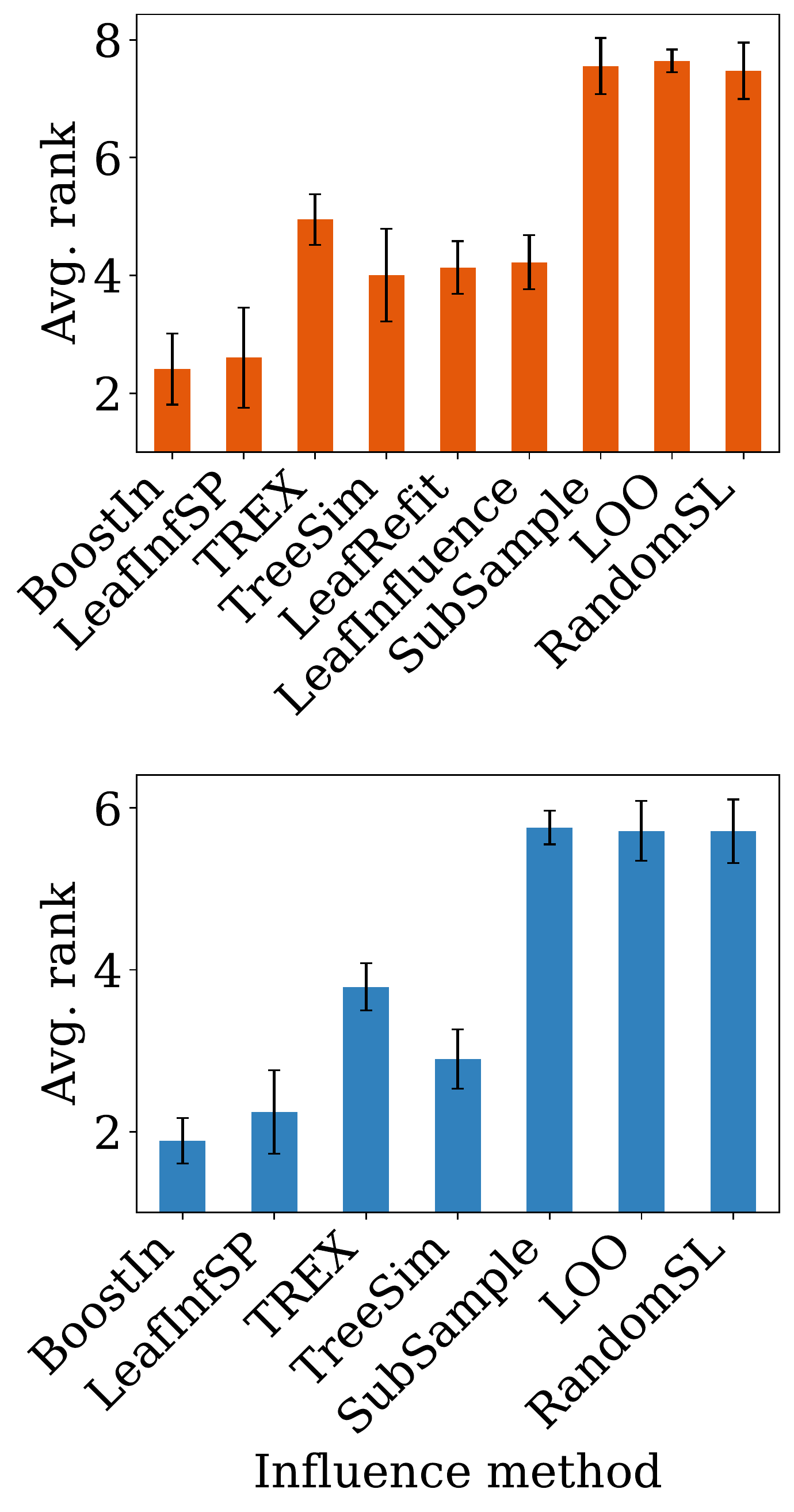}
  \caption{Accuracy}
  \label{app_fig:multi_remove_acc_rank}
\end{subfigure}
\begin{subfigure}{\tw\textwidth}
  \centering
  \includegraphics[width=1.0\linewidth]{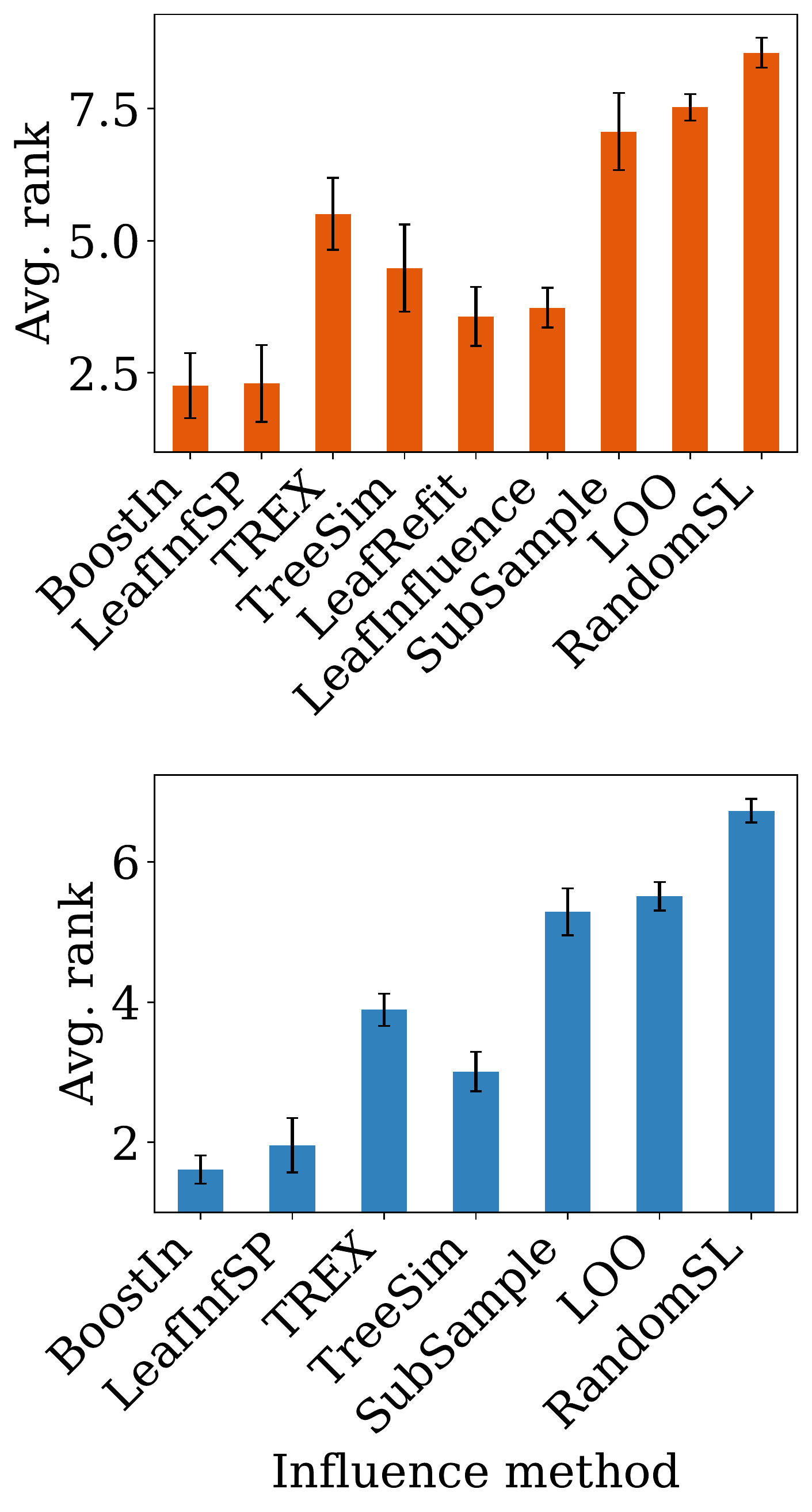}
  \caption{AUC}
  \label{app_fig:multi_remove_auc_rank}
\end{subfigure}
\caption{Average ranks after removing training examples for multiple test instances and evaluating on a held-out test set using different predictive performance metrics. \emph{Top row}: SDS data sets; \emph{Bottom row}: all data sets. Results are averaged over checkpoints, tree types, and data sets. Error bars represent 95\% confidence intervals computed over data sets. Lower is better.}
\label{app_fig:multi_remove_rank}
\end{figure}

\newpage
\subsection{Adding Noise (Multiple Test): Additional Analysis}
\label{app_sec:multi_test_add_noise}

Figure~\ref{app_fig:multi_add_noise_rank} shows the rankings when using different predictive performance measures~(e.g., accuracy or AUC) when computing ranks for the noise addition experiment. In contrast to loss, methods rank higher the more they decrease accuracy or AUC as training examples are removed. Overall, BoostIn is a clear favorite in terms of loss, but both BoostIn and TreeSim perform best for accuracy and AUC.

\renewcommand{\tw}{0.325}

\begin{figure}[h]
\centering
\begin{subfigure}{\tw\textwidth}
  \centering
  \includegraphics[width=1.0\linewidth]{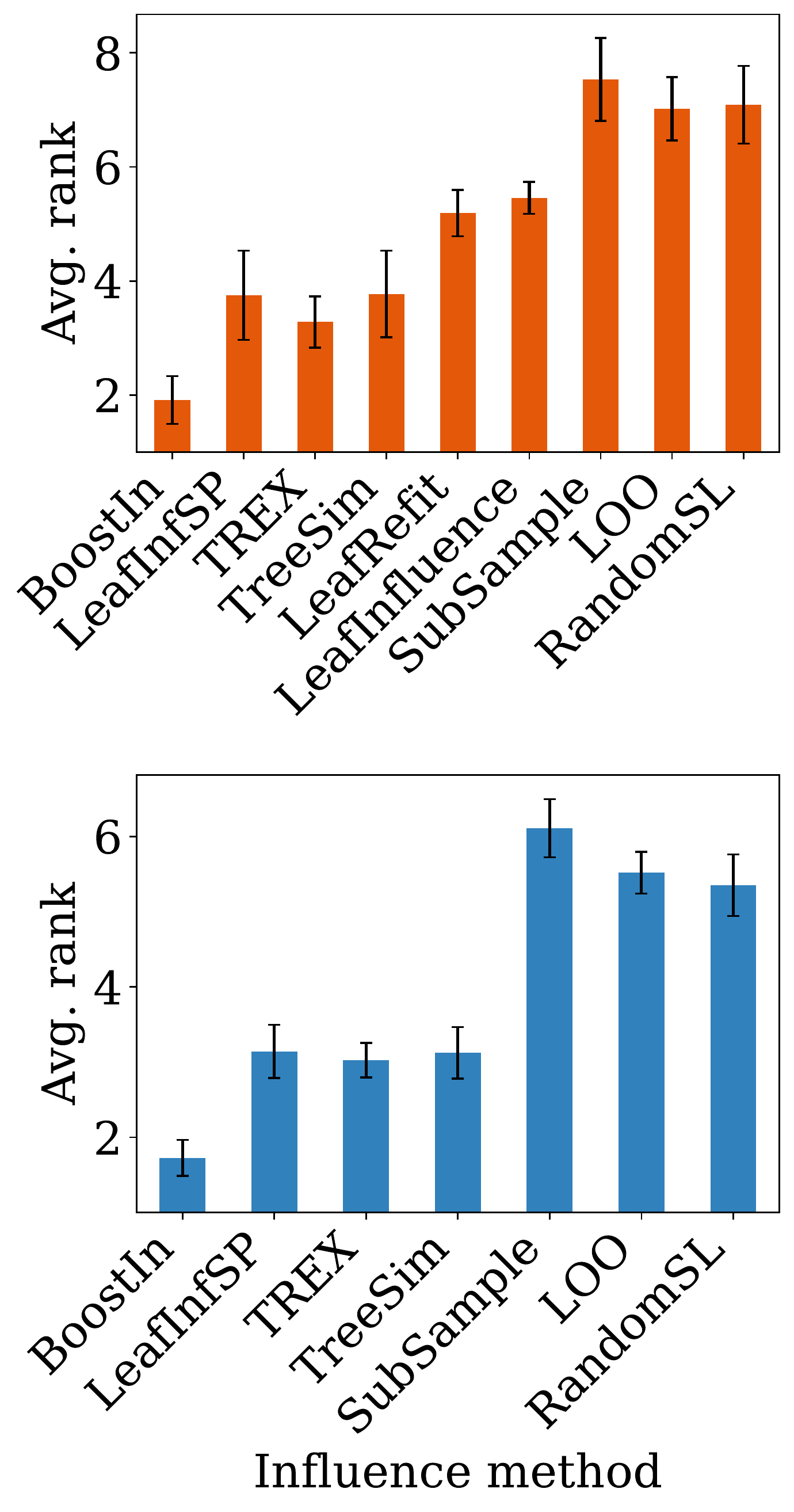}
  \caption{Loss}
  \label{app_fig:multi_add_noise_loss_rank}
\end{subfigure}
\begin{subfigure}{\tw\textwidth}
  \centering
  \includegraphics[width=1.0\linewidth]{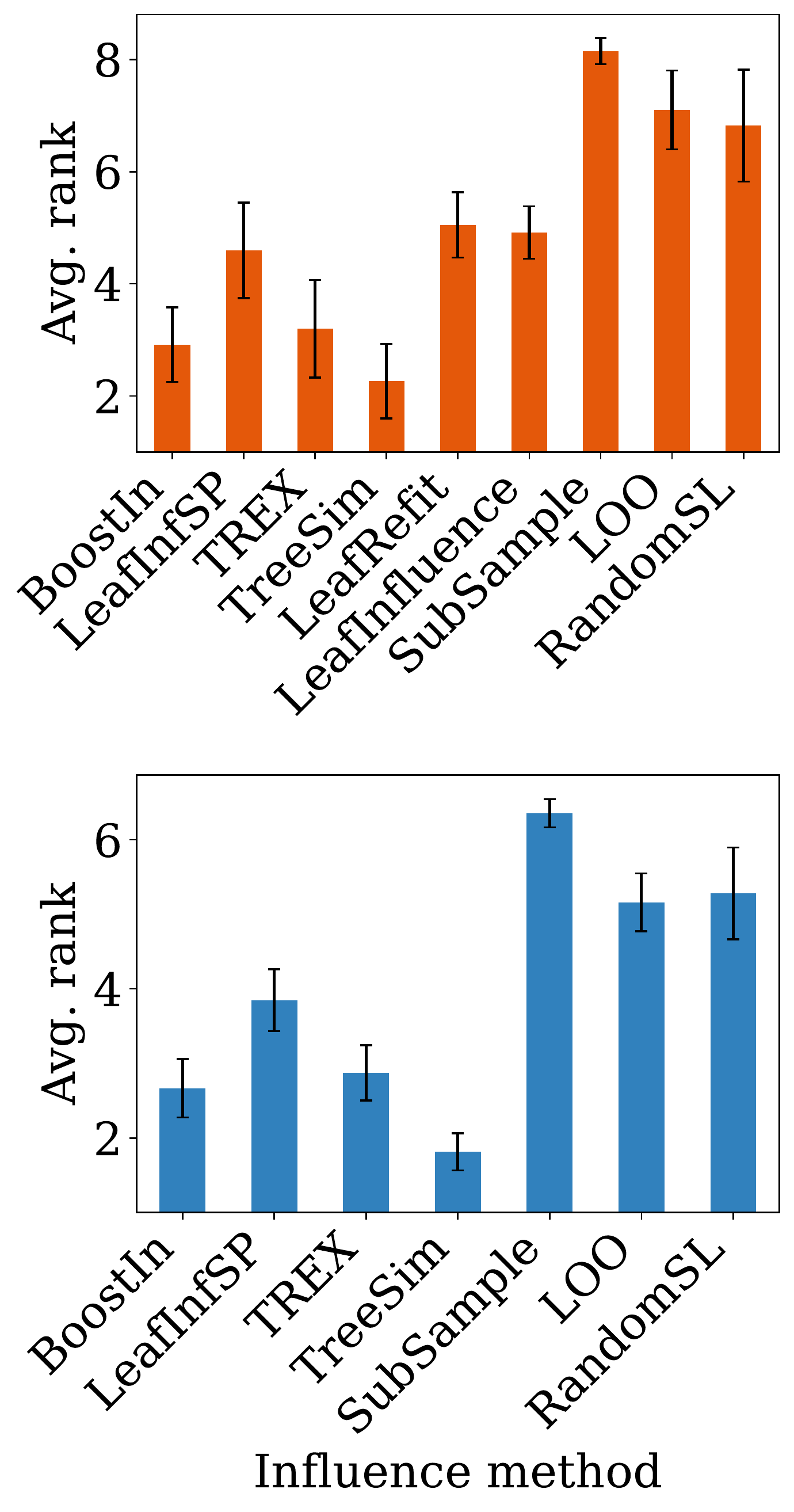}
  \caption{Accuracy}
  \label{app_fig:multi_add_noise_acc_rank}
\end{subfigure}
\begin{subfigure}{\tw\textwidth}
  \centering
  \includegraphics[width=1.0\linewidth]{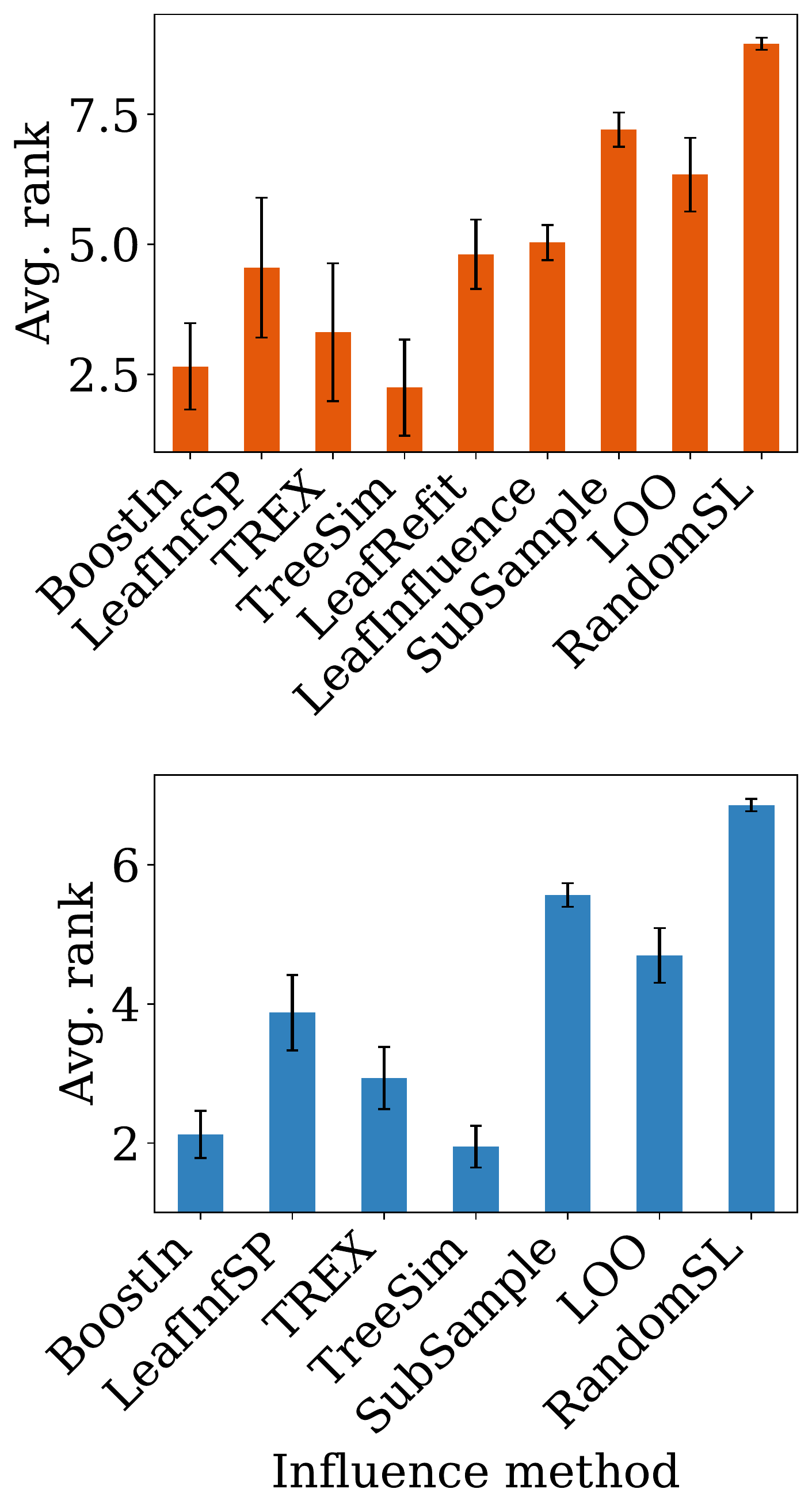}
  \caption{AUC}
  \label{app_fig:multi_add_noise_auc_rank}
\end{subfigure}
\caption{Average ranks after add noise to training examples for multiple test instances and evaluating on a held-out test set using different predictive performance metrics. \emph{Top row}: SDS data sets; \emph{Bottom row}: all data sets. Results are averaged over checkpoints, tree types, and data sets. Error bars represent 95\% confidence intervals computed over data sets. Lower is better.}
\label{app_fig:multi_add_noise_rank}
\end{figure}

\newpage
\subsection{Fixing Noisy/Mislabelled Examples (Multiple Test): Additional Analysis}
\label{app_sec:multi_test_fix_noise}

Figure~\ref{app_fig:multi_fix_noise_rank} shows the average rankings of each method when measuring predictive performance on a held-out test set as noisy/mislabelled training examples are checked and fixed. Methods rank higher the more they decrease loss and increase accuracy or AUC. We also add an additional baseline:~BoostIn~(self), which measures the influence of each training example on itself,~i.e.,~$\mathcal{I}_{BoostIn}(z_i, z_i)$; those values are then used to order the training examples to be checked/fixed. Overall, BoostIn and LeafInfSP rank highest in terms of loss and AUC; however, TREX and TreeSim tend to work better in increasing accuracy.

\renewcommand{\tw}{0.325}

\begin{figure}[h]
\centering
\begin{subfigure}{\tw\textwidth}
  \centering
  \includegraphics[width=1.0\linewidth]{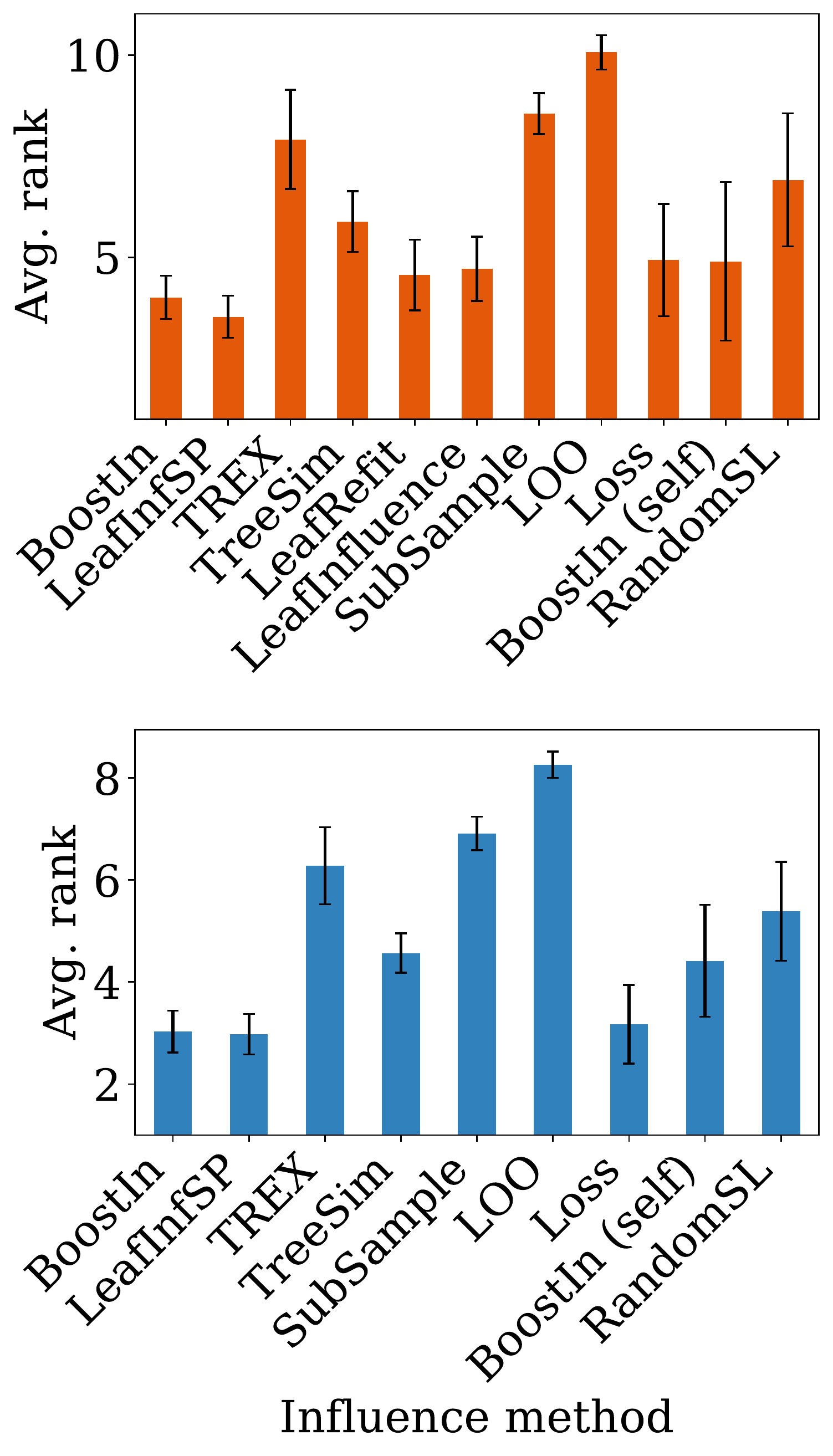}
  \caption{Loss}
  \label{app_fig:multi_fix_noise_loss_rank}
\end{subfigure}
\begin{subfigure}{\tw\textwidth}
  \centering
  \includegraphics[width=1.0\linewidth]{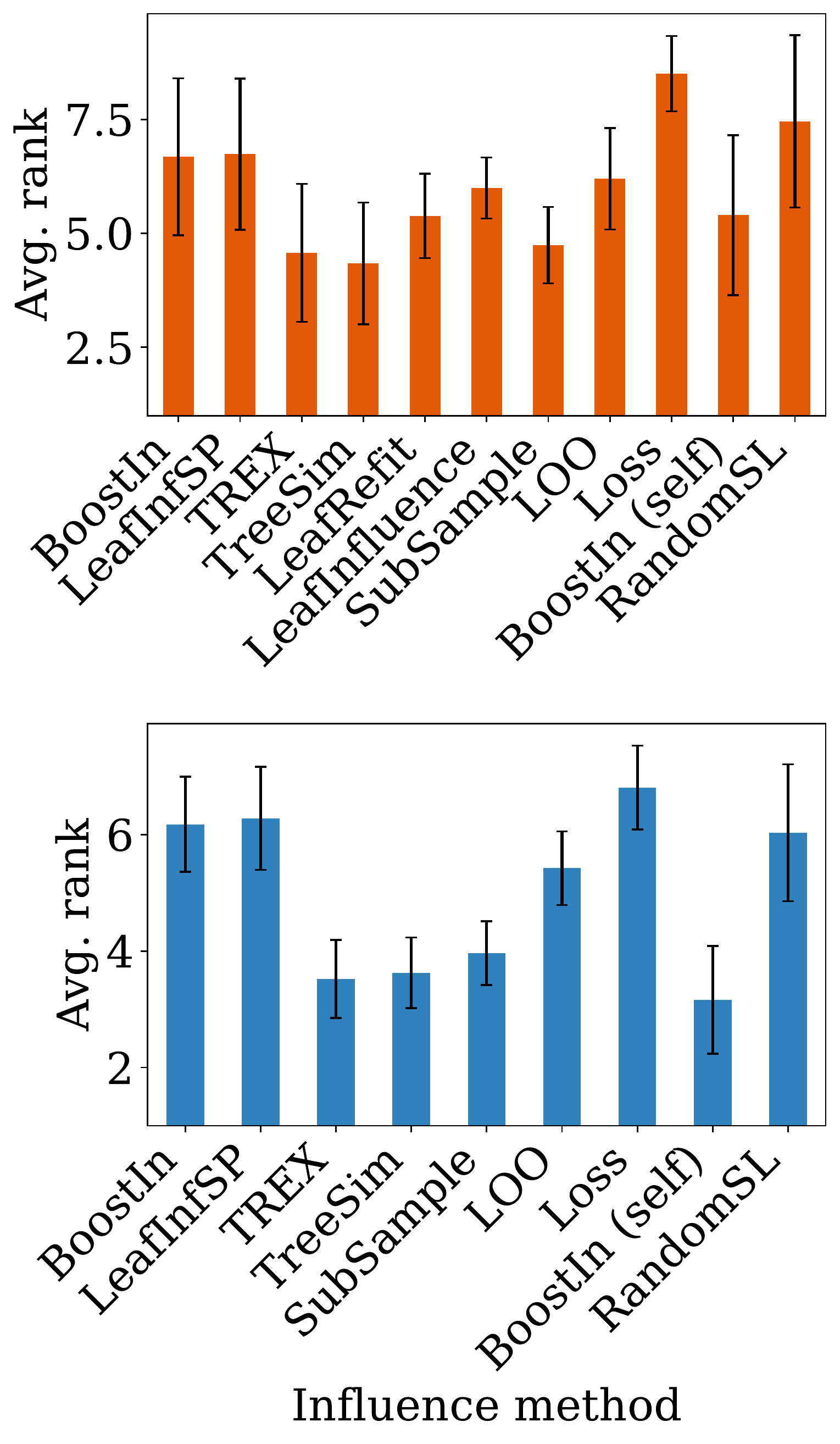}
  \caption{Accuracy}
  \label{app_fig:multi_fix_noise_acc_rank}
\end{subfigure}
\begin{subfigure}{\tw\textwidth}
  \centering
  \includegraphics[width=1.0\linewidth]{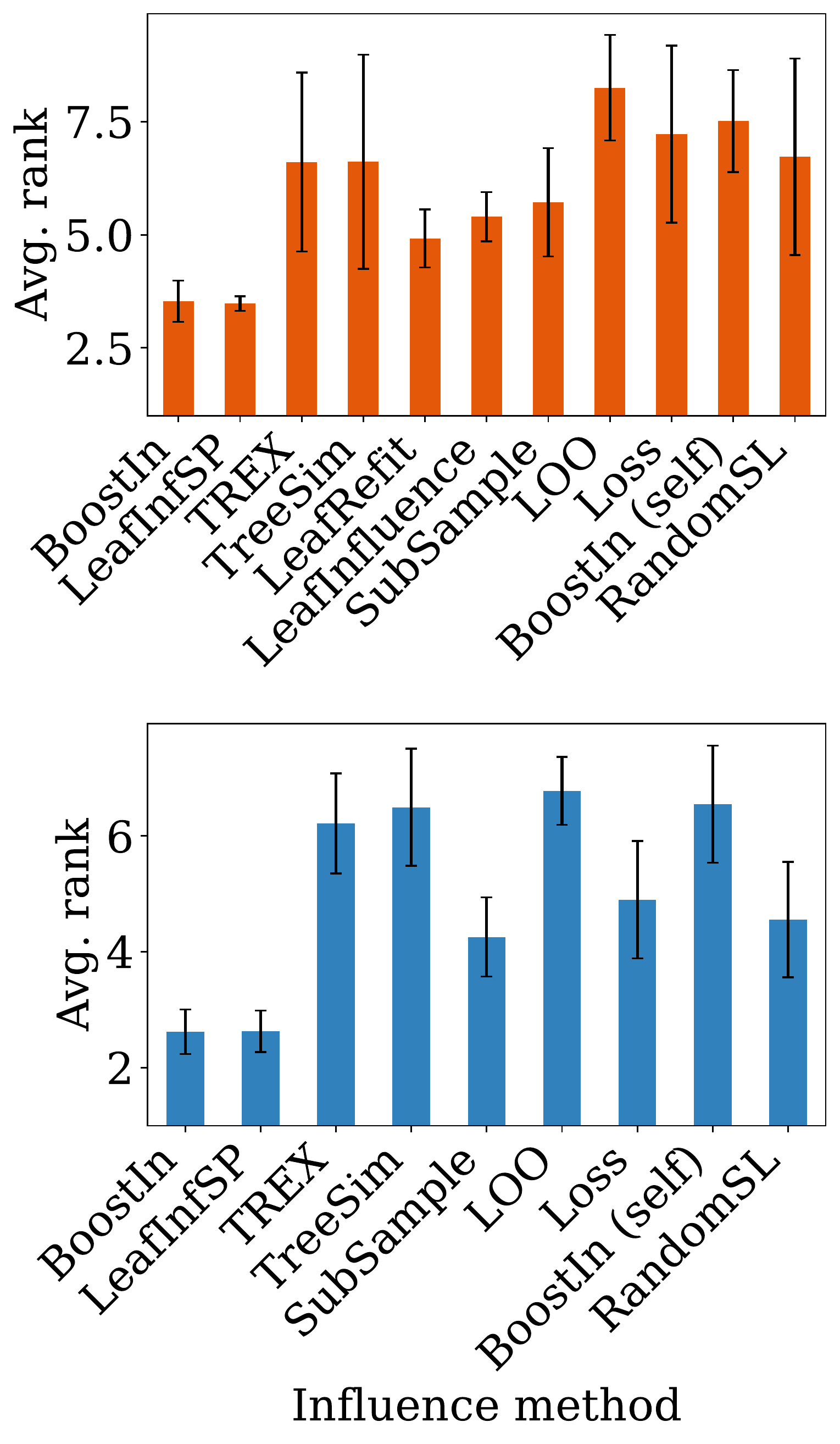}
  \caption{AUC}
  \label{app_fig:multi_fix_noise_auc_rank}
\end{subfigure}
\caption{Average ranks when measuring the predictive performance on the held-out test set after checking/fixing any noisy/mislabelled training examples. \emph{Top row}: SDS data sets; \emph{Bottom row}: all data sets. Results are averaged over checkpoints, tree types, and data sets. Error bars represent 95\% confidence intervals computed over data sets. Lower is better.}
\label{app_fig:multi_fix_noise_rank}
\end{figure}

\newpage
\subsection{Runtime Comparison}
\label{app_sec:runtime}

Table~\ref{app_tab:runtime} shows the total time of each method to compute all influence values for a single test instance (fit time + influence time). Each experiment is repeated 5 times, and results are averaged over GBDT types. Results are only shown for the SDS data sets since LeafRefit and LeafInfluence are too intractable to run on the non-SDS data sets.

\begin{table}[htb]
    \centering
    \begin{tabular}{@{}lcccccccc@{}}
        \toprule
        Data set & TreeSim & BoostIn & LeafInfSP & TREX & SubS. & LOO & LeafRefit & LeafInf. \\
        \midrule
        Bean & 0.260 & 1.021 & 1.362 & 3.999 & 1414 & 4194 & 132503 & 70618 \\
        Compas & 0.027 & 0.110 & 0.139 & 0.598 & 175 & 248 & 2763 & 2302 \\
        Concrete & 0.028 & 0.175 & 0.231 & 0.269 & 469 & 101 & 463 & 346 \\
        Credit & 0.068 & 0.212 & 0.282 & 0.837 & 449 & 3250 & 35359 & 33430 \\
        Energy & 0.024 & 0.164 & 0.248 & 0.260 & 411 & 68 & 317 & 214 \\
        German & 0.005 & 0.031 & 0.043 & 0.420 & 101 & 21 & 70 & 48 \\
        HTRU2 & 0.123 & 0.357 & 0.382 & 0.861 & 506 & 1969 & 43107 & 40499 \\
        Life & 0.107 & 0.445 & 0.596 & 0.936 & 2604 & 1440 & 3414 & 2706 \\
        Naval & 0.291 & 1.140 & 1.358 & 4.390 & 2288 & 5417 & 38599 & 33484 \\
        Power & 0.230 & 1.072 & 1.331 & 6.636 & 1658 & 3390 & 30938 & 26201 \\
        Spambase & 0.121 & 0.522 & 0.659 & 1.201 & 2724 & 3064 & 7935 & 6313 \\
        Surgical & 0.171 & 0.528 & 0.617 & 1.694 & 1170 & 4135 & 36798 & 33455 \\
        Wine & 0.171 & 1.020 & 1.281 & 2.792 & 1862 & 2763 & 14549 & 11419 \\
        \bottomrule
    \end{tabular}
    \caption{Time~(in seconds) to compute all influences values for a single test instance for the SDS data sets. Each experiment is repeated 5 times, and results are averaged over GBDT types.}
    \label{app_tab:runtime}
\end{table}

\newpage
\subsection{Correlation Between Influence Methods: Additional Analysis}
\label{app_sec:correlation}

\begin{figure}[h]
\centering
\begin{subfigure}{0.4\textwidth}
  \centering
  \includegraphics[width=1.0\linewidth]{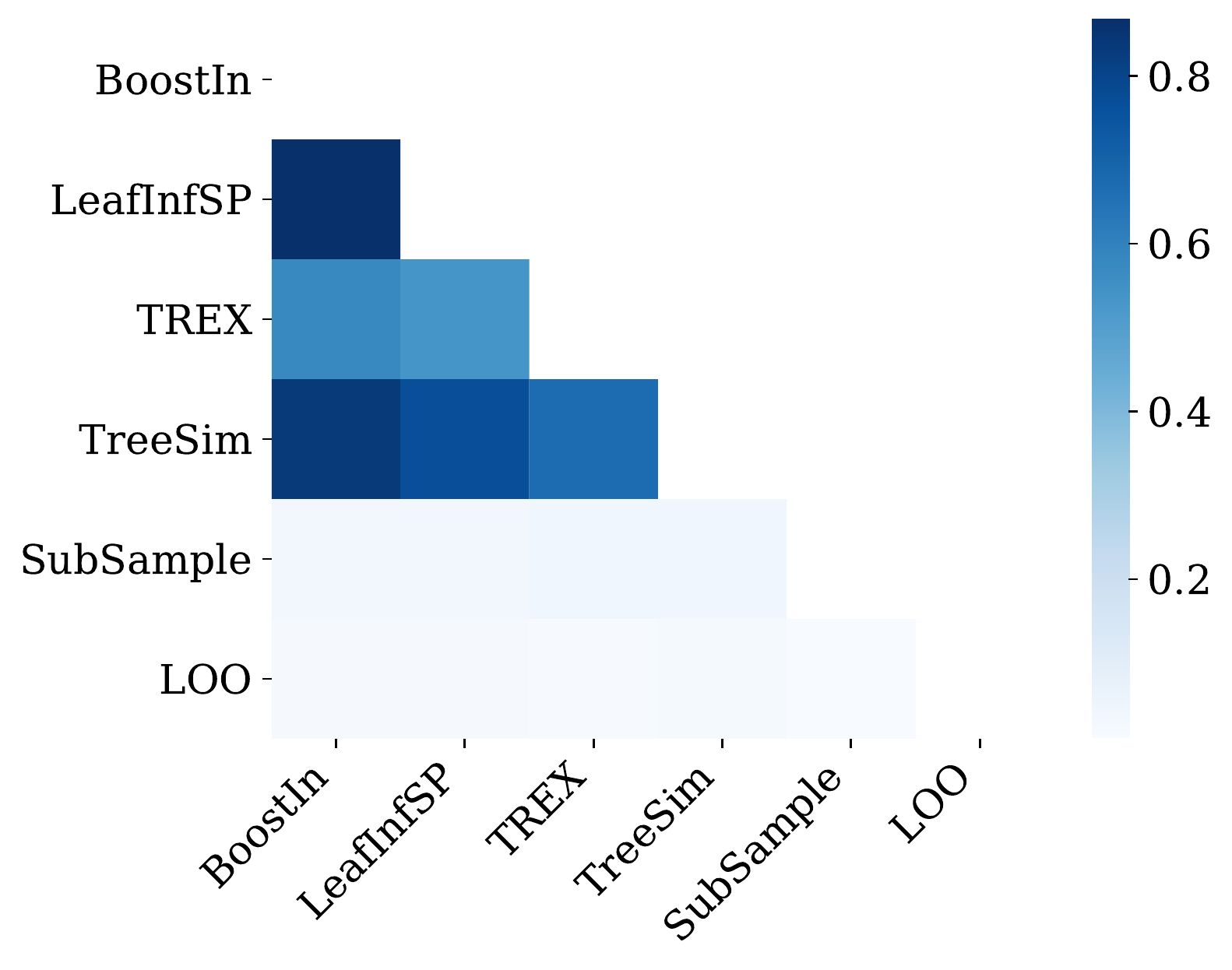}
  \caption{Classification data sets}
  \label{fig:spearman_classification}
\end{subfigure}%
\begin{subfigure}{0.4\textwidth}
  \centering
  \includegraphics[width=1.0\linewidth]{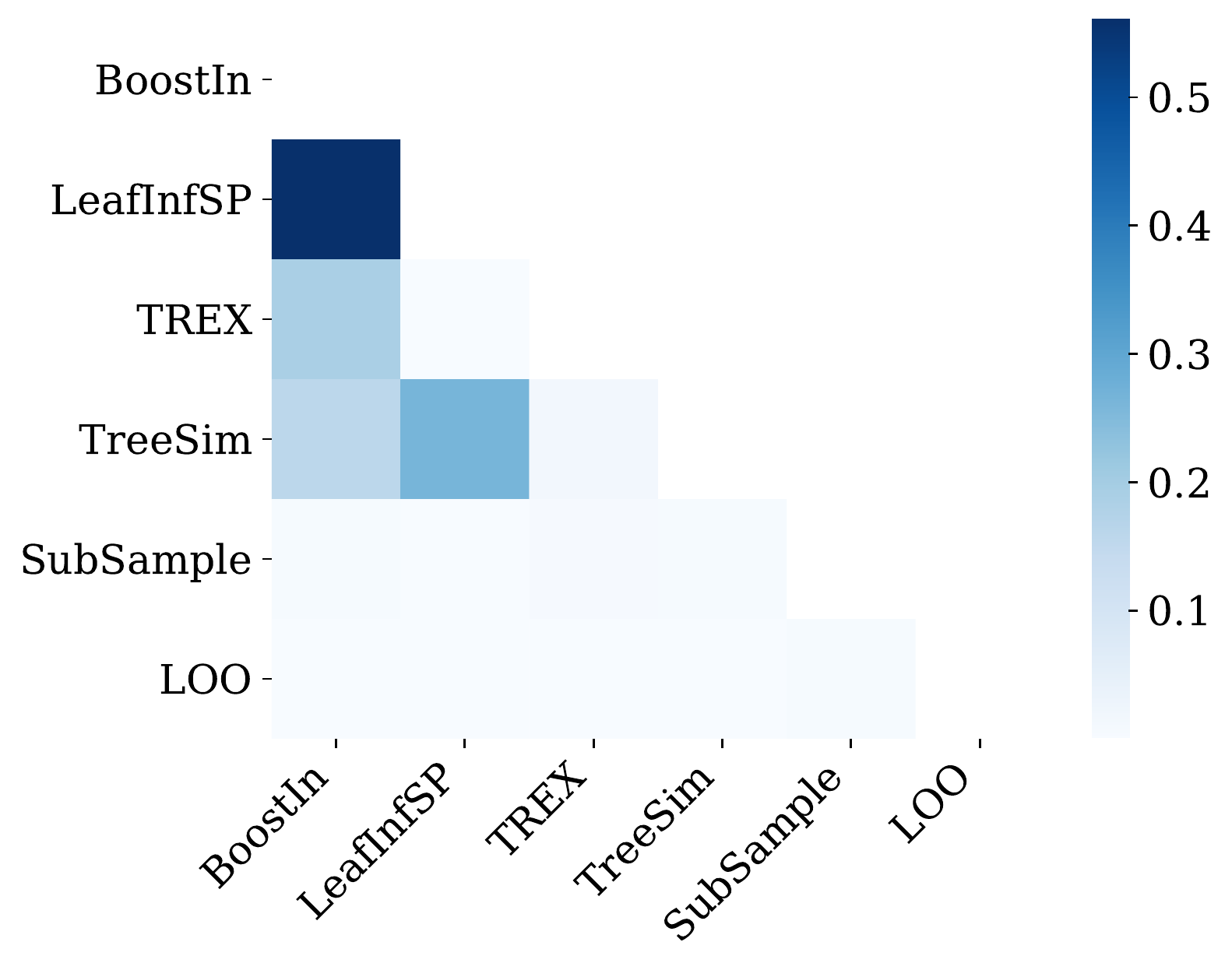}
  \caption{Regression data sets}
  \label{fig:spearman_regression}
\end{subfigure}
\caption{Spearman correlation coefficient between influence methods averaged over 100 test examples, all GBDT types, and either classification or regression data sets.}
\label{app_fig:correlation_datasets}
\end{figure}

Figure~\ref{app_fig:correlation_gbdt_types} shows additional correlation heatmaps averaged over the SDS data sets for each GBDT type; overall, the trends remain the same across GBDT types. Figure~\ref{app_fig:correlation_datasets} shows the correlation between influence methods averaged over either all classification or regression data sets; overall, the methods are much less correlated for the regression data sets than the classification data sets~(note the difference in legend values in both subplots).

\renewcommand{\tw}{0.24}

\begin{figure}[h]
\centering
\begin{subfigure}{\tw\textwidth}
  \centering
  \includegraphics[width=1.0\linewidth]{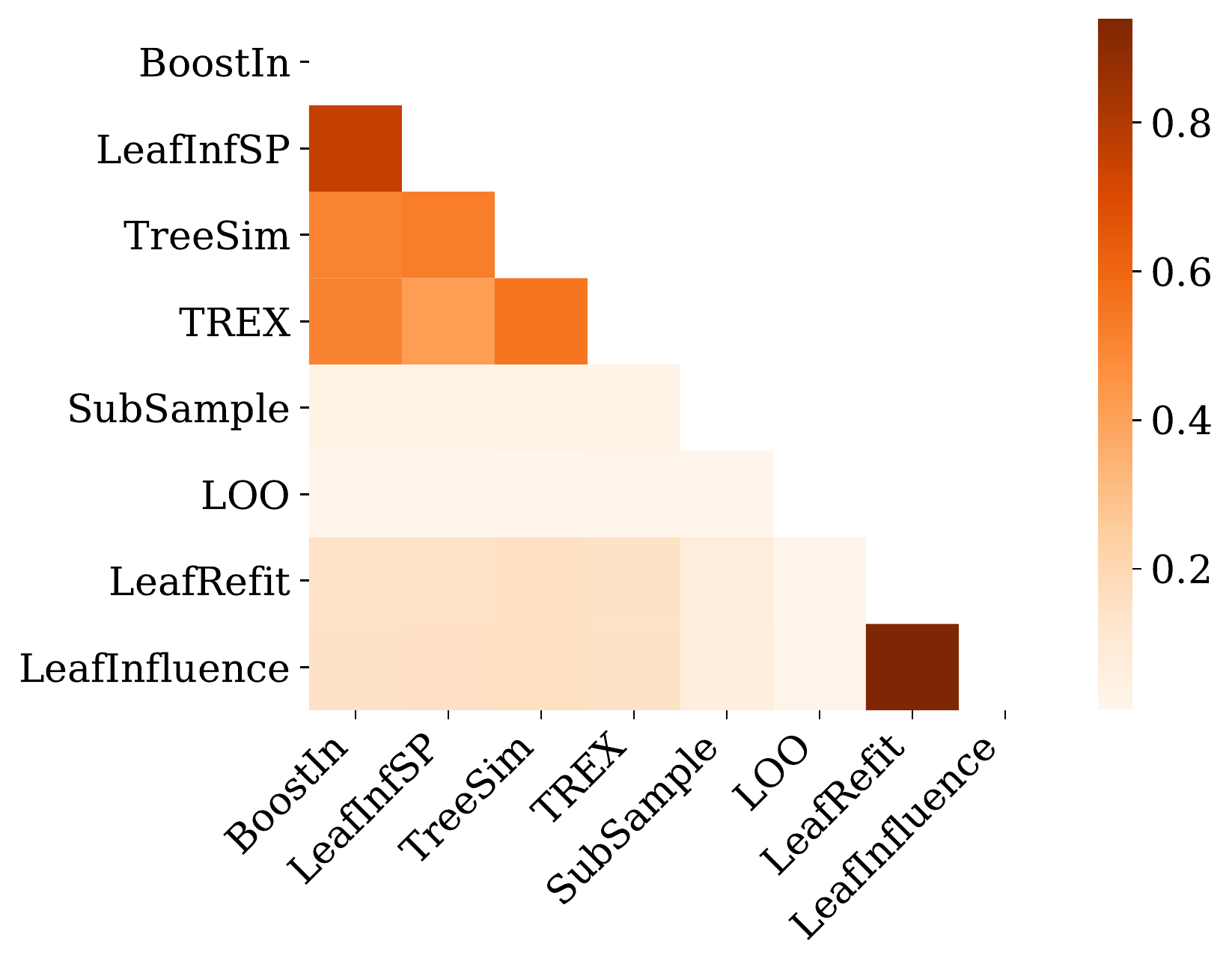}
  \caption{LGB}
  \label{fig:lgb_spearman_li}
\end{subfigure}%
\begin{subfigure}{\tw\textwidth}
  \centering
  \includegraphics[width=1.0\linewidth]{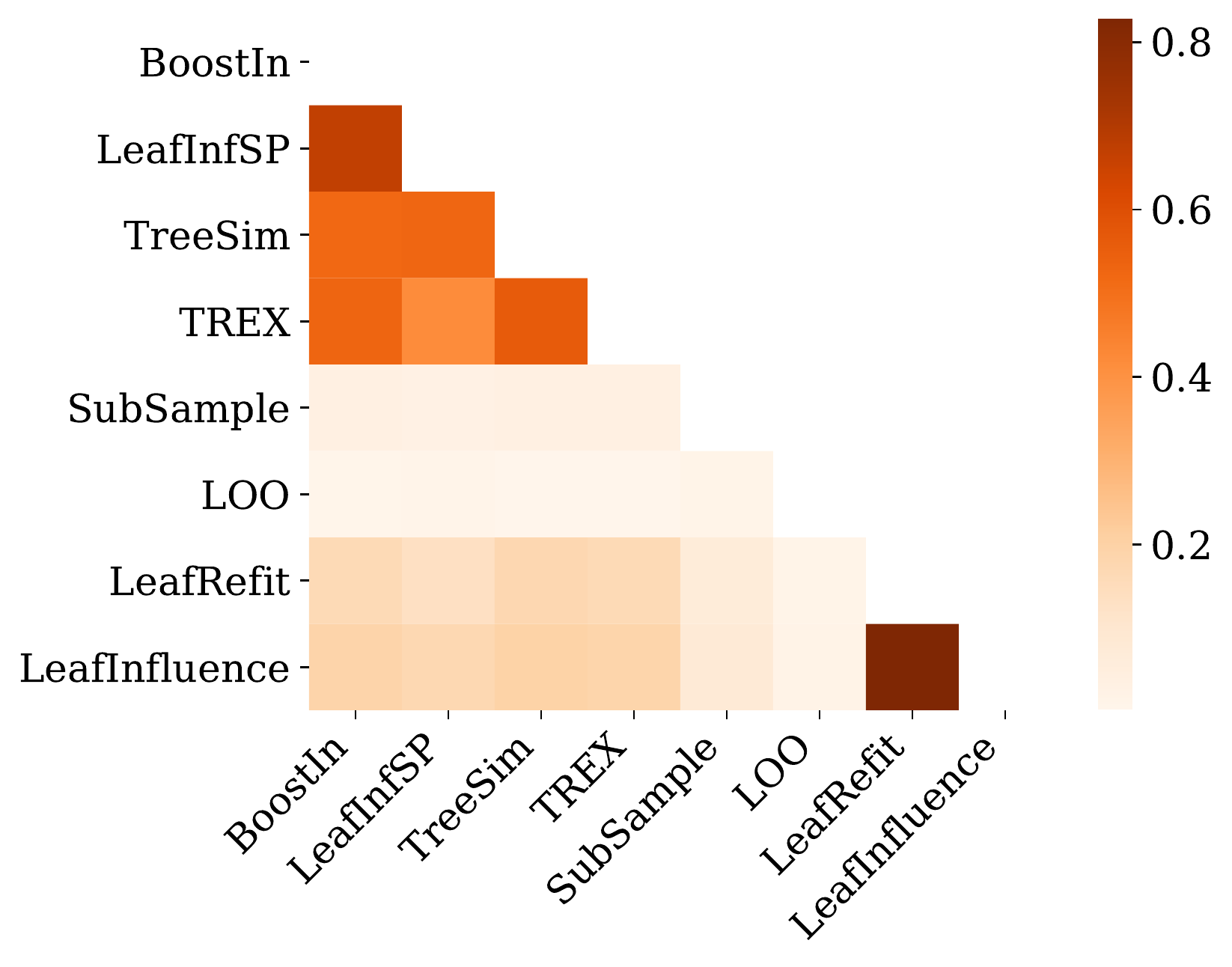}
  \caption{SGB}
  \label{fig:sgb_spearman_li}
\end{subfigure}
\begin{subfigure}{\tw\textwidth}
  \centering
  \includegraphics[width=1.0\linewidth]{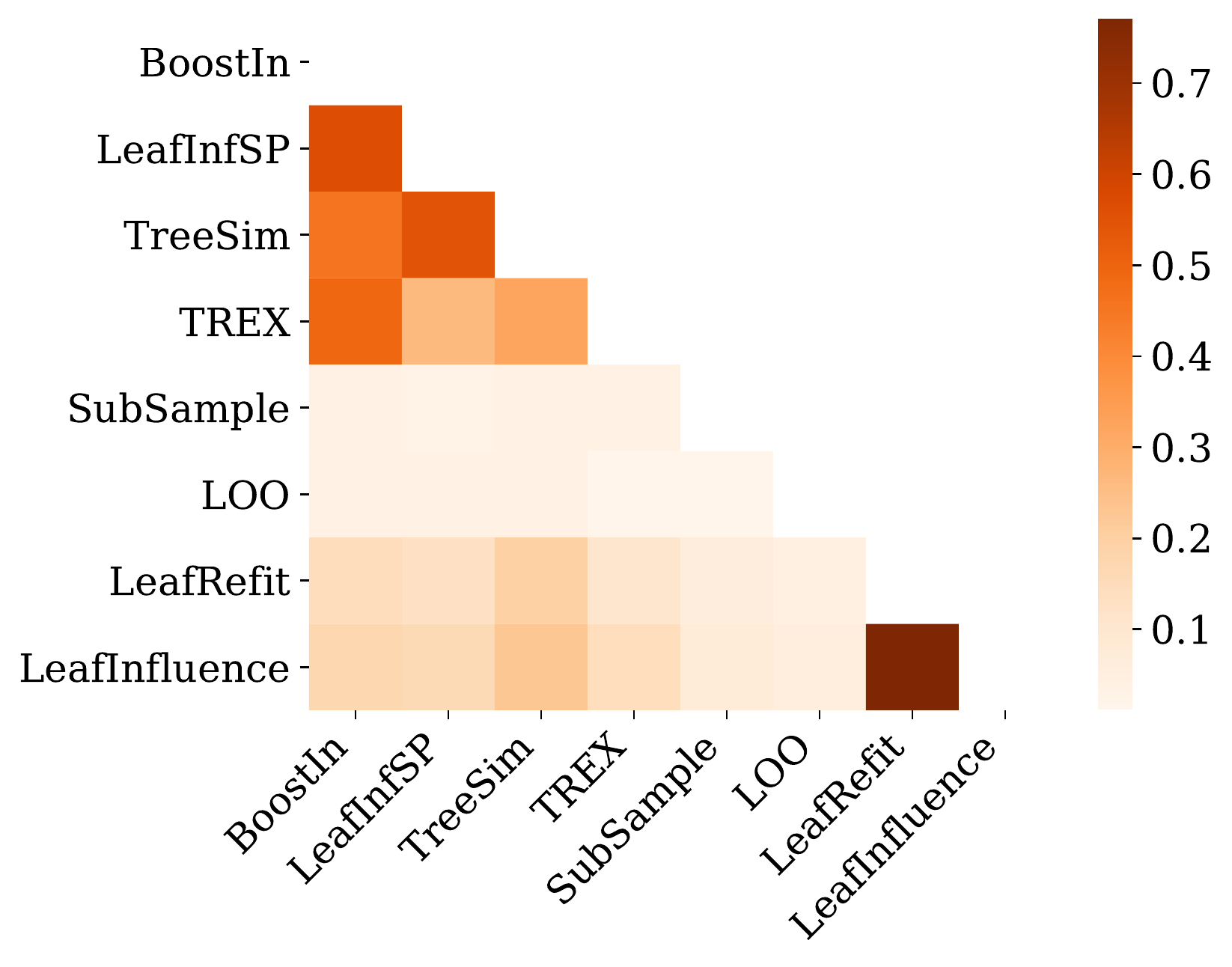}
  \caption{XGB}
  \label{fig:xgb_spearman_li}
\end{subfigure}
\begin{subfigure}{\tw\textwidth}
  \centering
  \includegraphics[width=1.0\linewidth]{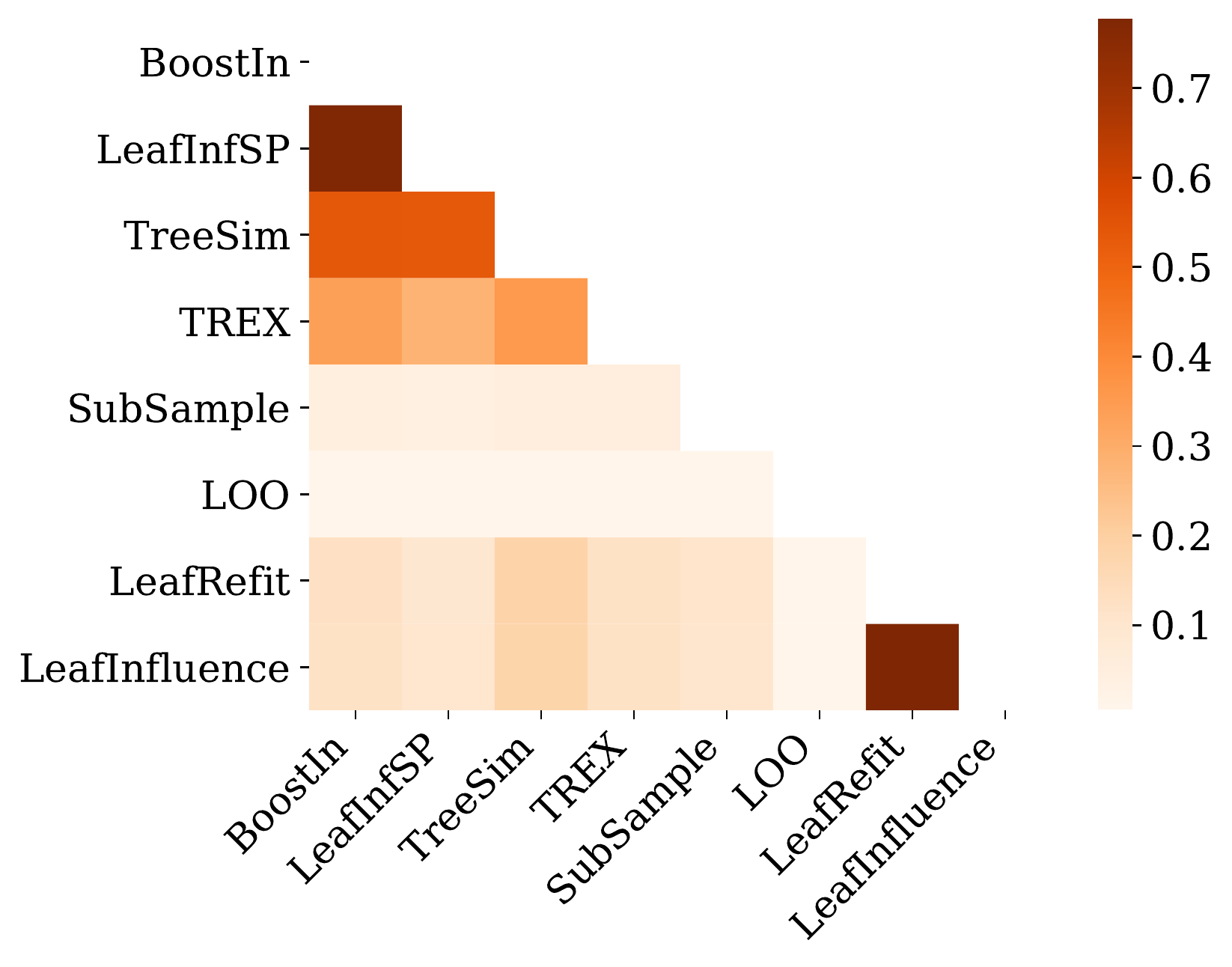}
  \caption{CB}
  \label{fig:cb_spearman_li}
\end{subfigure}
\caption{Spearman correlation coefficient between influence methods for each GBDT type, averaged over 100 test examples and SDS data sets.}
\label{app_fig:correlation_gbdt_types}
\end{figure}

\newpage

\subsection{The Structural Fragility of LOO: Additional Examples}
\label{app_sec:loo_structural_fragility}

Figure~\ref{app_fig:loo_fragility} shows additional examples of LOO choosing the \emph{single-most} influential example.

\renewcommand{\tw}{0.335}
\begin{figure}[H]  %
\centering
\begin{subfigure}{\tw\textwidth}
  \centering
  \includegraphics[width=1.0\linewidth]{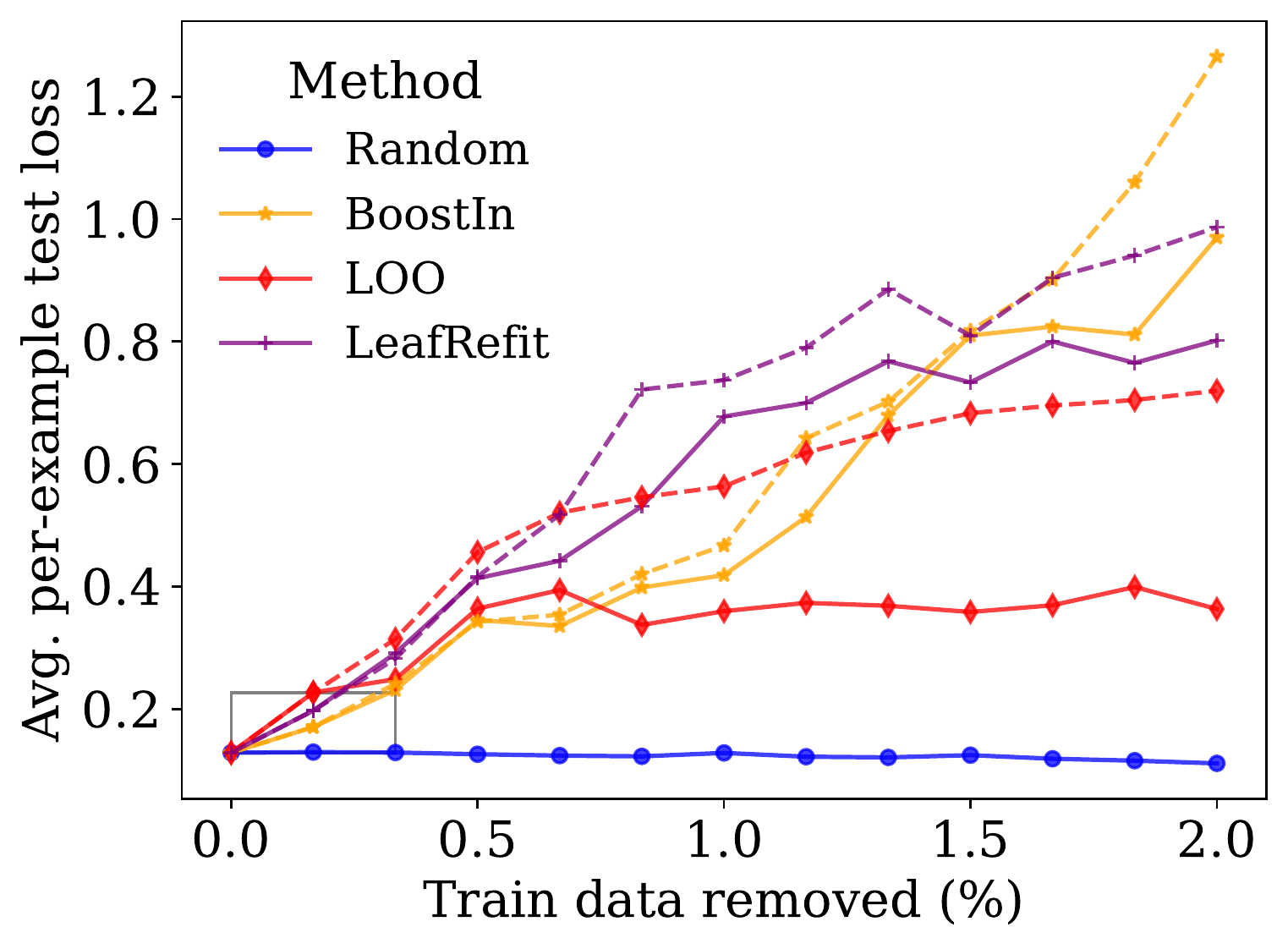}
  \caption{LGB: Energy}
  \label{fig:lgb_energy_reinf}
\end{subfigure}%
\begin{subfigure}{\tw\textwidth}
  \centering
  \includegraphics[width=1.0\linewidth]{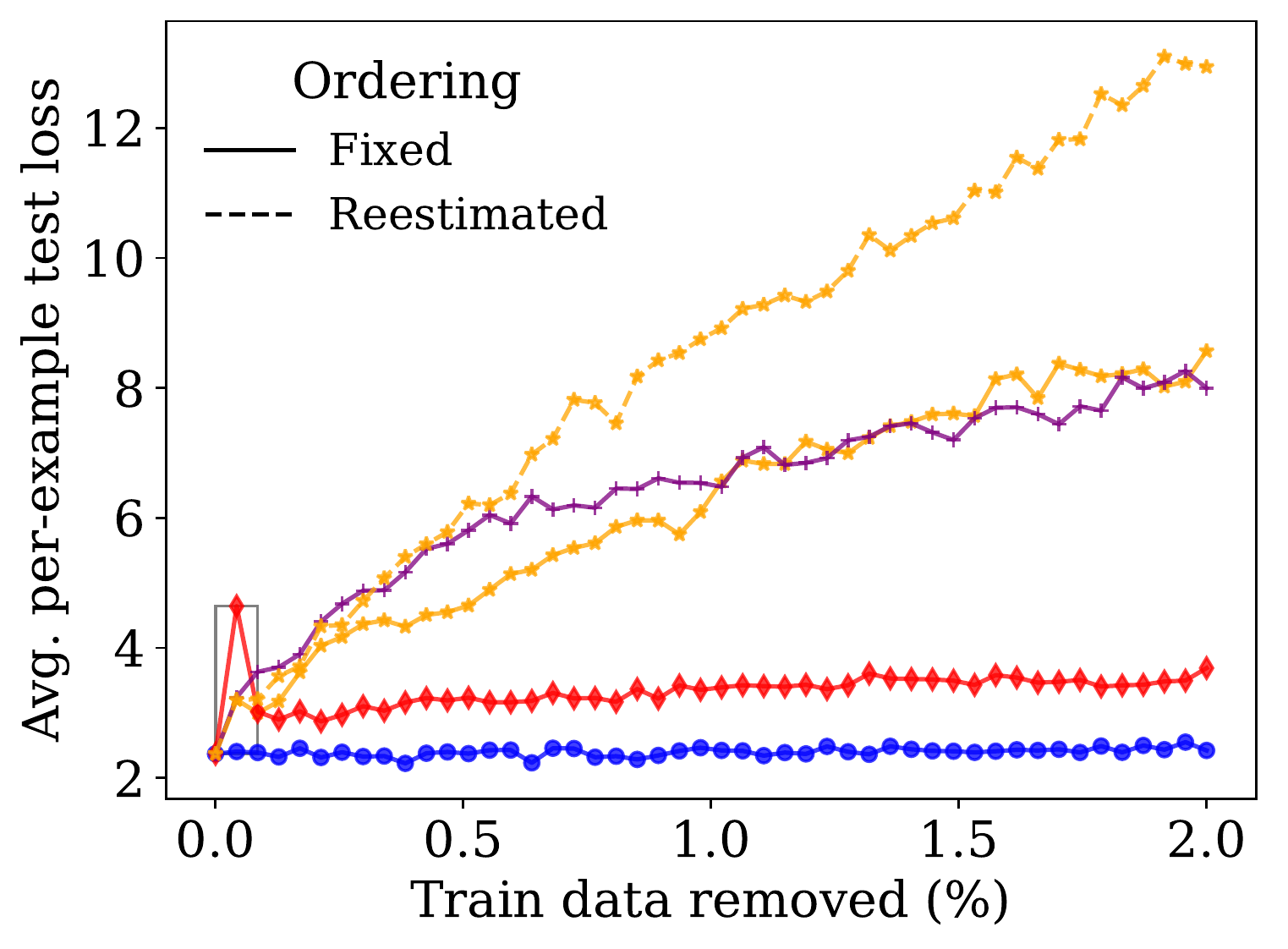}
  \caption{LGB: Life}
  \label{fig:lgb_life_reinf}
\end{subfigure}
\\
\begin{subfigure}{\tw\textwidth}
  \centering
  \includegraphics[width=1.0\linewidth]{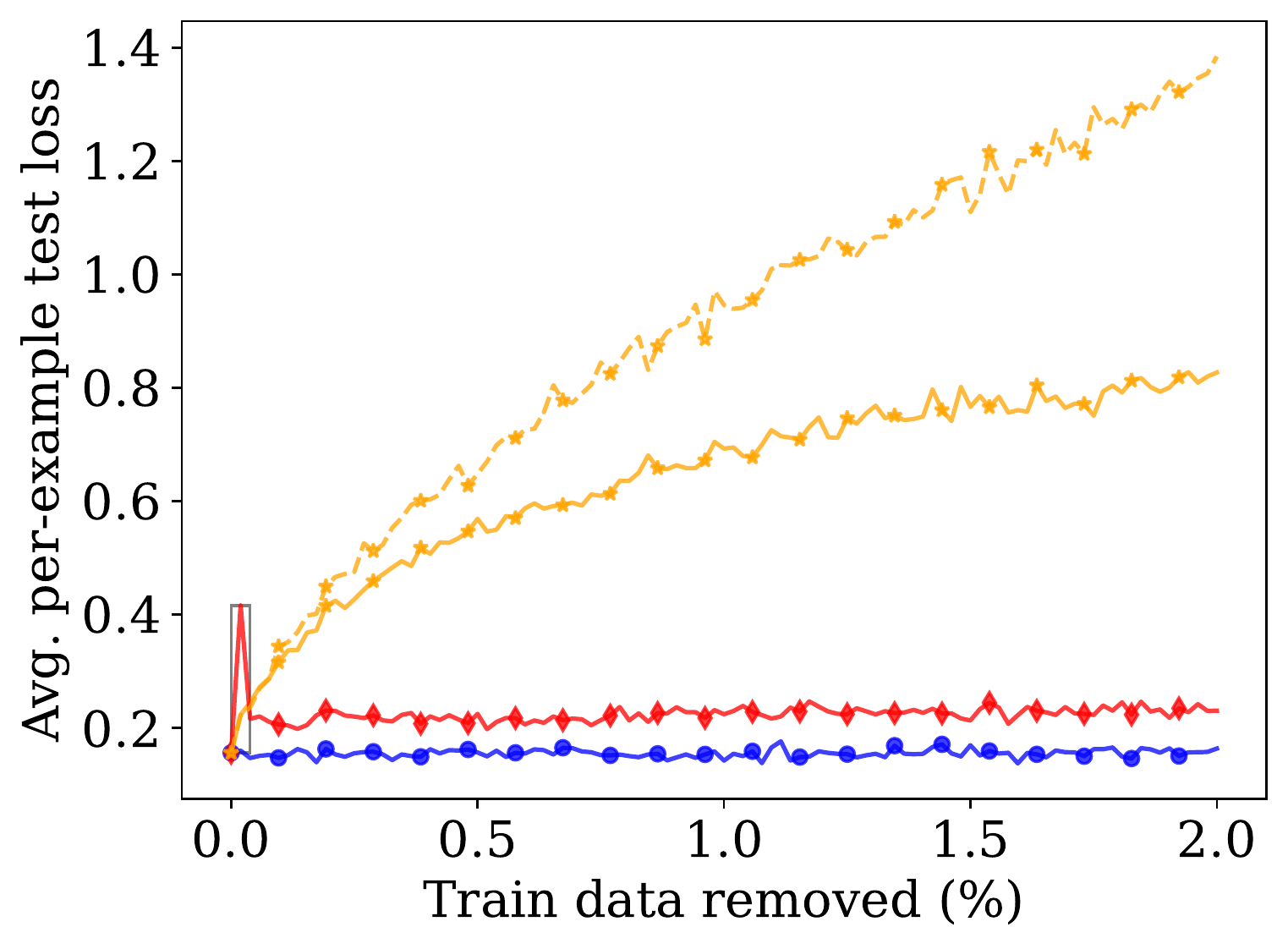}
  \caption{LGB: Wine}
  \label{fig:lgb_wine_reinf}
\end{subfigure}%
\begin{subfigure}{\tw\textwidth}
  \centering
  \includegraphics[width=1.0\linewidth]{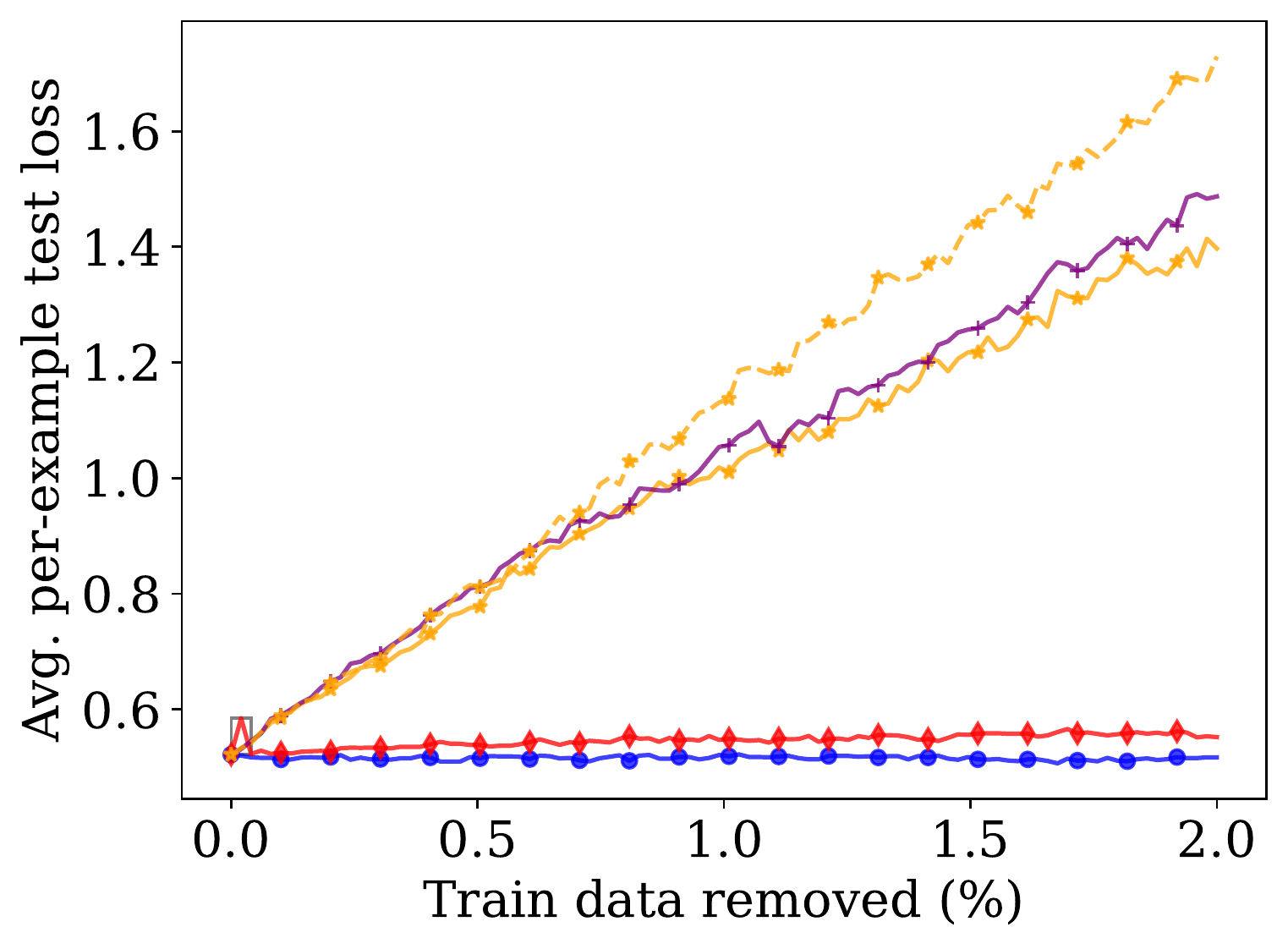}
  \caption{XGB: COMPAS}
  \label{fig:xgb_compas_reinf}
\end{subfigure}
\\
\begin{subfigure}{\tw\textwidth}
  \centering
  \includegraphics[width=1.0\linewidth]{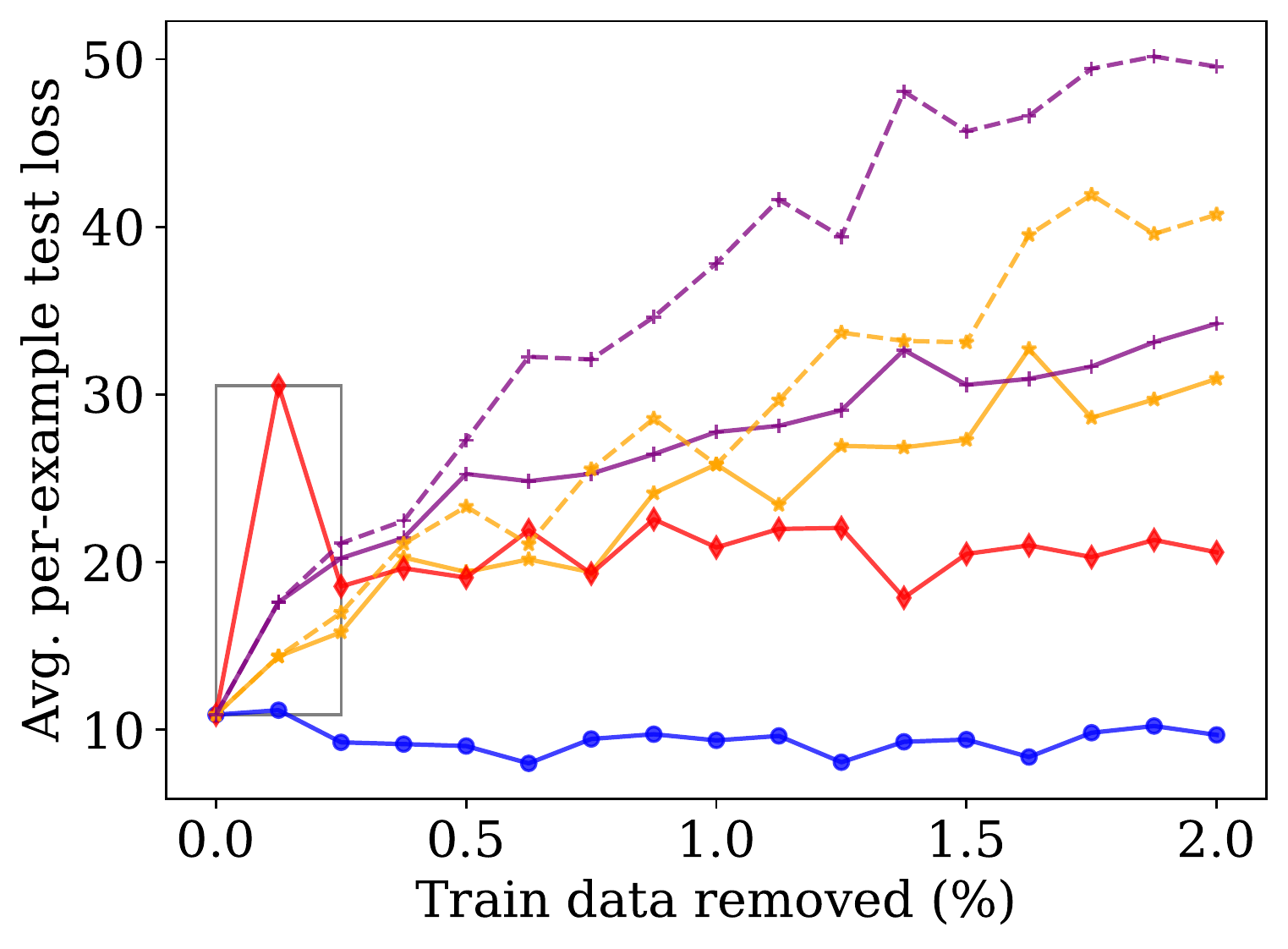}
  \caption{XGB: Concrete}
  \label{fig:xgb_concrete_reinf}
\end{subfigure}%
\begin{subfigure}{\tw\textwidth}
  \centering
  \includegraphics[width=1.0\linewidth]{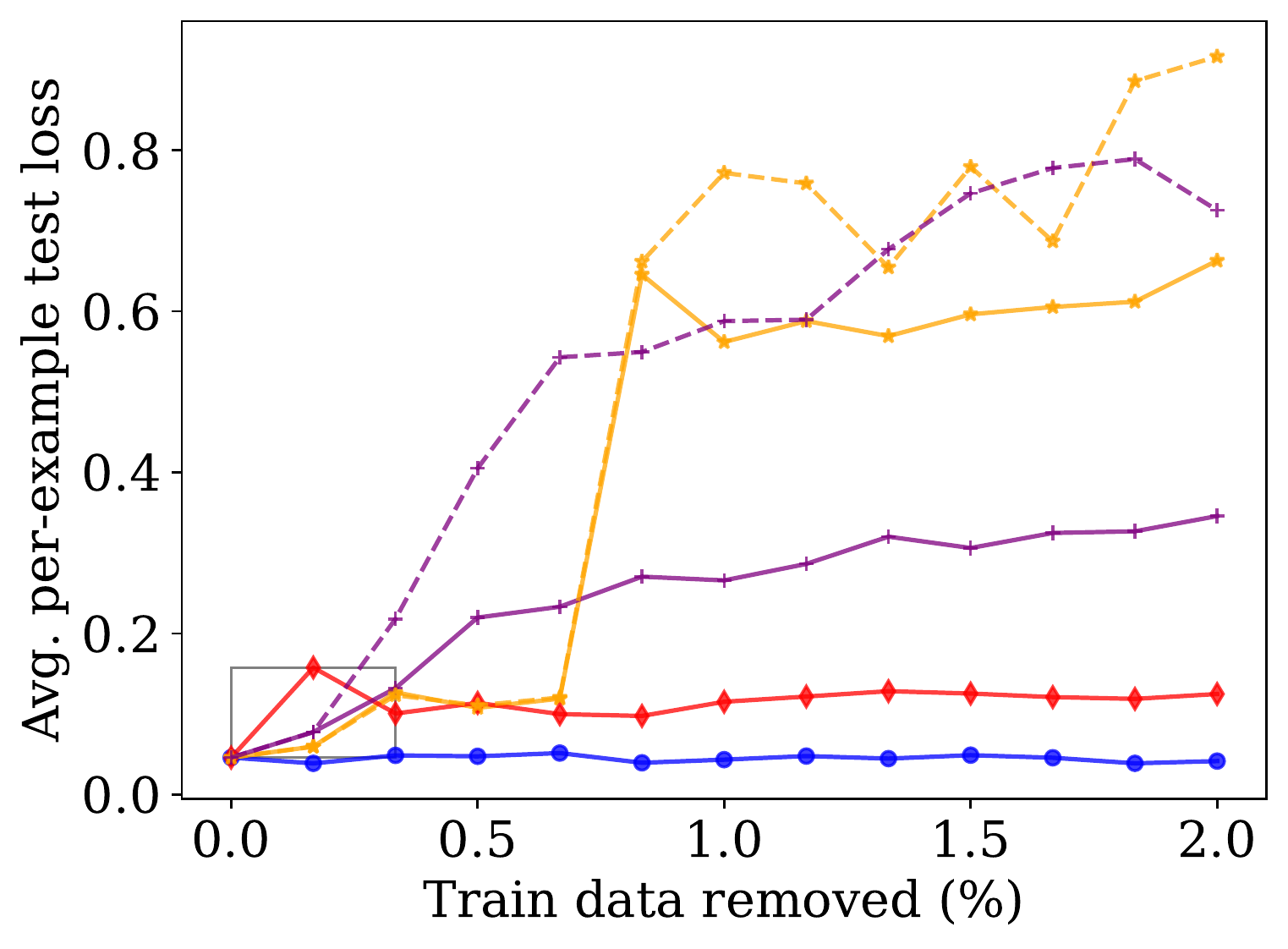}
  \caption{XGB: Energy}
  \label{fig:xgb_energy_reinf}
\end{subfigure}
\\
\begin{subfigure}{\tw\textwidth}
  \centering
  \includegraphics[width=1.0\linewidth]{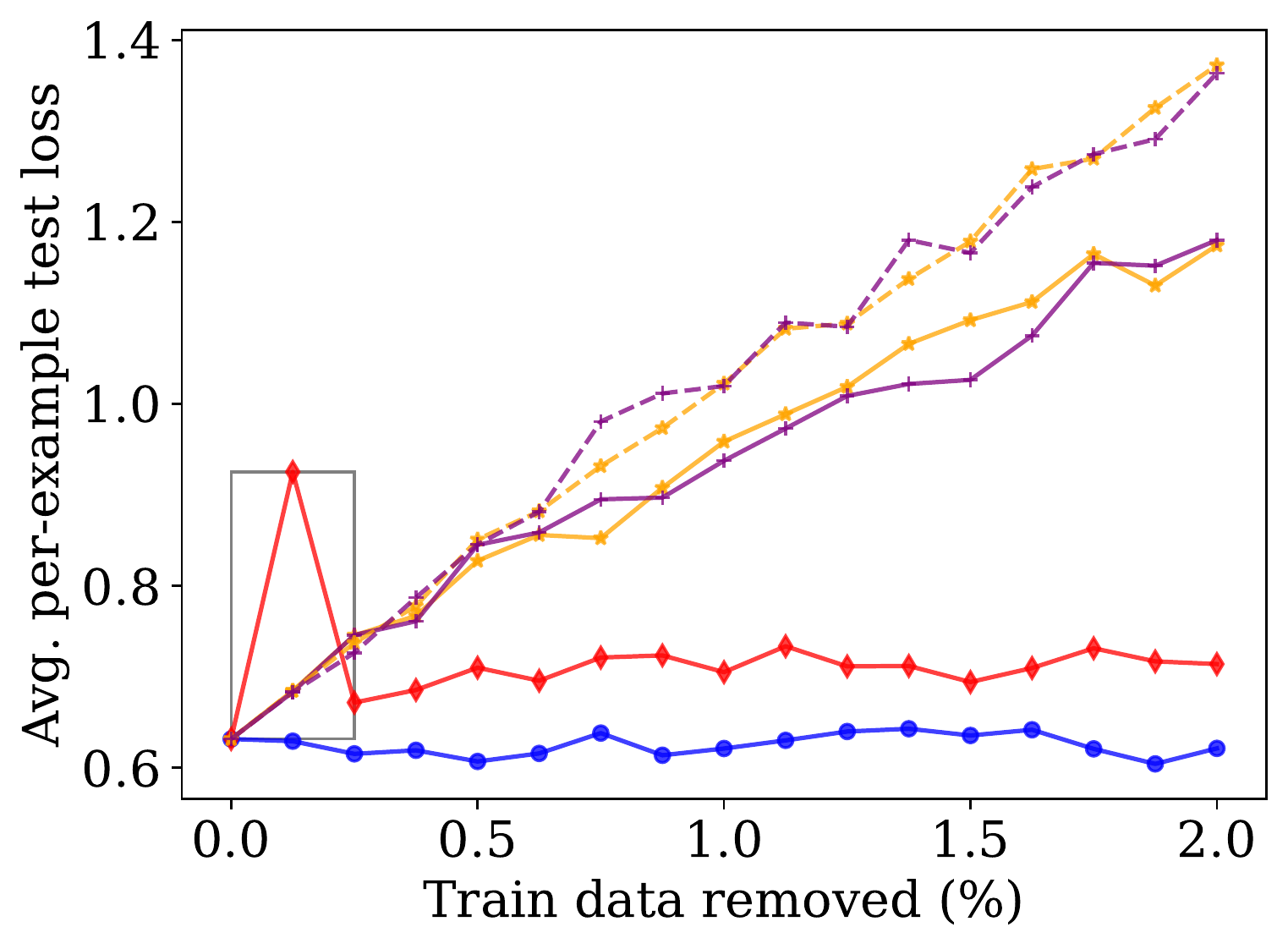}
  \caption{XGB: German Credit}
  \label{fig:xgb_german_credit_reinf}
\end{subfigure}%
\begin{subfigure}{\tw\textwidth}
  \centering
  \includegraphics[width=1.0\linewidth]{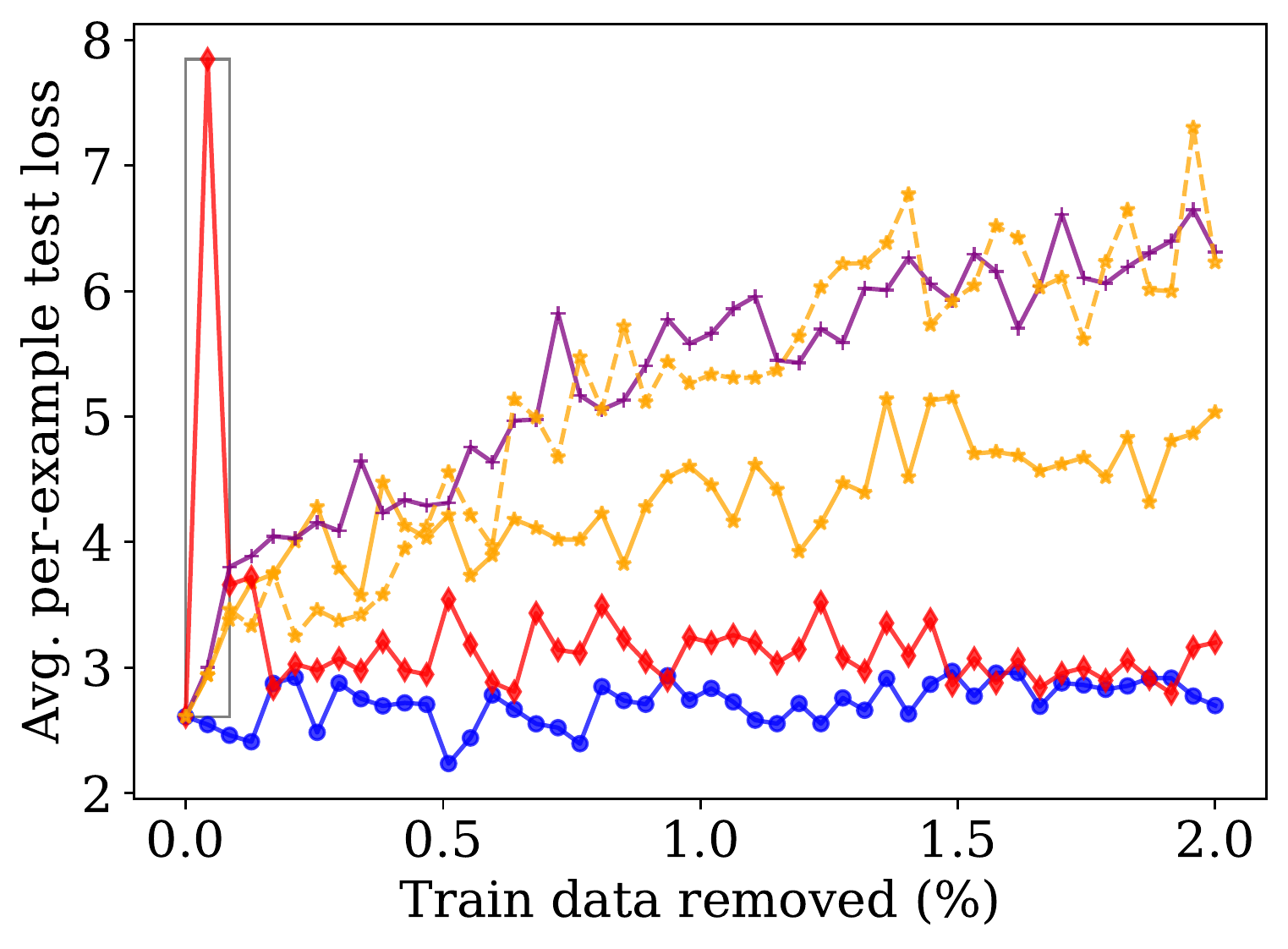}
  \caption{XGB: Life}
  \label{fig:xgb_life_reinf}
\end{subfigure}
\caption{Change in loss (mean over 100 test instances) as training examples are removed \emph{one at a time}. Higher is better. The gray box highlights the ability of LOO to identify the single-most influential training example on the given test instance.}
\label{app_fig:loo_fragility}
\end{figure}

\bibliography{main}

\end{document}